# Empirical and Experimental Insights into Data Mining Techniques for Crime Prediction: A Comprehensive Survey

Kamal Taha*

Department of Electrical Engineering and Computer Science, Khalifa University, kamal.taha@ku.ac.ae

This survey paper presents a comprehensive analysis of crime prediction methodologies, exploring the various techniques and technologies utilized in this area. The paper covers the statistical methods, machine learning algorithms, and deep learning techniques employed to analyze crime data, while also examining their effectiveness and limitations. We propose a methodological taxonomy that classifies crime prediction algorithms into specific techniques. This taxonomy is structured into four tiers, including methodology category, methodology sub-category, methodology techniques, and methodology sub-techniques. Empirical and experimental evaluations are provided to rank the different techniques. The empirical evaluation assesses the crime prediction techniques based on three criteria, while the experimental evaluation ranks the algorithms that employ the same sub-technique, the different sub-techniques that employ the same technique, the different techniques that employ the same methodology sub-category, the different methodology sub-categories within the same category, and the different methodology categories. The combination of methodological taxonomy, empirical evaluations, and experimental comparisons allows for a nuanced and comprehensive understanding of crime prediction algorithms, aiding researchers in making informed decisions. Finally, the paper provides a glimpse into the future of crime prediction techniques, highlighting potential advancements and opportunities for further research in this field.

## 1 INTRODUCTION

Criminal activity is a global issue affecting developed and underdeveloped nations, with severe consequences on the economy and individuals' well-being, leading to social and societal problems [1, 2]. Crime poses significant challenges for many societies in large cities worldwide. Its presence in neighborhoods can have adverse effects on the physical and mental health of urban residents, increasing their risk of violence and potentially leading to depression from constant exposure to assaults, blows, and gunfire [3]. Illegal activities such as terrorism, armed robbery, and drug violations often involve multiple offenders who are linked through various relationships such as kinship, friendship, co-workers, or business associates [4]. These criminals form a network in which they interact and play different roles in committing illegal acts [5]. Terrorist networks often consist of members with common religious beliefs or joint terrorist training, fostering trust and cooperation in attack planning [6]. Understanding the associations between these entities is crucial for uncovering criminal activities and fighting crimes. To achieve this, investigators use a technique called link analysis [7, 4, 8], which generates leads and uncovers information that may be concealed within a criminal network.

Local authorities prioritize reducing crime rates through active plans and police intervention. Predicting crime is essential for enhancing public safety and efficiently allocating resources [9]. However, it's difficult because criminal activities vary over time and location, and the relationships between various crime types (like Theft, Robbery, Assault, Damage) change and offer detailed insights into criminal behavior [9]. Mathematical, statistical, or computational models capable of predicting crime events in advance can aid in generating preventive measures for high-risk areas and speeding up the process of solving crimes. Many studies have been conducted worldwide, and many cities have made their data available for such investigations. Assuming crime prediction can be based solely on hotspot maps overlooks evidence that integrating time factors with spatial data greatly improves predictive accuracy [10]. Machine learning models specifically adjusted for spatiotemporal prediction in low population density regions, addressing class imbalance, significantly improved crime hotspot prediction accuracy [11]. Deep learning and traditional learning methods, such as classification, clustering, regression, and social network-based methods, are used to predict and prevent crimes by identifying hot spots and potential times of occurrence.

Data-driven techniques, such as machine learning and statistical modeling, have shown promise in predicting when and where crimes are likely to occur, enabling law enforcement to take proactive measures. Integrating cross-type and spatiotemporal correlations through comprehensive urban data analysis has demonstrated that recognizing the complex connections among various crime types and their evolution over time and across locations can greatly improve prediction accuracy [12]. This insight into the dynamic nature of criminal activity underscores the importance of adopting multifaceted analytical approaches to enhance the effectiveness of crime prediction models [12]. Various techniques and research are explored for predicting crimes. Given the multitude of crime prediction techniques available, it is imperative to undertake a comprehensive examination of these methodologies. This review paper is crucial for comparing their effectiveness and conducting experiments to explore potential approaches for advancement.

Our aim in this survey paper is to conduct a thorough evaluation that offers valuable insights into the strengths and weaknesses of various crime prediction techniques, which will contribute to future research in this domain. Our approach provides a meticulous and detailed understanding of modern algorithms for predicting crimes and their practical applications. By conducting *empirical* and *experimental* assessments, we present a comprehensive evaluation of contemporary and state-of-the-art crime prediction algorithms. To facilitate a more precise and accurate classification of these techniques, we propose a taxonomy based on methodology, organizing algorithms into nested hierarchical, specific, and finely grained categories.

To carry out our analysis, we extensively reviewed more than 150 papers, focusing on finely grained and specific techniques. We sourced these papers from esteemed publishers such as IEEE and ACM, specifically searching for those that describe algorithms utilizing the techniques of interest. In order to ensure the relevance and representativeness of our paper selection, we ranked them based on their recency and novelty, selecting the top papers that provided a diverse range of information on the techniques. Through this comprehensive evaluation, our goal is to offer insights into the strengths and weaknesses of different crime prediction techniques and provide guidance for future research in this field.

* Corresponding author.

## 1.1 Motivation and Key Contributions

This paper addresses a significant challenge in the field of crime prediction algorithms: the difficulty of categorizing these algorithms into groups that are both specific and informative. The prevalent method of grouping algorithms into broad, non-specific categories leads to confusion, inaccurately lumping unrelated algorithms together, and evaluating them using the same metrics. To overcome this limitation, we propose a new methodological taxonomy that hierarchically classifies algorithms based on their techniques, offering a more detailed and nuanced categorization. Our taxonomy divides crime prediction algorithms into four main classes: classification, clustering, regression, and social network-based methods. Each class is further broken down into four tiers, ranging from broad methodology categories to specific sub-techniques, as illustrated in Fig. 1. This hierarchical structure—Methodology Category → Methodology Sub-category → Methodology Techniques → Methodology Sub-techniques—enables a more precise classification, facilitating better comparison and evaluation of algorithms.

The rationale for adopting this hierarchical taxonomy is multifaceted:
1. **Enhanced Comparability**: It allows for more granular comparison within the same sub-category or technique level, helping identify the strengths and weaknesses of different algorithms.
2. **Facilitates Evolution Tracking**: The taxonomy's adaptability accommodates new categories and sub-categories, enabling tracking of the field's evolution without disrupting the overall structure.
3. **Promotes Interdisciplinary Collaboration**: By providing a common framework, the taxonomy fosters collaboration across fields such as criminology, computer science, and statistics.
4. **Improves Algorithm Development**: Identifying specific areas needing innovation guides focused research efforts, advancing the development of crime prediction algorithms.
5. **Aids in Systematic Literature Review**: The structured approach of the taxonomy streamlines the literature review process, highlighting gaps and future research opportunities.
6. **Enhances Understanding of Algorithmic Impact**: A detailed classification aids in assessing algorithms' societal impacts, including ethical considerations and biases.
7. **Facilitates Customization and Adaptation**: Understanding the taxonomy helps practitioners customize algorithms to specific needs, enhancing their practical application.
8. **Standardizes Evaluation Metrics**: Detailed classification enables the standardization of evaluation metrics, ensuring fair and accurate performance assessments.

However, it's important to note that not all methods fit neatly into a four-level structure. Some methods are better represented as techniques due to their specificity, and forcing a hierarchical structure could obscure their unique characteristics. Recognizing this, our taxonomy maintains clarity and precision by avoiding unnecessary complexity.

The comprehensive survey conducted in this paper, along with the proposed taxonomy, enables more accurate evaluation and comparison of algorithms. We also perform **empirical and experimental evaluations** to assess the effectiveness of the different approaches. These evaluations compare various algorithmic categories and techniques, providing a nuanced understanding of available algorithms and guiding researchers in selecting the most appropriate method for their needs.

In conclusion, this paper makes a significant contribution to the field of crime prediction by introducing a methodological taxonomy that offers a more systematic and comprehensive approach to categorizing algorithms. This taxonomy not only aids in the precise evaluation and development of crime prediction algorithms but also facilitates the field's growth by encouraging interdisciplinary collaboration and innovation. By adopting this taxonomy, researchers can make well-informed decisions, contributing to the development of more accurate/effective crime prediction algorithms.

## 1.2 Current Survey Papers on Crime Prediction

Thomas and Sobhana [13] conducted a study on worldwide strategies for crime prediction and forecasting. They classified the techniques into three groups: neural network, statistical, and Spatiotemporal approaches. The effectiveness of these techniques was assessed based on precision and accuracy in different domains. Additionally, they explored papers related to AI and deep learning that utilize statistical methods to anticipate the timing and manner of crime occurrences. Yang et al. [14] conducted a review of existing research that focuses on uncovering connections between crime incidents and extracting information about entities involved. They categorized techniques used for identifying links between different crime incidents and extracting relations between entities. Kawthalkar et al. [15] conducted a comprehensive examination of technological advancements aimed at improving crime prevention methods beyond existing systems. They investigated various techniques such as crime mapping and predictive policing, considering criminological theories. Kounadi et al. [16] performed a systematic literature review on Spatial Crime Forecasting, analyzing empirical studies and emphasizing the concepts and methodologies employed. Their objective was to address research inquiries regarding the significance of space in forecasting, the effectiveness of predictive techniques, and strategies for validating models.

Saravanan et al. [17] explored the possibility of analyzing extensive data repositories to explore the relationship between various socioeconomic factors and criminal incidents. Their main goal was to detect anomalies and patterns within the data and create computational models that could forecast crime using data mining methods. Prabakaran and Mitra [18] conducted a survey on diverse data mining techniques employed in crime prediction. Ramesh and Maheswari [19] conducted research on various categories of cybercriminal activities, which encompassed targeting individuals, property, and government institutions. They investigated preventive measures to combat cybercrime. Alsaqabi et al. [20] performed a study aiming to forecast the key factors influencing crime in Saudi Arabia. Belesiotis et al. [21] conducted a study to assess the effectiveness of different types of data in forecasting crime levels. Their focus was to enhance accuracy by integrating multiple data sources. The researchers also explored the influence of these predictions on identifying areas with high crime rates. Bshayer et al. [22] investigated the application of machine learning models for predicting specific crimes in a particular location. They examined available data on recurring incidents and employed Decision Tree and Naive Bayes algorithms for analysis. Zhuo and Libed [23] uncovered the main factors contributing to crime rates in Rizal Province. They utilized the statistical technique of Pearson Correlation to examine the relationships between various factors and the occurrence of criminal activities.

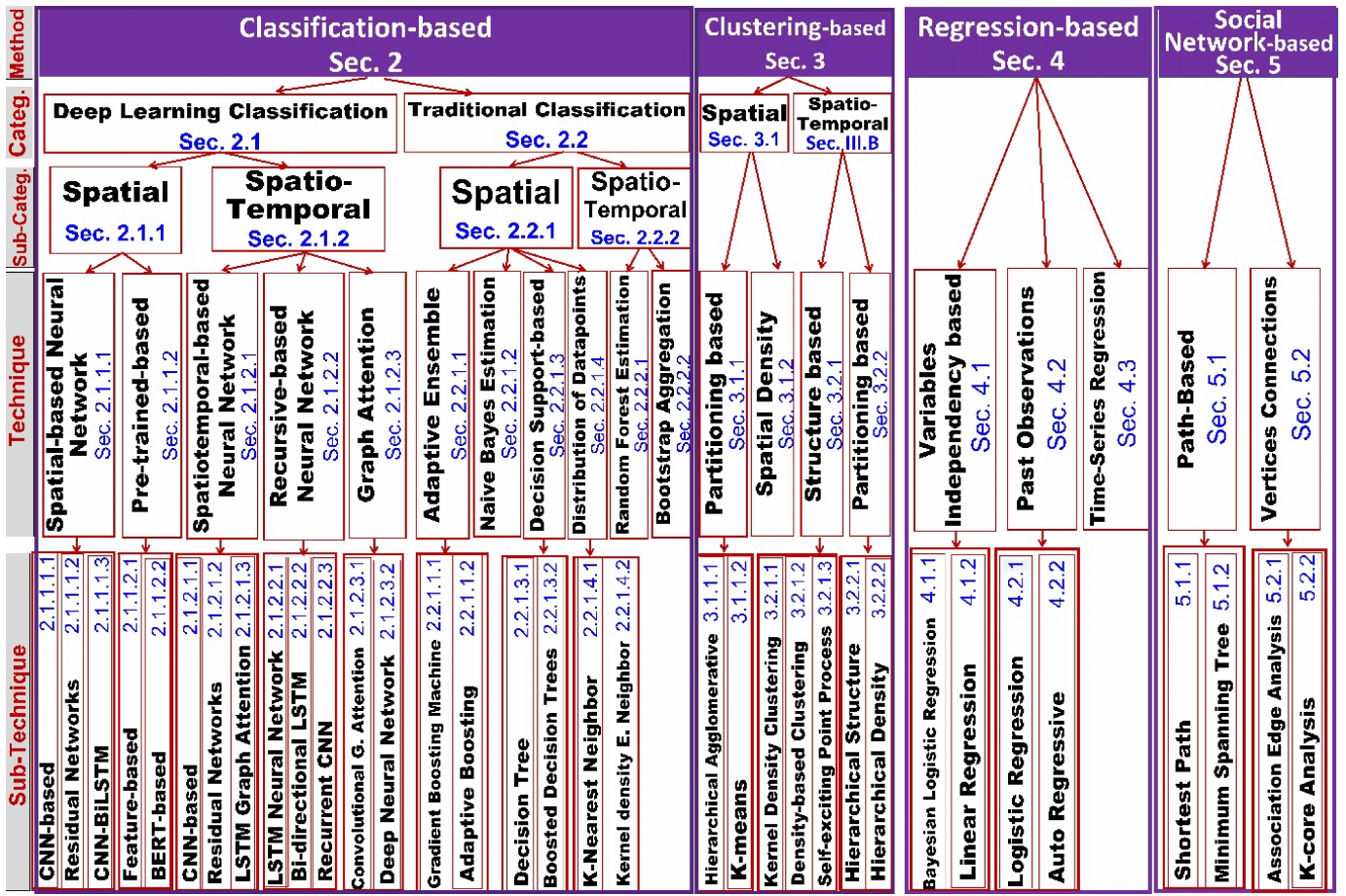

**Fig. 1.** Our methodology-based taxonomy that categorizes crime prediction algorithms into fine-grained classes in a hierarchical manner, as follows: method → category → sub-category → technique → sub-technique. For each method, category, sub-category, technique, and sub-technique, the figure also shows the section number in the manuscript that discusses it.

## 2 CRIME PREDICTION BASED ON CLASSIFICATION

### 2.1 Classification Based on Deep Learning Methods

#### 2.1.1 Spatial-Based Classification

##### 2.1.1.1 Neural Network-Based Spatial Classification

###### 2.1.1.1.1 Convolutional Neural Networks (CNN)-Based Spatial Classification

The integration of spatial-based CNNs in crime prediction capitalizes on the analysis of spatial data, which includes crime incident locations and attributes. This approach develops predictive models utilizing a broad spectrum of information, encompassing geographical coordinates of crime incidents, demographic characteristics of the area, and socio-economic factors. The CNN architecture plays a pivotal role in this process through its convolution layer, which filters the input data to identify patterns and features relevant to crime prediction. Subsequent pooling layers streamline the data further by reducing its dimensionality and extracting key features, ensuring that the most significant information is retained for analysis. The fully connected layers then take over, classifying the input by transforming the refined output from the convolution and pooling layers into scores. These scores categorize the input into various categories, such as crime type or location, based on the patterns recognized in the data. Finally, an activation function, like softmax, is applied to these scores to derive a final probability distribution over the different categories. This sophisticated process enables the accurate prediction of crime by analyzing and interpreting complex spatial data through the lens of CNNs, thus providing a powerful tool for law enforcement and public safety agencies.

*The rationale behind the usage of the technique*: Crime data exhibits complex and non-linear relationships between the input features and the occurrence of crimes. CNNs, with their multiple layers and non-linear activation functions, are suited to capture these relationships. They learn hierarchical representations of the data, enabling the model to discern intricate patterns that might not be easily identifiable with traditional linear models. The ability of CNNs to automatically learn relevant features from the input data makes them highly effective for crime prediction.

*Conditions for the optimal performance of the technique*: Incorporating deep contextualized word representations like ELMo and BERT can elevate the feature representations used in the CNN by providing more informative and contextualized input to the model which are not dependent on the decoder, leading to a general enhancement in the performance. This improves the model's ability to understand the meaning of words and sentences. By leveraging deep contextualized word representations, the CNN model can better capture nuances and semantic relationships within the crime data, enabling more accurate predictions.

*Limitations of the technique*: CNNs necessitate a substantial amount of training data to perform effectively, which can be a problem due to privacy concerns in crime data availability. CNNs are not impervious to partiality or inaccuracies. Utilizing CNN-based analysis for crime detection may result in erroneous detections or missed crimes, as factors like lighting and shadows.

TABLE 1: FEATRURING RSEARCH PAPERS (PP.) THAT HAVE EMPLOYED **CNN-BASED SPATIAL TECHNIQUES** AND THEIR EVALUATION IN TERMS OF SCALABILITY (SC.), INTERPRETABILITY (IN.), ACCURACY (AC.), AND EFFICIENCY (EF.)

| Pp. | Sc. | In. | Ac. | Ef. | Description |
|---|---|---|---|---|---|
| Duan [24] | Fair | Acceptable | Good | Fair | The authors proposed a Spatiotemporal Crime Network (STCN) that uses deep CNNs for automatic crime-referenced feature extraction, aiming to predict crime locations and timings by analyzing complex relationships with space, time, and environments. This model predicts the next day's crime risk in urban areas using high-dimensional data. |
| Fu et [25] | Unsatisfactory | Acceptable | Fair | Unsatisfactory | The authors proposed a CNN based approach for inferring crime rankings from street view images using a preference learning framework, improving on previous safety level prediction methods by utilizing spatiotemporal correlations between images and crime data. It aligns street view images with urban perceptions and accurately maps them to crime rankings. |
| Wei et [26] | Fair | Unsatisfactory | Fair | Unsatisfactory | The authors proposed CrimeSTC, a deep learning framework designed for predicting urban crime, leveraging spatial, temporal, and categorical data through a blend of CNN, GRU, fully connected layers, and graph convolutional networks. It integrates dynamic and static data inputs via a joint training module to accurately forecast crime numbers. |
| Onan [27] | Fair | Fair | Good | Acceptable | The author introduced a deep learning approach for sentiment analysis on Twitter product reviews, leveraging a CNN-LSTM architecture enhanced with TF-IDF weighted Glove word embeddings. The architecture combines convolution layers with varying N-gram sizes and an LSTM layer, demonstrating superior predictive performance over traditional methods. |
| Patel [28] | Acceptable | Fair | Good | Good | The authors introduced the Dimension-Based Generic Convolution Block (DBGC), a flexible component for CNNs that enables generic architecture with dimension-wise kernel selection for height, width, and depth, utilizing the separable convolution principle. It's adaptable for use across multiple dimensions simultaneously, with a dimension selector block. |

### 2.1.1.1.2 Residual Network-Based Spatial Classification

In a spatial-based residual network designed to process spatial data, the input is typically a grid or map representing a specific geographical region, populated with crime data associated with various locations. This network aims to predict crime rates or types based on the provided input by leveraging advanced neural network techniques. To enhance the network's efficiency and learning capability, it incorporates Rectified Linear Unit (ReLU) nonlinearities and batch normalization between double- or triple-layer skips. Including ReLU nonlinearities between the skip connections is a strategic approach that significantly accelerates the network's convergence during the training phase. Furthermore, the synergistic combination of ReLU nonlinearities and batch normalization not only contributes to improved performance but also ensures faster convergence in these deep neural networks. This methodological approach is essential for processing and learning from spatial data effectively, enabling the network to make accurate predictions about crime rates or types within the geographical area under study.

*The rationale behind the usage of the technique*: Residual Networks resolve the issue of vanishing gradients by optimizing the training of neural networks, which reduces accuracy saturation. This enhancement can improve the performance of the networks for crime prediction. This can lead to improved accuracy and reduced saturation, which is important for crime prediction. By mitigating the problem of vanishing gradients, Residual Networks enable more effective training of neural networks, enhancing their performance and accuracy in crime prediction tasks.

*Conditions for the optimal performance of the technique*: If a nonlinear layer is omitted, it may enable the network to comprehend more intricate connections among the input characteristics. This also reduces the number of parameters in the model, making it more efficient to train. By creating a shortcut that bypasses a nonlinear layer, the network can effectively retrieve information from a prior layer that might have been lost. By incorporating skip connections that bypass nonlinear layers, the network gains the ability to capture complex relationships within data.

*Limitations of the technique*: The technique of skip connections in neural networks has several constraints, including the following: (1) involving high computational complexity, (2) being prone to overfitting (which can happen when the model is overly complex or there is inadequate training data), (3) utilizing high memory, (4) presenting challenges during training, and (5) having limited scalability. The above limitations need to be carefully addressed when implementing the approach to crime prediction.

TABLE 2: FEATURING RSEARCH PAPERS (PP.) THAT HAVE EMPLOYED **RESIDUAL NETWORK-BASED SPATIAL TECHNIQUES** AND THEIR EVALUATION IN TERMS OF SCALABILITY (SC.), INTERPRETABILITY (IN.), ACCURACY (AC.), AND EFFICIENCY (EF.)

| Pp. | Sc. | In. | Ac. | Ef. | Description |
|---|---|---|---|---|---|
| Hara [29] | Fair | Unsatisfactory | Acceptable | Fair | The authors introduced a 3D CNN architecture based on ResNets, enhancing action representation. They provided detailed insights into the training process of our 3D ResNets. They emphasized the effectiveness of ResNets with 3D convolutional kernels, showcasing robust performance even with the large parameter count, thanks to training on Kinetics. |
| Jan [30] | Fair | Fair | Acceptable | Acceptable | The authors introduced DeepCrime, a spatiotemporal CNN method designed for detecting crimes in video footage. They conducted a comparison of two leading networks for spatiotemporal analysis, C3D and ResNet3D for binary crime classification. Findings indicate C3D's superiority in binary classification, while ResNet3D excels in multi-class scenarios. |
| Matereke [31] | Unsatisfactory | Unsatisfactory | Acceptable | Good | The authors investigated crime classification using 3D deep learning algorithms, specifically 3D CNN and 3D Residual Network (ResNet-18). The study focuses on binary classification to predict the occurrence of theft, utilizing a 30-day temporal window for training and evaluating the impact of spatial grid resolutions on model performance. |

*2.1.1.1.3 CNN-BiLSTM-Based Spatial Classification*

The Spatial-Based CNN-BiLSTM is an advanced deep learning model that integrates the strengths of Convolutional Neural Networks (CNNs) and Bidirectional Long Short-Term Memory (BiLSTM) networks, aiming to accurately predict crime rates in specific areas. The model operates by first utilizing the CNN layer to extract spatial features from crime data. This is achieved through the use of filters for convolutional operations on the input, which produce a detailed feature map. These extracted features are subsequently passed to the BiLSTM layer, which is adept at understanding and predicting temporal dynamics, thanks to its bidirectional analysis of data sequences. The Spatial-Based CNN-BiLSTM model efficiently predicts crime rates by combining CNNs for spatial analysis and BiLSTMs for temporal insights, enhanced by ReLU nonlinearities and batch normalization for faster learning and stability. This setup accelerates training convergence and enhances performance, making the model highly effective in utilizing historical crime data for accurate future crime predictions.

*The rationale behind the usage of the technique*: By employing BiLSTMs, the system can access a greater amount of data, leading to an enhanced contextual understanding of the algorithm. This means that the network can consider not only the words that come before a given word, but also the words that come after it. The network can better understand the context in which the word is used. This enables the network to grasp the intricate context in which each word is employed.

*Conditions for the optimal performance of the technique*: Incorporating 2D convolution in BiLSTM can boost contextual understanding and feature capture, leading to enhanced classification performance. By augmenting BiLSTM with CNN layers, aspect category classification can be tackled. Local patterns and features in the data can be identified and classified using 2D convolutional layers. This enables the classification of local patterns and features in data.

*Limitations of the technique*: The incorporation of 2D convolutional layers in the BiLSTM architecture may result in slower training speed and increased computational requirements compared to using LSTM alone. This is due to the heightened intricacy and computational requirements linked with the integration of 2D convolutional layers. The incorporation of such layers may prolong the training process and cause a decrease in performance speed during inference.

**TABLE 3:** FEATRURING RSEARCH PAPERS (PP.) THAT HAVE EMPLOYED **CNN BILSTM-BASED SPATIAL TECHNIQUES** AND THEIR EVALUATION IN TERMS OF SCALABILITY (SC.), INTERPRETABILITY (IN.), ACCURACY (AC.), AND EFFICIENCY (EF.)

| Pp. | Sc. | In. | Ac. | Ef. | Description |
|---|---|---|---|---|---|
| Krung. [32] | Unsatisfactory | Good | Acceptable | Good | The authors proposed crime prediction models using unidirectional and bidirectional LSTM networks. They presented a hybrid model combining legal expertise and deep learning (including LSTM, BiLSTM, and SMOTE techniques) to analyze specific sections of the Thai Criminal Code. |
| Gandhi [33] | Unsatisfactory | Good | Good | Fair | The authors highlight the critical role of SQL injection detection in database security, proposing a superior CNN-BiLSTM hybrid model over other machine learning algorithms, and recommends combining such models with firewalls and static code analysis for improved SQL injection protection. |
| Nidhi [34] | Fair | Fair | Acceptable | Fair | The authors examined the use of CNN-BiLSTM and other ML models for crime prediction. They aimed to enhance city safety in Boston by employing various ML models and time series analysis to predict and improve crime prevention. They performed exploratory data analysis, implemented multiple ML algorithms, conducted time series analysis, and forecasted crime trends. |
| Saha [35] | Fair | Acceptable | Acceptable | Fair | The authors presented a system for analyzing and visualizing crime information from large text repositories. It uses a CNN-BiLSTM network for extracting crime indicators and offers a query and retrieval system to provide users with insights into crime patterns and statistics. Users can navigate related articles and access aggregated statistics across various dimensions. |

2.1.1.2 Pre-Trained-Based Spatial Classification

*2.1.1.2.1 Feature-Based Spatial Classification*

Spatial-based pre-trained feature-based models, such as CNNs, offer a powerful approach to enhancing crime prediction through the incorporation of spatial mechanisms. By fine-tuning these models with geocoded crime data, they become capable of predicting the likelihood of criminal activities in specific areas. In contrast, there are alternative approaches to crime prediction that focus on pre-trained models. These models, designed specifically for crime prediction, rely on crime-related features such as crime rates, demographics, and socio-economic factors to estimate the likelihood of criminal incidents in particular regions. Feature embedding techniques use bidirectional LSTM (Long Short-Term Memory) to create word vectors that capture a word's contextual nuances by analyzing its surrounding context in both forward and backward directions, making it vital for many natural language processing tasks and applications.

*The rationale behind the usage of the technique*: This technique uses the context of the entire sentence to represent the embedding of a word, which causes the embeddings to grasp the meaning of the word within that context. Embeddings can be generated for the same word in different contexts across multiple sentences. Embeddings can be generated for the same word in different contexts and sentences.

*Conditions for the optimal performance of the technique*: To improve crime prediction using this technique, a biLSTM-enhanced model can be integrated into a supervised model by freezing the weights of the biLSTM and utilizing its enhanced representation as input for the Recurrent Neural Networks (RNNs) that can capture long-term dependencies and sequential patterns in input sequences.

*Limitations of the technique*: The meanings of words can vary based on the context in which they are used, and it can be challenging for the technique to interpret the intended meaning in all cases. The meanings of words in different sentences are influenced by subtle contextual factors. Also, the technique may not be efficient in processing unfamiliar words.

TABLE 4: FEATRURING RSEARCH PAPERS (PP.) THAT HAVE EMPLOYED FEATURED-BASED SPATIAL TECHNIQUES AND THEIR EVALUATION IN TERMS OF SCALABILITY (SC.), INTERPRETABILITY (IN.), ACCURACY (AC.), AND EFFICIENCY (FE.)

| Pp. | Sc. | In. | Ac. | Ef. | Description |
|---|---|---|---|---|---|
| Verma [36] | Unsatisfactory | Fair | Good | Good | The authors put forth two deep learning models for crime prediction, namely the standard CNN architecture and pre-trained CNN architectures. These models utilize the learning capabilities of facial images to infer crime prediction. They used standard CNN architecture and pre-trained CNN architectures, namely VGG-16, VGG-19, and InceptionV3. |
| Barathi [37] | Fair | Good | Acceptable | Unsatisfactory | The authors proposed a pre-trained CNN-based algorithm for crime prediction using 3D convolutional networks. The algorithm first performs feature extraction, then proceeds to the model training process and predicts the occurrence of violence. The study emphasized the importance of video frame pre-processing and training parameters like CNN network. |

*2.1.1.2.2 BERT-Based Spatial Classification*

Spatial-based BERT is an innovative approach that integrates spatial and text-based data to predict outcomes, particularly focusing on crime prediction. This technique leverages a comprehensive data set that includes crime-related text data, such as news articles, social media posts, and police reports, alongside spatial data, which encompasses crime hotspots, demographic information, and geographic features like parks and public transportation. By combining these diverse data sources, Spatial-based BERT is uniquely positioned to analyze and predict future crime patterns based on both the textual content of crime reports and the spatial distribution of crime incidents.

Spatial-based BERT integrates spatial data and text-based information, such as social media posts, news articles, police reports, and demographic information, with geographic features like crime hotspots, parks, and public transportation, to predict future crime patterns. This model employs a multi-layer transformer encoder and a token embedding layer to process the combined data. The encoder extracts significant representations by mapping input tokens into high-dimensional vector representations. These vectors capture the contextual and spatial nuances of the data, enabling the model to analyze the information thoroughly. By training with both historical crime data and current inputs, Spatial-based BERT provides a nuanced tool for accurate crime prediction, leveraging the detailed nature of text data alongside empirical, location insights.

*The rationale behind the usage of the technique*: A model can capture a significant amount of contextual information by analyzing the input text bidirectionally, considering both the preceding and following words and the surrounding context of each word. This is essential for classification of crime data, as it enables the model to deduce the meaning of words based on their context and predict the current word's label.

*Conditions for the optimal performance of the technique*: To improve the technique, one can incorporate a random sequence to forecast training tokens instead of solely depending on BERT's masked language model, which only predicts masked tokens. By sharing parameters between layers, the model can learn to extract more complex features from the input data, which can improve its ability to make accurate predictions.

*Limitations of the technique*: There are several limitations associated with the technique, including its extensive size due to the training structure and corpus, slow training time caused by the multitude of weights that require updating, longer computational time, and its design intended for integration with other systems. Therefore, the technique may require specialized optimization to improve its performance.

TABLE 5: FEATRURING RSEARCH PAPERS (PP.) THAT HAVE EMPLOYED BERT-BASED SPATIAL TECHNIQUES AND THEIR EVALUATION IN TERMS OF SCALABILITY (SC.), INTERPRETABILITY (IN.), ACCURACY (AC.), AND EFFICIENCY (EF.)

| Pp. | Sc. | In. | Ac. | Ef. | Description |
|---|---|---|---|---|---|
| Sandagiri [38] | Unsatisfactory | Good | Good | Acceptable | The authors presented an approach for predicting crime utilizing Twitter and weather data. The approach consists of two modules, namely the crime detection module and the crime prediction module. The detection module was developed using a BERT-based approach, while the prediction module was implemented using an Artificial Neural Network (ANN). |
| Alkhatib [39] | Fair | Unsatisfactory | Fair | Good | The authors introduced CAN-BERT, a network intrusion detection system leveraging the BERT model, tailored for the CAN bus protocol to identify cyber-attacks. It utilizes the "masked language model" training objective to understand the sequence of IDs in CAN bus communications by predicting masked CAN IDs using their surrounding context. |
| Rifat [40] | Unsatisfactory | Unsatisfactory | Good | Acceptable | The authors presented an approach for real-time SMS spam detection using pre-trained Google BERT models. By integrating shallow machine learning techniques with BERT algorithms, the study addresses phishing attacks through advanced SMS spam filtering. It reviews recent advancements, challenges, and future prospects in spam and phishing text detection. |
| Sandagiri [41] | Fair | Fair | Acceptable | Acceptable | The authors introduced an innovative method to enhance crime prediction by analyzing Twitter posts with the BERT model. It outlines the process of collecting Twitter data via API, employing NLP techniques for data cleaning, and then applying BERT for distinguishing crime-related content. It leverages real-time and user-generated content to predict crime patterns. |
| Onan [42] | Fair | Good | Good | Acceptable | The authors introduced a text classification method that merges hierarchical graph-based modeling with BERT for dynamic fusion across seven stages, from linguistic feature extraction to advanced text representation. This approach blends linguistic and domain knowledge with graph learning and attention mechanisms, showcasing the synergy between graph neural networks and transformer architectures in text classification. |

### 2.1.2 Spatiotemporal-Based Classification

#### 2.1.2.1 Neural Network-Based Spatiotemporal Classification

*2.1.2.1.1 CNN-Based Spatiotemporal Classification*

The spatiotemporal CNN architecture integrates multiple layers, including convolutional and pooling layers, to meticulously filter and condense input data for pattern and feature identification. These layers work in tandem to distill essential features from the data, setting the stage for advanced analysis. Following this, the fully connected layers come into play, transforming the refined output into scores that categorize the input into specific categories, such as crime types or locations. This classification process leverages the output layer, which is adeptly designed to produce a probability distribution over these categories. The final step in this computational process involves the application of an activation function. Depending on the nature of the classification—be it multi-class or binary—the model employs either softmax activation for multi-class classification, providing a probability distribution across various crime types or the likelihood of crime occurrence in specific areas, or a single sigmoid activation for binary classification, focusing on the presence or absence of crime. This comprehensive approach ensures a nuanced analysis of the data, enabling the model to offer precise predictions regarding crime patterns and occurrences.

Bhatt et al. [43] categorized the latest advancements in CNN architectures into eight distinct groups: (1) Spatial Exploitation-Based CNNs: this leverages varying filter sizes to extract different levels of granularity, (2) CNNs with multiple paths: this utilizes cross-layer connectivity to mitigate the vanishing gradient problem, (3) Deep CNN architectures: this is accomplished through additional nonlinear mappings and advanced feature hierarchies, demonstrate that network depth significantly enhances learning capabilities, (4) Feature-Map Exploitation Based CNNs: this enhances network generalization by dynamically selecting and adjusting feature maps, (5) Multi-Connection Depending on the Width: this switched from deep and narrow designs to wide and thin architectures, (6) Exploitation-Based Feature-Map CNNs: this enhance computer vision tasks by dynamically selecting and adjusting feature maps to improve network generalization, (7) CNNs based on attention: this utilizes hierarchical abstractions and focus on relevant features for improved localization and recognition, and (8) Dimension-Based CNNs: this improves computational efficiency by using depth-wise separable convolutions to separately encode spatial and channel-wise information.

*__The rationale behind the usage of the technique__*: (1) CNNs excel in learning spatial and temporal crime patterns, capturing local and broader contexts by incorporating time, thus predicting future crimes based on past data, (2) CNNs automatically learn relevant patterns from data, identifying complex patterns crucial for crime prediction, (3) CNNs handle diverse, high-dimensional crime datasets efficiently, analyzing various data types to predict and classify crime events, (4) Beyond classifying crimes, CNNs also localize future occurrences, aiding in resource allocation and targeted interventions, and (5) These models adapt and scale across different areas and crime types, updating with new data for robust crime analysis and prediction.

*__Conditions for the optimal performance of the technique__:* (1) Optimal CNN architecture and hyperparameter tuning are crucial for effective spatial-temporal pattern recognition, with experimentation encouraged for better generalization, (2) Choosing the right temporal resolution affects accuracy, where finer granularity captures detailed patterns but may increase data and computational demands, (3) Higher spatial resolution improves location prediction precision but also raises computational complexity, (4) Employing regularization and dropout prevents overfitting, aiding in model generalization to new data, and (5) Enhancing predictions by incorporating external data such as weather conditions or social media can provide additional contextual insights.

*__Limitations of the technique__:* (1) Crime patterns are shaped by complex spatial and temporal interactions. Accurately integrating these factors into CNNs requires sophisticated architectures, making design and training resource-intensive, (2) The complexity of spatiotemporal data and the high capacity of CNNs pose a risk of overfitting, especially with limited data, resulting in poor generalization to new data, (3) CNNs, particularly deep models, are often "black boxes" with complex structures that make understanding their decisions difficult, which can be problematic for applications requiring trust, such as crime prediction, (4) Using CNNs for crime prediction can amplify existing biases in the data, potentially leading to unfair or discriminatory outcomes, and (5) CNN models may not perform well across different regions due to varying crime patterns and demographics, requiring the development of region-specific models, which is resource-intensive.

**TABLE 6:** FEATURING RSEARCH PAPERS **(PP.)** THAT HAVE EMPLOYED **CNN-BASED SPATIOTEMPORAL TECHNIQUES** AND THEIR EVALUATION IN TERMS OF SCALABILITY **(SC.)**, INTERPRETABILITY **(IN.)**, ACCURACY **(AC.)**, AND EFFICIENCY **(EF.)**

| Pp. | Sc. | In. | Ac. | Ef. | Description |
|---|---|---|---|---|---|
| Zhuan [44] | Fair | Unsatisfactory | Acceptable | Good | The authors introduced the Spatio-Temporal Neural Network (STNN) to forecast crime hot spots by incorporating spatial data. STNN treats crime prediction as a classification task across geographic areas. By analyzing historical crime data, the model identifies key areas likely to become future hot spots and employs spatial data embedding to enhance accuracy. |
| Esquivel [45] | Unsatisfactory | Unsatisfactory | Good | Good | The authors introduced a novel Convolutional Long Short-Term Memory Neural Network (CLSTM-NN) for predicting crime events in Baltimore, USA, focusing on larceny and street robbery. Utilizing matrices of past crime events based on spatial coordinates, the CLSTM-NN model aims to leverage spatial and temporal correlations to forecast future crimes. |
| Wawrzynak [46] | Fair | Fair | Acceptable | Good | The authors presented a deep learning framework for crime forecasting, using ANNs optimized by Gram-Schmidt orthogonalization and a virtual leave-one-out test for input selection and neural architecture refinement, respectively. It combines spatiotemporal analysis, CNNs, and LSTM to enhance short-term crime prediction accuracy, focusing on minimizing mean square errors with unstructured police records. |

*2.1.2.1.2  Residual Network-Based Spatiotemporal Classification*

A Spatiotemporal Residual Network (STRN) model can be used to analyze spatiotemporal crime data by considering both the spatial and temporal features of the data. It processes data through a series of convolutional layers to extract spatial features and then passes it through a series of recurrent layers to extract temporal features.

*The rationale behind the usage of the technique*: (1) ResNet-based spatiotemporal models excel in managing the complexity of crime data, which encompasses both spatial and temporal dimensions. They adeptly learn spatial hierarchies and temporal sequences, essential for predicting crime occurrences, (2) Through skip connections, ResNets enhance gradient flow during training, mitigating the vanishing gradient issue. This enables the learning of complex patterns in crime data more accurately and efficiently, (3) ResNet-based models are well-suited for diverse and large crime datasets, crucial for real-world applications across varied data types and volumes, and (4) Combining ResNets with spatiotemporal modeling significantly improves crime prediction and classification, helping identify crime hotspots and trends.

*Conditions for the optimal performance of the technique*: (1) Customize ResNet architecture to data complexity and tasks, using deeper layers and residual connections to handle complex patterns and avoid gradient problems, (2) Hyperparameter optimization, including learning rate, layers, and batch size, is key, using methods like grid, random search, or Bayesian optimization, (3) Apply regularization and dropout to prevent overfitting, crucial for limited data scenarios, (4) Enhance temporal prediction accuracy by incorporating LSTM or GRU layers with ResNet, and (5) Use GIS data overlays and spatial clustering to improve spatial pattern analysis in crime data.

*Limitations of the technique:* (1) The complex nature of ResNets makes them difficult to interpret, posing challenges for trust and ethical considerations in crime prediction, (2) Accurately capturing temporal dynamics is difficult due to unpredictable factors affecting crime patterns, (3) Performance can be impacted by the level of spatial detail, with high granularity leading to sparse data and low granularity missing local patterns, (4) Models might not perform well in different regions due to varying local factors, limiting their widespread applicability, (5) Utilizing deep learning for crime prediction raises issues around bias, privacy, and community stigmatization, and (6) The evolving nature of crime, driven by socio-economic and environmental changes, challenges the adaptability of static models.

**TABLE 7:** FEATURING RSEARCH PAPERS **(PP.)** THAT HAVE EMPLOYED **RESIDUAL NETWORK-BASED SPATIOTEMPORAL TECHNIQUES** AND THEIR EVALUATION IN TERMS OF SCALABILITY **(SC.)**, INTERPRETABILITY **(IN.)**, ACCURACY **(AC.)**, AND EFFICIENCY **(EF.)**

| Pp. | Sc. | In. | Ac. | Ef. | Description |
|---|---|---|---|---|---|
| Materke [47] | Unsatisfactory | Acceptable | Good | Fair | The authors compared the performance of three Spatiotemporal deep learning algorithms for crime prediction. They compared Spatio Temporal Residual Network (ST-ResNet), Deep Multi View Spatio Temporal Network (DMVST-Net), and Spatio Temporal Dynamic Network (STD-Net)—for crime prediction, using the Chicago crime dataset. |
| Khoei [48] | Fair | Fair | Acceptable | Acceptable | The authors introduced a deep learning approach using a 50-layer Residual Neural Network (ResNet-50) for detecting and classifying Denial of Service (DoS) attacks on smart grid intrusion systems. By converting tabular data into images, the model enhances performance. ResNet-50's effectiveness was evaluated using accuracy and detection probability. |
| Li and Huang [49] | Good | Acceptable | Good | Acceptable | The authors proposed ST-HSL, a framework that integrates a multi-view spatial-temporal convolution network and hypergraph structure learning to encode local and global crime dependencies, refining crime predictions by capturing spatial-temporal patterns and high-order relationships. ST-HSL leverages self-supervised learning and contrastive learning to improve model accuracy. It incorporates residual connections to strengthen the framework's neural network architecture, ensuring comprehensive crime pattern representations through element-wise addition over previous embeddings. |

*2.1.2.1.3  LSTM Graph Attention-Based Spatiotemporal Classification*

The LSTM Graph Attention (GAT) architecture is a sophisticated neural network model designed for predicting crime by leveraging both temporal and spatial data dependencies. This innovative approach integrates LSTM networks, which excel in capturing temporal dependencies within sequential data, with Graph Attention mechanisms that focus on spatial dependencies. The latter is achieved by assigning attention weights to nodes within a graph, reflecting their importance to a specific task based on the spatial relationships and relevance. To implement this model for crime prediction, one begins by constructing a graph from crime data, where nodes symbolize geographical locations and edges depict the spatial relationships among these locations. This graph representation allows the LSTM GAT network to model crime patterns effectively over time and space. The training process involves feeding the network with historical crime data, enabling it to learn and identify underlying patterns and correlations. The LSTM GAT model leverages LSTM networks and Graph Attention mechanisms to predict crime by understanding temporal and spatial data relationships. This fusion allows for the representation of crime patterns over time and space, offering law enforcement critical insights for strategic crime prediction and potentially improving public safety.

*The rationale behind the usage of the technique*: The technique proves to be valuable due to its ability to capture temporal dependencies in crime patterns, consider spatial dependencies between neighboring areas, and effectively incorporate multiple types of input data, leading to more accurate predictions. Crime patterns have temporal dependencies that LSTM captures to understand them over time. Spatial dependencies also exist, where crimes in neighboring areas affect the likelihood of crime in a location. LSTM GAT integrates multiple types of input data.

*Conditions for the optimal performance of the technique*: To enhance the technique, consider using relevant features that capture spatial and temporal relationships between crime events, appropriate hyperparameters (e.g., learning rate, number of epochs, batch size, number of hidden layers), an optimal graph structure that captures spatial and temporal dependencies, and regularization techniques like dropout, early stopping, and weight decay to prevent overfitting by imposing constraints on parameters, thereby improving the accuracy of the LSTM GAT model.

*Limitations of the technique*:  The limitations are limited data availability, model complexity, and generalizability. If historical data is scarce, accuracy may be affected. The model requires significant computational resources due to its complexity. Also, it may not generalize well to new locations or changes in crime patterns over time, as it relies on historical crime patterns in a specific location to make predictions and may not adapt to changes in underlying patterns. The performance of the LSTM GAT model may be constrained by limited data availability, making it difficult to train the model effectively and obtain accurate predictions.

TABLE 8: RSEARCH PAPERS (PP.) THAT HAVE EMPLOYED LSTM GRAPH ATTENTION-BASED SPATIOTEMPORAL TECHNIQUES AND THEIR EVALUATION IN TERMS OF SCALABILITY (SC.), INTERPRETABILITY (IN.), ACCURACY (AC.), AND EFFICIENCY (EF.)

| Pp. | Sc. | In. | Ac. | Ef. | Description |
|---|---|---|---|---|---|
| Rayhan [50] | Unsatisfactory | Good | Good | Fair | The authors introduced a technique for forecasting criminal activity using customized feature embeddings based on the type of crime and geographic location. These embeddings are utilized by three sparse LSTM graph attention models capture daily and weekly patterns of crime. The method ends with a location-based attention mechanism to assess patterns and predict. |
| Manju [51] | Fair | Unsatisfactory | Acceptable | Unsatisfactory | The authors discussed enhancing surveillance video analysis by integrating 3D CNN (3DCNN) with LSTM and Bidirectional LSTM to recognize human actions and predict abnormal events. It captures motion information from sequential frames and utilizing temporal patterns, highlighting the importance of learning spatial and temporal structures for early action prediction |

2.1.2.2 Recursive Neural Network Spatiotemporal Classification

*2.1.2.2.1 LSTM Neural Network Spatiotemporal Classification*

To train a Spatiotemporal CNN for crime prediction, the process begins by gathering comprehensive data on crime incidents along with associated spatiotemporal features. These features might include population density, local events, and the timing of incidents, all crucial for understanding the context of crime occurrences. This data must then be aggregated and encoded into a format suitable for processing by the neural network. The technique not only leverages the spatial and temporal aspects of data through the CNN but also incorporates the strengths of recurrent neural networks (RNNs) to uncover underlying connections within the sequential dataset. RNNs, with their dynamic memory allocation and flexible architectures such as LSTM) and Gated Recurrent Units (GRU), excel in capturing the order of occurrences in predictive tasks. This combination allows for managing input sequences of varying lengths and is particularly adept at maintaining the continuity of data through time, which is essential for accurate crime prediction. After preparing the data, a Spatiotemporal CNN with integrated RNN features is designed to process sequential data. The model is trained using backpropagation and stochastic gradient descent to minimize errors. Once trained, it predicts crime by analyzing new data, providing insights for prevention and improving safety.

*The rationale behind the usage of the technique*: This technique integrates an extra module into a neural network, which is crucial for improving a model's capacity to grasp grammatical relationships. This module can identify the relevant information at a future stage in the sequence and also detect the point at which it becomes insignificant. The attention mechanism allows the model to focus on relevant parts of the sequence.

*Conditions for the optimal performance of the technique*: The method can be improved by conducting supervised training on a group of training sequences, employing an optimization algorithm like gradient descent, and using backpropagation through time to compute gradients. This allows the model to converge on the optimal set of parameters that minimize the difference between its predicted and expected outputs.

*Limitations of the technique*: The technique needs additional training data to attain proficiency. If the input data is not structured in a sequential format, it may have limited efficacy in the classification, rendering it inadequate for certain data types. The structure of input data impacts the efficacy of the model. Certain data types require specialized preprocessing. For instance, text data may require tokenization or stemming.

TABLE 9: FEATURING RSEARCH PAPERS (PP.) THAT HAVE EMPLOYED LSTM NEURAL NETWORK-BASED SPATIOTEMPORAL TECHNIQUES AND THEIR EVALUATION IN TERMS OF SCALABILITY (SC.), INTERPRETABILITY (IN.), ACCURACY (AC.), AND EFFICIENCY (EF.)

| Pp. | Sc. | In. | Ac. | Ef. | Description |
|---|---|---|---|---|---|
| Yi [52] | Unsatisfactory | Unsatisfactory | Fair | Good | The authors suggested utilizing a Continuous Conditional Random Field (CCRF) as the foundation for a Neural Network to predict crime. They introduced a LSTM component to capture the non-linear relationship between the inputs and outputs of each region, and a modified Stacked Denoising AutoEncoder component to model the interactions between regions. |
| Yuan [53] | Unsatisfactory | Acceptable | Good | Acceptable | The authors proposed a Convolutional LSTM neural network model to develop a technique for forecasting traffic accidents. To overcome the issue of spatial heterogeneity in the data, the researchers put forth a structure that includes spatial graph characteristics and a spatial model ensemble. |
| Mei [54] | Unsatisfactory | Unsatisfactory | Fair | Fair | The authors used LSTM models to analyze the predictability of commercial, pedestrian, and residential robberies, highlighting the impact of spatial and temporal scales on prediction accuracy. It showed variation in predictability among robbery types. It finds that increased spatial granularity reduces predictability and that each robbery type exhibits unique temporal patterns |

*2.1.2.2.2 Bidirectional LSTM Spatiotemporal Classification*

A Spatiotemporal Bidirectional LSTM (BiLSTM) is a type of neural network that uses both spatial and temporal information to predict future events.

The input data consists of features such as crime location, time, type, demographic data, and socio-economic factors. The network is trained on a historical crime dataset, where the input data is spatiotemporal features, and the output data is whether a crime occurred or not. The bidirectional aspect of the LSTM allows it to process the input data in both forward and backward directions, which helps to capture both past and future information and improve prediction accuracy.

*The rationale behind the usage of the technique*: (1) BiLSTM networks are tailored for recognizing long-term trends in crime patterns, leveraging their ability to understand sequential data's temporal dependencies, (2) By processing data from both past and future, BiLSTMs gain comprehensive insights, enhancing prediction accuracy by utilizing information from dual temporal contexts, (3) Integrating BiLSTMs with spatial analysis improves modeling of how crime varies across locations and times, capturing the essence of criminal activities' spatiotemporal nature, (4) BiLSTMs' proficiency in predicting future crimes and classifying crime types enables identification of potential hotspots and crime natures, aiding in resource allocation and targeted interventions, and (5) The architecture of BiLSTMs allows for the modeling of complex and non-linear interactions within crime data, addressing the multifaceted influences on crime rates.

*Conditions for the optimal performance of the technique*: **(1)** The model's performance can vary significantly based on the granularity of the temporal data (e.g., hourly, daily). The choice should reflect the application's requirements and the nature of the crime patterns, (2) Similarly, the spatial granularity (e.g., street level, neighborhood level) affects model performance. Higher resolution can provide more detailed predictions but requires more data and computational power, (3) Choosing the right batch size and the number of epochs for training is critical. Smaller batch sizes can offer more frequent updates and potentially better generalization, while larger batches can speed up the training process, and (4) Integrating domain knowledge into the model, either through feature engineering or model architecture decisions, can improve its accuracy.

*Limitations of the technique*: (1) Bi-LSTM models' complexity and their ability to capture long-term dependencies can lead to overfitting if training data is not large or diverse enough, reducing generalizability to new data or regions, (2) Integrating spatial information into Bi-LSTMs is challenging due to complex spatial dynamics affecting crime patterns, necessitating modifications or combining with models like CNNs to capture spatial features, (3) The "black box" nature of Bi-LSTM models complicates understanding their decision-making process and identifying driving features, posing issues for applications requiring transparency, such as law enforcement and public policy, and (4) Without regular updates to reflect changes in urban environments, societal behaviors, or law enforcement practices, Bi-LSTM models' effectiveness may decline, necessitating ongoing data collection and model retraining.

**TABLE 10:** FEATURING RSEARCH PAPERS **(PP.)** THAT HAVE EMPLOYED **BIDIRECTIONAL LSTM -BASED SPATIOTEMPORAL TECHNIQUES** AND THEIR EVALUATION IN TERMS OF SCALABILITY **(SC.)**, INTERPRETABILITY **(IN.)**, ACCURACY **(AC.)**, AND EFFICIENCY **(EF.)**

| Pp. | Sc. | In. | Ac. | Ef. | Description |
|---|---|---|---|---|---|
| Siami [55] | Acceptable | Fair | Good | Good | The authors conducted a behavioral analysis and comparison of Bidirectional LSTMs (BiLSTMs) and LSTM models, with the aim of investigating the potential benefits of additional layers of training data in tuning the model parameters. The findings revealed that the use of BiLSTM-based modeling led to more accurate predictions compared to traditional LSTM models. |
| Ren [56] | Unsatisfactory | Unsatisfactory | Good | Acceptable | The authors proposed an attentional based BiLSTM model to predict Malicious URLs. Initially, the URLs underwent preprocessing and were transformed into word vectors through pre-trained Word2Vec. Subsequently, a BiLSTM model along with an attention mechanism was trained to obtain URL sequences features and categorize them. |
| Freitas [57] | Fair | Unsatisfactory | Acceptable | Acceptable | The authors proposed a deep learning model to predict crimes in criminal sub-trajectories. The model combines two factors, location and time, and uses a BiLSTM to learn the underlying criminal activities from the sub-trajectory data of offenders. The BiLSTM operates at the embedding levels of location and time. |
| Onan [58] | Fair | Acceptable | Good | Acceptable | The author introduced SRL-ACO, a novel text augmentation method combining Semantic Role Labeling (SRL) and Ant Colony Optimization (ACO) to automatically produce extra training data for enhancing NLP model accuracy, which is demonstrated across multiple text classification tasks. It employs a pre-trained SRL model from, utilizing BiLSTM and self-attention layers for identifying semantic roles. |
| Onan [59] | Fair | Fair | Good | Good | The author introduced a convolutional RNN architecture that leverages bidirectional LSTM and GRU layers for analyzing sentiment, focusing on understanding both past and future contexts. This model applies a group-wise enhancement mechanism to prioritize significant features, alongside convolution and pooling layers to distill high-level features and condense the feature space. It showcased the effectiveness of bidirectional LSTM within this architecture. |
| Onan [60] | Good | Fair | Good | Good | The author introduced a novel three-layer stacked bidirectional LSTM framework for detecting sarcasm in text, incorporating a unique inverse gravity moment-based term-weighted word embedding with trigram to maintain word order. It represents the first comprehensive effort to evaluate both unsupervised and supervised term-weighted neural language models, including word2vec, fastText, and GloVe, for sarcasm identification. |

*2.1.2.2.3 Recurrent Convolutional Neural Network*

The Spatiotemporal Recurrent Convolutional Neural Network (SRCNN) integrates the strengths of CNNs and RNNs to effectively grasp both spatial and temporal dependencies within input data. This advanced neural network architecture excels in predicting crime in urban settings by leveraging historical crime statistics, temporal factors such as time of day, and demographic information, thus determining the potential for crime in specific locales and times. Its capacity to adapt to newly encountered data and identify unusual patterns renders it particularly valuable in dynamic scenarios where crime trends evolve. SRCNN continuously refines its internal representations with fresh data, enabling it to pinpoint areas or intervals of anomalously high crime rates, which could indicate a necessity for heightened surveillance measures.

The SRCNN merges CNNs and RNNs to analyze spatial and temporal data intricacies, including urban crime prediction using historical crime data, demographics, and time. It dynamically updates its learning from new data to detect changing patterns and anomalous behaviors, enhancing surveillance needs. SRCNN also deeply understands textual data's relational patterns, especially syntax, through iterative input weighting for sequential, structured forecasts. This ability to process data in order and adapt to new information makes SRCNN effective in applications requiring detailed pattern analysis and forecasting.

*The rationale behind the usage of the technique*: By using Recursive Neural Networks, the quadratic links connecting hidden child nodes in LSTM enable the capture of compositional semantics by linking hidden vectors across child units. This serves to reinforce the co-occurring properties of compositional semantics. Thus, LSTM can accurately capture the meaning of complex expressions.

*Conditions for the optimal performance of the technique*: To enhance the accuracy of the technique, one can merge numerical and symbolic data within a model architecture, and tackle information patterns of varying sizes and topologies during the learning phase. These data patterns encapsulate both the information and logical connections present in the data. By combining numerical and symbolic data, the strengths of both information are leveraged.

*Limitations of the technique*: While utilizing the technique for data extraction, there exists a chance of extracting data from the parameters themselves, which entails acquiring enough information to entirely reconstruct the models. Training GANs can be computationally intensive and prone to issues such as mode collapse or vanishing gradients, requiring careful hyperparameter tuning.

**TABLE 11**: FEATURING RSEARCH PAPERS **(PP.)** THAT HAVE EMPLOYED **RECURRENT CNN-BASED SPATIOTEMPORAL TECHNIQUES** AND THEIR EVALUATION IN TERMS OF SCALABILITY **(SC.)**, INTERPRETABILITY **(IN.)**, ACCURACY **(AC.)**, AND EFFICIENCY **(EF.)**

| Pp. | Sc. | In. | Ac. | Ef. | Description |
|---|---|---|---|---|---|
| Pareja [61] | Fair | Unsatisfactory | Good | Good | The authors presented a technique for predicting crimes that adjusts the graph convolutional network (GCN) model over time without relying on node embeddings. The approach involves utilizing an RNN to update the GCN parameters, capturing the graph sequence's dynamism. The authors examined two different architectures for this parameter evolution process. |
| Stec [62] | Acceptable | Unsatisfactory | Fair | Acceptable | The authors introduced a technique for predicting crimes that utilizes RNN. This approach integrates crime data with other information such as public transportation and census data to offer more comprehensive insights. The authors argue that crime trends are influenced by a range of environmental, economic, and sociological factors, and incorporating this data can provide a nuanced understanding of raw crime data. |
| Kabir [63] | Unsatisfactory | Fair | Fair | Fair | The authors put forward an RNN-based model that incorporates convolutional layers and a recurrent network. This model is designed to extract features from specific facial regions in images. They combined CNN for facial feature extraction and RNN to analyze temporal dependencies, achieving unparalleled accuracy in identifying human abnormalities. |
| Huang [64] | Fair | Fair | Acceptable | Fair | The authors created a crime prediction model using a deep neural network that can detect changing crime patterns and analyze the relationships between crime and data. The model incorporates spatial, temporal, and categorical signals into hidden representation vectors, which are then analyzed using attention-based RNN to capture the dynamic nature of crime. |
| Onan [65] | Fair | Fair | Good | Good | The author employed RNN and various machine learning techniques to analyze over 154,000 instructor evaluation reviews for opinion mining. We used three text representation methods (TP, TF, TF-IDF) with traditional ML, and four word embedding techniques (word2vec, GloVe, fastText, LDA2vec) with deep learning models. Results reveal that RNN with an attention mechanism using GloVe embeddings, significantly outperformed traditional classifiers. |
| Wang [66] | Acceptable | Fair | Good | Acceptable | The authors introduced HAGEN, a Graph Convolutional Recurrent Network tailored for crime prediction, combining adaptive graph learning and Diffusion Convolution Gated Recurrent Units (DCGRU) to capture both spatial crime correlations and temporal dynamics. Utilizing a homophily-aware constraint, HAGEN optimizes regional graph learning to ensure neighboring nodes reflect similar crime patterns, enhancing the model's accuracy through homophily-based graph convolution. Empirical tests on real-world datasets confirm HAGEN's superiority in crime forecasting. |

2.1.2.3  Graph Attention-Based Spatiotemporal Classification

*2.1.2.3.1  Convolutional Graph Attention-Based Spatiotemporal Classification*

The Spatiotemporal Convolutional Graph Attention (ST-CGA) model integrates spatiotemporal CNNs with graph attention networks (GATs) to enhance crime prediction capabilities. It is tailored for analyzing spatiotemporal data, specifically crime incidents, by spatially and temporally mapping crime data onto graphs where incidents are nodes interconnected by edges. This model uses CNNs to parse spatial features and employs GATs to understand the dynamic relationships and interactions between different locations and times. Through bidirectional attention mechanisms, ST-CGA captures nuanced entity characteristics, including semantic meanings, syntactic structures, and contextual relevance, converting crime text into a comprehensible graph format. By training on historical crime data, ST-CGA discerns underlying patterns and trends, leveraging the synergistic power of CNNs and GATs to forecast future incidents effectively. This graph-based approach allows for a deeper extraction and analysis of complex relationships within the data, making ST-CGA a potent tool for predicting crimes with high accuracy.

*The rationale behind the usage of the technique*: The method's implementation of bi-directional attention facilitates the creation of mutual information among text components depicted as nodes in a graph, along with the measurement of their relationships. This allows the precise measurement of the relationships between nodes, leading to a highly informative representation. The method can capture the intricate dependencies and contextual nuances between text components, resulting in a rich representation that can be leveraged for downstream tasks.

*Conditions for the optimal performance of the technique*: Utilizing GCN in the technique enables information exchange among entity nodes within the graph and generate revised node representations. Also, incorporating a two-tier neural attention mechanism for sequential data enhances interpretability, leading to greater insights into the results. The integration of Graph GCN allows for iterative information propagation and aggregation among entity nodes, capturing higher-order dependencies and refining the node representations, which improves performance.

*Limitations of the technique*: The method faces difficulty where its effectiveness deteriorates as the length of text sequences grows. In particular, the technique might encounter obstacles in distinguishing nodes that receive information from multiple other nodes. This can result in a lack of precision and accuracy in the technique's performance. The challenge of distinguishing nodes receiving information from multiple sources in lengthy text sequences can lead to reduced precision and accuracy.

TABLE 12: FEATURING RSEARCH PAPERS (PP.) THAT HAVE EMPLOYED **RECURRENT CONVLUTIONAL GRAPH ATTENTION-BASED SPATIOTEMPORAL TECHNIQUES** AND THEIR EVALUATION IN TERMS OF SCALABILITY (SC.), INTERPRETABILITY (IN.), ACCURACY (AC.), AND EFFICIENCY (EF.)

| Pp. | Sc. | In. | Ac. | Ef. | Description |
|---|---|---|---|---|---|
| Xiao [67] | Fair | Good | Good | Acceptable | The authors proposed the Attentional Factorization Machine model that utilizes a Convolutional Graph Attention (CGA) network to understand the significance of each feature interaction through data for crime prediction. The model incorporates a neural attention network to assess the significance of each feature interaction directly from data. |
| Hu [68] | Acceptable | Unsatisfactory | Acceptable | Acceptable | The authors proposed a Dual-robust Enhanced Spatiotemporal Learning Network method for crime prediction. It employs a convolutional self-attention structure representation module to encode the underlying crime patterns at each time slot, where specific relevant patterns are given weight to encode the present pattern. Also, a convolutional structure is utilized to improve the sequence-level pattern from the representation of each time slot. |
| Yu [69] | Fair | Fair | Good | Good | The authors proposed a method for predicting criminal activity using a multi-layer perceptron neural network. The method involves utilizing a CGA network to encode the second-order interactions among characteristics. The model combines multi-layer perceptual neural networks with pairwise ranking factorization to enhance item recommendation accuracy. |
| Mascorro [70] | Unsatisfactory | Acceptable | Acceptable | Fair | The authors utilized a CGA network to forecast criminal activity, incorporating a structure composed of four 3D convolutional layers and two fully connected layers. It identifies suspicious behavior indicative of shoplifting before the crime occurs. It focuses on analyzing pre-crime conduct by extracting behavioral features from video footage. |
| Onan [71] | Fair | Acceptable | Good | Good | The authors introduced GTR-GA, an approach that leverages graph-based neural networks and genetic algorithms to enhance text data augmentation, utilizing a model, HetGAPN, for capturing complex text relationships through graph attention and node aggregation techniques. This method generates diverse and high-quality augmented text, showing significant improvements in NLP tasks like sentiment analysis and text classification, and addressing data scarcity challenges. |
| Zhao [72] | Fair | Good | Good | Acceptable | The authors developed an end-to-end framework, AttenCrime, incorporating Graph Attention Network (GAT) and Temporal Convolution Network to effectively capture dynamic spatiotemporal features and fuse multiple domain data through a Cross-Domain Attention Network for precise crime forecasting. It demonstrated superior performance in urban crime prediction. |

### 2.1.2.3.2 Deep Neural Network-Based Spatiotemporal Classification

Spatiotemporal Deep Neural Networks (ST-DNNs) have emerged as a sophisticated neural network model, increasingly utilized for crime prediction tasks in urban areas due to their capability to process both spatial and temporal data efficiently. To effectively train an ST-DNN, it is essential to preprocess data using advanced techniques such as CNNs, RNNs, LSTM networks, and transformers. These methodologies empower the model to discern and interpret complex spatial and temporal patterns within the data.

The training of ST-DNNs integrates supervised and unsupervised learning, including clustering and anomaly detection, with adjustments in weights and biases. This comprehensive strategy not only boosts the accuracy of crime predictions but also ensures effective generalization of new data, employing methods like cross-validation and regularization. Once refined, ST-DNNs become crucial for law enforcement, offering precise predictions for better resource and personnel deployment in high-risk urban areas, thereby enhancing safety and security.

*The rationale behind the usage of the technique*: The advantage of using deep neural networks for crime prediction is that they can continuously learn and adapt as new data becomes available. This means that as new crime data becomes available, the model can update its predictions in real-time, ensuring that it remains accurate and effective. This ability to adapt and evolve makes deep neural networks a powerful tool for crime prediction.

*Conditions for the optimal performance of the technique*: To improve the technique, use (1) relevant data, (2) appropriate DNN architecture, and (3) hyperparameter tuning. The DNN needs to be trained on a diverse dataset that accurately represents crime patterns and demographics. A DNN that is too small may not learn underlying patterns, while one that is too large may overfit data. Hyperparameter tuning includes learning rate and regularization.

*Limitations of the technique*: The use of deep neural networks for crime prediction is not without its challenges. For example, there is a risk of bias in the data used to train the model, which could result in inaccurate or discriminatory predictions. Also, there are concerns around privacy and surveillance, particularly if the model is used to identify individuals who may be at risk of committing a crime.

**TABLE 13:** FEATURING RSEARCH PAPERS **(PP.)** THAT HAVE EMPLOYED **DEEP NEURAL NETWORK-BASED SPATIOTEMPORAL TECHNIQUES** AND THEIR EVALUATION IN TERMS OF SCALABILITY **(SC.)**, INTERPRETABILITY **(IN.)**, ACCURACY **(AC.)**, AND EFFICIENCY **(EF.)**

| Pp. | Sc. | In. | Ac. | Ef. | Description |
|---|---|---|---|---|---|
| Lin [73] | Unsatisfactory | Acceptable | Good | Good | The authors utilized both geographical features and deep neural network (DNN) techniques to develop a crime prediction model. This model incorporates feature learning techniques and offers an impartial benchmark for comparison. It facilitates the replication, dissemination, and iterative enhancement of the knowledge model. |
| Chun [74] | Unsatisfactory | Acceptable | Acceptable | Good | The authors utilized a DNN to predict criminal activity. Their predictions are based on the arrest booking records' historical data. The study provides access to the datasets used for crime prediction by researchers and analyzes prominent approaches applied in ML algorithms to predict crime, offering insights into different trends and factors related to criminal activities. |
| Guo [75] | Unsatisfactory | Fair | Fair | Good | The authors proposed a DNN-based method for crime prediction. Their method employs a DNN to model high-order feature interactions and can be trained end-to-end without requiring any feature engineering. The model does not need any pre-training. It learns both high- and low-order feature interactions. |

## 2.2 Classification Based on Traditional Methods

### 2.2.1 Spatial-Based Classification

#### 2.2.1.1 Adaptive Ensemble-Based Spatial Classification

##### 2.2.1.1.1 Gradient Boosting Machine (GBM)-Based Classification

Gradient Boosting Machine (SGBM) is a machine learning algorithm that leverages the principles of Gradient Boosting Machines to enhance crime prevention strategies by predicting the probability of crime in specific locations. This model synthesizes vast amounts of historical data to identify complex patterns, crucial for formulating effective crime prevention measures. The core strategy involves building a predictive model through the aggregation of multiple weak models, each trained on a subset of the data. These weak models are subsequently amalgamated using a boosting algorithm, culminating in a robust model capable of making precise predictions.

One innovative aspect of the SGBM approach is its method for handling continuous features. The algorithm discretizes these features into bins and selects the leaf node that most significantly reduces loss. This selection process is critical for optimizing the model's performance. The influence of gradient instances on information gain varies according to their magnitude, with larger instances exerting greater influence. This differential impact ensures that the model accurately reflects the importance of various data points, thereby enhancing its predictive accuracy.

*The rationale behind the usage of the technique*: The technique's vertical growth of decision trees using a leaf-wise split approach can improve classification accuracy. Also, incorporating Exclusive Feature Bundling and Gradient-Based One-Side Sampling can increase algorithm accuracy. Randomly dropping instances with small gradients while keeping instances with significant gradients helps to maintain accuracy. Vertical growth and exclusive feature bundling improves accuracy.

*Conditions for the optimal performance of the technique*: The technique uses a log loss function to assess its classification efficacy. Incorporating a personalized version of this function with the technique can augment its classification accuracy. By tailoring the log loss function, greater command over the learning process can be attained, facilitating the attainment of optimal model performance. This personalized version of the log loss function allows for fine-tuning the process to better align with the requirements.

*Limitations of the technique*: Overfitting is a potential issue with decision trees, where the tree becomes too deep and complex, making it difficult to interpret and apply to new data. This can happen when the tree is split at the leaf level, resulting in overly complex trees that capture the idiosyncrasies in the training data. These trees may not be effective at predicting new observations.

**TABLE 14:** FEATURING RSEARCH PAPERS **(PP.)** THAT HAVE EMPLOYED **GRADIENT BOOSTING MACHINE-BASED SPATIAL TECHNIQUES** AND THEIR EVALUATION IN TERMS OF SCALABILITY **(SC.)**, INTERPRETABILITY **(IN.)**, ACCURACY **(AC.)**, AND EFFICIENCY **(EF.)**

| Pp. | Sc. | In. | Ac. | Ef. | Description |
|---|---|---|---|---|---|
| Vomfell [76] | Unsatisfactory | Acceptable | Good | Acceptable | The authors predicted property crimes using GBM and data from taxi trips, Twitter, and Foursquare, computing a weighted average of base model forecasts. They demonstrated that incorporating new features related to human activity enhances the prediction of crimes influenced by local conditions, moving beyond reliance on past crime data and census information. |
| Inglvch [77] | Fair | Unsatisfactory | Acceptable | Fair | The authors compared crime prediction using GBM, logistic regression, and linear regression, finding gradient boosting most accurate. This research aimed to explore the use of statistical tools for predicting crime rates in urban areas, employing clustering techniques to identify spatial crime patterns and uncover influencing factors. |
| Asad [78] | Unsatisfactory | Unsatisfactory | Fair | Acceptable | The authors used machine learning to predict malware infections based on computer characteristics, using supervised learning and GBM. They explored the efficacy of LightGBM, Decision Tree Classifier, and Neural Networks through four experiments. They demonstrated the LightGBM's superiority as a gradient boosting algorithm for malware prediction. |
| Yu [79] | Good | Acceptable | Fair | Fair | The authors identified Spatiotemporal patterns in crime clusters and used them to construct a global crime pattern for future forecasting. The method selects relevant local spatio-temporal patterns for constructing a global crime pattern, aimed at enhancing future crime prediction. It offers timely forecasts with minimal data inputs. |

*2.2.1.1.2 Adaptive Boosting-Based Spatial Classification*

It is based on Adaptive Boosting (AdaBoost) that combines multiple weak learners to create a strong predictor. It works by first dividing the city or region into smaller spatial units, such as neighborhoods or blocks. Then, it analyzes the crime data for each spatial unit over a certain period, such as a week or a month. It uses AdaBoost to combine multiple weak predictors, each of which is based on a different set of spatial and temporal features, to create a strong predictor for each spatial unit.

*The rationale behind the usage of the technique*: (1) AdaBoost is an ensemble method that improves prediction by combining weak classifiers into a strong one, adjusting to focus more on incorrectly predicted data points, (2) This technique utilizes geographical patterns to enhance crime prediction accuracy, considering factors like socioeconomic conditions and urban design, (3) AdaBoost and spatial analysis together improve crime type classification by learning from location-based patterns and characteristics, (4) The combination of AdaBoost and spatial analysis adjusts to changing crime patterns, making it suitable for various contexts and dynamic environments, and (5) Accurate crime prediction allows for better resource allocation by law enforcement, focusing on high-risk areas to prevent crime.

*Conditions for the optimal performance of the technique*: (1) The success of AdaBoost depends on having a large, accurate, relevant, and complete dataset, particularly detailed historical crime data with times, locations, types, and other influencing factors, (2) AdaBoost's effectiveness varies with parameter settings, including the number and depth of weak learners and the learning rate. Optimal parameters are found through experimentation and cross-validation, (3) Accounting for varying crime patterns across different locations and times is essential, potentially requiring specialized models for spatial data and time series analysis, and (4) Enhancing the model with insights from criminologists, law enforcement, and urban planners can inform feature selection, model interpretation, and the application of predictive insights.

*Limitations of the technique*: (1) AdaBoost is highly sensitive to noise and outliers, which can cause overfitting in messy crime data, affecting performance on new data, (2) It overlooks spatial autocorrelation in crime data, potentially reducing prediction accuracy, (3) The complexity of AdaBoost increases with spatial data, leading to longer model training and prediction times, (4) Models may not generalize well to new areas, a significant issue for spatial crime prediction, (5) The complexity of AdaBoost's multiple weak learners makes it hard to interpret, limiting stakeholder understanding, and (6) It creates static models that may not adapt to changing crime patterns without retraining.

**TABLE 15:** FEATURING RSEARCH PAPERS (**PP.**) THAT HAVE EMPLOYED **ADAPTIVE BOOSTING-BASED SPATIALL TECHNIQUES** AND THEIR EVALUATION IN TERMS OF SCALABILITY (**SC.**), INTERPRETABILITY (**IN.**), ACCURACY (**AC.**), AND EFFICIENCY (**EF.**)

| Pp. | Sc. | In. | Ac. | Ef. | Description |
|---|---|---|---|---|---|
| Hossain [80] | Fair | Fair | Acceptable | Good | The authors proposed an algorithm for predicting crime that used supervised learning techniques and both spatial and temporal crime data. They analyzed historical crime data to discern patterns. They trained their model using decision tree and K-Nearest neighbor algorithms, and improved accuracy with Adaptive Boosting algorithms and Random Forest. |
| Yuki [81] | Fair | Unsatisfactory | Acceptable | Unsatisfactory | The authors utilized Adaptive Boosting to determine the likelihood of specific types of crime occurring at particular times and locations in Chicago. They employed five algorithms to forecast crime types based on time and location, finding that tree-based algorithms closely match actual outcomes, offering high accuracy with various tree classifiers. |
| Al-Ghush.[82] | Fair | Good | Fair | Good | The authors assessed various machine learning techniques, including Adaptive Boosting, for predicting crime types and identifying crime categories. They used temporal and spatial data from crime records to pinpoint crime hotspots and predict crime types. The study revealed that gradient boosting, random forest, and decision tree outperformed other models. |
| Onan [83] | Acceptable | Fair | Good | Good | The author assessed 66,000 MOOC reviews through machine and deep learning, focusing on ensemble learning techniques like AdaBoost, term weighting schemes, and word embeddings, using models ranging from Naïve Bayes to LSTM. It contrasts the efficiency of machine learning, involving TP, TF, TF-IDF, and supervised learning, with deep learning's use of word2vec, fastText, and GloVe in various architectures. |

2.2.1.2   Naïve Bays Estimation Neighbor-Based Spatial Classification

Spatial Naive Bayes (SNB) represents a specialized adaptation of the Naive Bayes algorithm, explicitly designed to incorporate the spatial relationships between data points. This variant is particularly effective in fields like crime prediction, where it models the interplay between the occurrence of crime and its environmental context. SNB enhances traditional Naive Bayes by integrating spatial neighbor estimation, a method that identifies and incorporates neighboring locations of a crime incident as predictors within the model. Such neighbors can be determined through various methodologies, including fixed radius or distance-based approaches, thereby enriching the model's ability to evaluate the likelihood of crime occurrences in new areas. SNB enhances crime prediction by incorporating spatial relationships, using a probabilistic classifier to assign risk categories to locations. It identifies high-risk areas by analyzing the environment around crime scenes with data not seen in training, guiding targeted prevention strategies.

*The rationale behind the usage of the technique*: Naive Bayes is a swift algorithm for training because it only requires computing the probability of each class and the probability of each class for distinct input values. This is due to the assumption that all features are independent, which contributes to the algorithm's speed. It requires fewer parameters.

*Conditions for the optimal performance of the technique*: The method can be improved by assigning feature-specific weights to each class. In this approach, the conditional probabilities of the text classifier are computed by utilizing feature-weighted frequencies derived from the training data. It can capture the relationships between text features and their corresponding categories.

*Limitations of the technique*: Any categorical variable in the test dataset that is not present in the training dataset would be assigned a probability of zero by Naive Bayes, making it impossible to generate any predictions. Also, the simplistic version of Naive Bayes does not support parallel computation.

**TABLE 16:** FEATURING RSEARCH PAPERS **(PP.)** THAT HAVE EMPLOYED **NAÏVE BAYS ESTIMATION NEIGHBOR-BASED SPATIALL TECHNIQUES** AND THEIR EVALUATION IN TERMS OF SCALABILITY **(SC.)**, INTERPRETABILITY **(IN.)**, ACCURACY **(AC.)**, AND EFFICIENCY **(EF.)**

| Pp. | Sc. | In. | Ac. | Ef. | Description |
|---|---|---|---|---|---|
| Kumar [84] | Fair | Acceptable | Acceptable | Fair | The authors assessed the use of data mining in identifying crime patterns, offering insights into the distribution and types of crimes in Cheltenham. They analyzed crime hotspots using big data and developed a Naive Bayes algorithm for crime identification and prediction. They sourced crime data from Cheltenham police data portal. |
| Jangra [85] | Unsatisfactory | Unsatisfactory | Fair | Good | The authors improved the precision of Naïve Bayes for crime prediction and found it to be more accurate than KNN. They found Naïve Bayes to surpass KNN in accuracy for crime prediction, despite crime prediction's complexity arising from numerous attributes. The study transitions from using KNN to adopting Naïve Bayes, highlighting its superiority. |
| Phua [86] | Unsatisfactory | Fair | Acceptable | Acceptable | The authors proposed a fraud detection approach that utilizes backpropagation and Naive Bayes on oversampled minority class data partitions. The findings indicate that with a constant decision threshold and cost matrix, employing a partitioning and multiple algorithms strategy slightly increases cost savings compared to altering the class distributions across the dataset |

2.2.1.3   Decision Support-Based Spatial Classification

2.2.1.3.1   Decision Tree-Based Spatial Classification

A Spatial Decision Tree (SDT) is a sophisticated, data-driven approach designed for predicting crime by analyzing the spatial patterns of criminal activity within specific areas. This decision-making method employs a tree-shaped model where the branches signify the outcomes of various tests, and the leaf nodes represent class labels, with each path from the root to a terminal point demonstrating a classification rule. SDT analyzes spatial data and crime patterns by classifying areas based on socio-demographic and environmental factors like poverty, population density, and land use. It divides the study area into smaller sections to examine crime presence and selects relevant predictors for each, adjusting to local area characteristics. The model uses a tree-shaped structure, with branches and leaves representing test results and outcomes, respectively. Decisions are made through a series of questions, leading to predictions about crime patterns.

*The rationale behind the usage of the technique*: Employing distinct subsets of features and decision rules during different classification phases leads to the effective handling of both categorical and continuous data and the handling of both linear and nonlinear relationships between input features and the output variable. This requires minimal data preparation. The cost of using the resulting tree is logarithmic in relation to the number of training data points.

*Conditions for the optimal performance of the technique*: Random Forests can enhance the method by constructing multiple decision trees from bootstrapped training samples and randomly selecting a subset of predictors. Another way to enhance the method is through boosting, which grows trees in a sequential manner while utilizing information from previously grown trees. By combining the predictions of multiple decision trees, boosting can create a more accurate model.

*Limitations of the technique*: The method is susceptible to instability. Even slight alterations in data can result in significant changes to the optimal decision tree's structure. When working with data that contains variables with varying levels, information gain tends to favor attributes with greater levels because they provide more discriminatory power in separating the data into different classes. Computations for uncertain data can be complex.

**TABLE 17:** FEATURING RSEARCH PAPERS **(PP.)** THAT HAVE EMPLOYED **DECISION TREE-BASED SPATIALL TECHNIQUES** AND THEIR EVALUATION IN TERMS OF SCALABILITY **(SC.)**, INTERPRETABILITY **(IN.)**, ACCURACY **(AC.)**, AND EFFICIENCY **(EF.)**

| Pp. | Sc. | In. | Ac. | Ef. | Description |
|---|---|---|---|---|---|
| Ivan [87] | Good | Unsatisfactory | Good | Fair | The authors focused on creating a prototype model for crime prediction using a decision tree (DT) algorithm. They chose this approach to predict crime. Using the WEKA Toolkit, we trained a J48 classifier on a preprocessed crime dataset, achieving a prediction accuracy of 94.25287%. This performance makes it a reliable option for crime prediction. |
| Kiani [88] | Fair | Unsatisfactory | Acceptable | Good | The authors created a machine learning algorithm to predict the type of crime in a specific area by analyzing past incident data. They used DT and Naive Bayes techniques. This method aims to generate superior training and testing datasets, remove insignificant attributes to tackle high-dimensional data, and optimize outlier parameters, demonstrating its efficacy. |
| Aldossari [89] | Fair | Acceptable | Fair | Good | The authors built a machine learning model to predict possible crime categories in a location by utilizing DT and Naive Bayes. This study focused on predicting crime categories in Chicago from 2013 to 2017, using Naïve Bayes and DT techniques. The DT classifier outperformed Naïve Bayes with a prediction accuracy of 91.59%, compared to 83.40% for Naïve Bayes. |

2.2.1.3.2   Boosted Decision Tree-Based Spatial Classification

The Spatial Boosted Decision Tree (SBDT) is an advanced machine learning algorithm that enhances prediction accuracy for crime patterns by integrating decision trees with spatial analysis. Utilizing geographical coordinates and other location-specific data, SBDT identifies relationships between various variables to forecast crime occurrences in distinct areas. It's an ensemble learning algorithm that improves prediction through the combination of multiple decision trees, each tree aiming to enhance the accuracy of its predecessors. SBDT combines spatial analysis with decision trees to predict crime patterns, focusing on the relationship between crime rates and factors like proximity to public transportation and commercial areas. It trains decision trees in sequence, each improving on the previous, to form a weighted model based on accuracy. This ensemble approach minimizes overfitting and improves predictions on new data, making SBDT a powerful tool for analyzing crime patterns through spatial and environmental data.

*The rationale behind the usage of the technique*: Since there may be many different factors that contribute to the likelihood of a crime occurrence, such as location, time of day, weather conditions, and demographic characteristics of the population, employing this technique is useful due to its ability to handle: (1) high-dimensional feature spaces, where there are many variables that may be relevant to the prediction task, and (2) non-linear relationships between the features and the target variable. In addition, this technique's flexibility allows it to adapt to changing patterns and dynamics in crime data, making it a valuable tool for law enforcement agencies in their crime prevention efforts.

*Conditions for the optimal performance of the technique:* The technique can be improved by: (1) selecting features based on domain knowledge, exploratory data analysis, and statistical methods), (2) balancing data with an equal number of positive and negative examples, to prevent it from being biased towards the majority class, (3) using appropriate boosting algorithm (the choice of the boosting algorithm should be based on the specific problem being solved), and (4) hyperparameter tuning. Incorporating ensemble methods, such as bagging or stacking, can enhance predictive performance by combining multiple models trained on different subsets of the data.

*Limitations of the technique*: The method suffers limitation in: (1) handling high-dimensional data, (2) capturing complex relationships, (3) requiring high quality data, and (4) accounting for temporal patterns. Boosted Decision Trees may not perform well if there are interactions between variables not captured by the model, or if there are missing or inconsistent data. Also, it may not be able to capture temporal patterns in crime data, which can lead to suboptimal predictions. The interpretability of the model may be compromised, as boosted decision trees tend to create complex and highly nonlinear models that are difficult to interpret and explain.

TABLE 18: FEATURING RSEARCH PAPERS (PP.) THAT HAVE EMPLOYED **BOOSTED DECISION TREE-BASED SPATIALL TECHNIQUES** AND THEIR EVALUATION IN TERMS OF SCALABILITY (SC.), INTERPRETABILITY (IN.), ACCURACY (AC.), AND EFFICIENCY (EF.)

| Pp. | Sc. | In. | Ac. | Ef. | Description |
|---|---|---|---|---|---|
| Kim [90] | Fair | Fair | Acceptable | Acceptable | The authors aimed to develop a model for accurately predicting crime using two classification algorithms, KNN and boosted decision tree, on a Vancouver Police Department (VPD) crime dataset from 2003 to 2018, containing over 560,000 records. Their approach involved data collection, classification, pattern identification, prediction, and visualization. |
| Hu [91] | Fair | Fair | Good | Fair | The authors proposed a crime prediction technique that utilizes various data sources. The study's unique aspect is the application of Kernel Density Estimation (KDE) and zoning district dataset. They combined distance and temporal decay effects to map crime distribution pattern. They utilized KDE with "generalized product kernels" and likelihood cross-validation |
| Alsirha [92] | Acceptable | Unsatisfactory | Unsatisfactory | Acceptable | The authors introduced a framework that combines Apache Spark, a distributed processing engine, with the Gradient Boosted Trees (GBT) classification algorithm to efficiently classify traffic data within a reasonable processing time. |

2.2.1.4   Distribution of Datapoints-Based Spatial Classification

*2.2.1.4.1   K-nearest Neighbor-Based Spatial Classification*

The algorithm employs k-nearest neighbors (k-NN) to predict the likelihood of new crime incidents by using spatial information to identify the closest historical data points. It classifies new incidents based on the proximity and frequency of crime classes among these neighbors, assigning the new incident to the most common class found. This method is particularly effective for complex decision boundaries or undefined data distributions, as it relies on the spatial features and historical context of neighboring incidents. The training and validation of the k-NN model allow for accurate predictions of crime occurrences. This approach leverages the concept of proximity for data classification, ensuring that each new data point is analyzed within the context of its nearest instances to produce a reliable class membership prediction.

*The rationale behind the usage of the technique*: When items are near each other within a dataset, they tend to exhibit similar characteristics or properties. A classification algorithm that considers the distances between data points can leverage the inherent relationships and similarities in datasets. An algorithm that considers the proximity of items can make accurate predictions and identify patterns that can't be identified by other methods. By leveraging the proximity concept, this method can effectively handle various types of data.

*Conditions for the optimal performance of the technique:* To surmount the hurdle presented by noisy or irrelevant features that can potentially impede the precision of the k-NN algorithm, evolutionary algorithms can be employed to optimize feature scaling. Also, scaling the features based on their mutual information with the training classes can enhance accuracy. This is useful when dealing with high-dimensional data. By employing evolutionary algorithms to optimize feature scaling, the k-NN algorithm becomes robust and efficient in handling high-dimensional data.

*Limitations of the technique:* Inconsistent feature scaling and the presence of noisy or irrelevant features can undermine the effectiveness of the method. Also, determining the optimal value of *k* can prove to be a challenging task, and the technique can be computationally inefficient. Testing multiple values of *k* can further increase the computational cost. The k-NN method's performance can be affected by imbalanced datasets, as it tends to favor the majority class, making it less suitable for imbalanced classification problems.

TABLE 19: FEATURING RSEARCH PAPERS (PP.) THAT HAVE EMPLOYED K-NEAREST NEIGBOR-BASED SPATIALL TECHNIQUES AND THEIR EVALUATION IN TERMS OF SCALABILITY (SC.), INTERPRETABILITY (IN.), ACCURACY (AC.), AND EFFICIENCY (EF.)

| Pp. | Sc. | In. | Ac. | Ef. | Description |
|---|---|---|---|---|---|
| Kumar [93] | Unsatisfactory | Acceptable | Good | Unsatisfactory | The authors proposed a tool for predicting crimes and frauds in urban areas. The K-Nearest Neighbor (KNN) prediction method was utilized to anticipate and detect criminal activity and fraudulent behavior in urban areas, as well as to monitor crime levels. The use of the KNN system has shown promise for precise crime analysis. |
| Liaha [94] | Fair | Unsatisfactory | Acceptable | Good | The authors examined KNN classification to forecast potential types of crimes based on different conditions. It focuses on predicting crime frequency by LSOA code and antisocial behavior crimes. It analyzed previously unknown, unstructured data. It evaluates K-means, Naive Bayes, and Linear Regression to predict states with the highest crime rates. |
| Tayebi [95] | Unsatisfactory | Unsatisfactory | Fair | Fair | The authors proposed a density based KNN model for predicting crime, which is based on point pattern analysis. It estimates the probability of criminal activity happening at a specific location by examining past occurrences. They aim to pinpoint crime-prone areas, raising public awareness and aiding in safer route planning. |

*2.2.1.4.2 Kernel Density Estimation Neighbor Spatial Classification*

Kernel Density Estimation (KDE) is a method to estimate the probability density function of a random variable. Spatial KDE is a variation of this method that estimates the density of events in a spatial region. Spatial Kernel Density Estimation Neighbor (SKDEN) uses this approach to identify high crime density areas and their neighboring areas likely to have high crime density. This can help predict future crime locations. After identifying high crime density areas, the characteristics that make them more susceptible to crime can be analyzed. To identify neighboring high crime density areas, a spatial weights matrix can be used to measure proximity.

*The rationale behind the usage of the technique*: (1) KDE estimates the probability density function of crime occurrences, identifying hotspots for targeted law enforcement efforts, (2) Analyzing spatial data with KDE reveals patterns and underlying causes of crime based on socio-economic factors, enhancing insights, (3) KDE-based techniques classify and predict crime locations, supporting the development of predictive policing models for preemptive actions, (4) Versatile across various crime types and scales, KDE can be tailored for different data sets and geographical areas, and (5) KDE maps engage the community by visualizing crime data, fostering awareness, and encouraging collaborative efforts

*Conditions for the optimal performance of the technique*: (1) Balance detail with computational efficiency in KDE's spatial resolution to avoid overfitting or missing local patterns, (2) Crucial for determining KDE's smoothing level, with optimal bandwidth capturing significant trends while preserving local detail, (3) Enhances prediction accuracy by analyzing trends, seasonal variations, and time-based crime patterns, (4) Selecting an appropriate kernel function, like Gaussian or Epanechnikov, based on data characteristics and analysis goals, (5) Improves predictions by accounting for crime's influence by nearby locations, using measures like Moran's I, and (6) Classifies areas accurately by defining neighborhoods through spatial dependency and nearest neighbor techniques.

*Limitations of the technique:* (1) It doesn't account for how nearby locations affect crime occurrences, potentially misleading in clustered areas, (2) Estimations near boundaries may be biased since data outside the study area aren't considered, possibly underestimating crime density near edges, (3) KDE struggles with datasets that have many variables due to the curse of dimensionality, (4) KDE might not generalize well for predicting specific crime types or understanding underlying causes without additional data and analysis, and (5) It assumes data homogeneity across the study area, which may not hold for crime data, missing local nuances.

TABLE 20: FEATURING RSEARCH PAPERS (PP.) THAT HAVE EMPLOYED KERNEL DENSITY ESTIMATION NEIGHBR-BASED SPATIALL TECHNIQUES AND THEIR EVALUATION IN TERMS OF SCALABILITY (SC.), INTERPRETABILITY (IN.), ACCURACY (AC.), AND EFFICIENCY (EF.)

| Pp. | Sc. | In. | Ac. | Ef. | Description |
|---|---|---|---|---|---|
| Neto [96] | Fair | Acceptable | Good | Good | The authors introduced a Marching Squares KDE approach to produce precise and quick crime hotspot maps. In their research, they evaluated the effectiveness of the technique by comparing it to the conventional Kernel Density Estimation method. They showed that MSKDE surpasses traditional KDE by producing a less anomalous map within the same timeframe. |
| Yang [97] | Acceptable | Fair | Good | Fair | The authors utilized KDE to develop a crime prediction model that uses historical data. The model fits a two-dimensional spatial probability density function to past crime records. Using Twitter-specific language analysis & statistical topic modeling, |
| Wang [98] | Acceptable | Unsatisfactory | Acceptable | Fair | The authors introduced an initial exploration into predicting criminal incidents using Twitter data. They improved crime prediction performance compared to KDE by automatically identifying discussion topics in a major city. The method leverages automated semantic analysis and comprehension of Twitter posts in natural language |
| Al Boni [99] | Fair | Acceptable | Fair | Acceptable | The authors proposed a localized KDE approach optimized with an evolutionary algorithm for hotspot analysis. The localized kernel density estimation (LKDE) technique effectively utilizes dense datasets to generate sharp, precise convolution kernels. When dealing with sparse data, it adjusts by expanding the kernel to incorporate more data. |
| Gerbr [100] | Acceptable | Fair | Fair | Acceptable | The authors examined whether incorporating Twitter data could enhance the accuracy of crime prediction compared to the conventional method that relies on kernel density estimation. By integrating topics into a crime prediction model, they demonstrated that incorporating Twitter data enhances the prediction accuracy for 19 out of 25 types of crimes. |

### 2.2.2 Spatiotemporal-Based Classification

#### 2.2.2.1 Random Forest Estimation-Based Spatiotemporal Classification

The Spatiotemporal Random Forest (STRF) is an advanced machine learning algorithm designed for predicting spatiotemporal events, notably crime. This method is an enhancement of the traditional Random Forest technique, incorporating both spatial and temporal dimensions to accurately forecast high-risk areas and periods for criminal activities. By leveraging historical crime data, demographic information, and a variety of other relevant variables, STRF aims to pinpoint potential future crime hotspots.

STRF leverages Random Forest, blending Bagging and Decision Trees (DTs) to construct diverse trees via random feature selection and feature bagging, minimizing overfitting. This method enables handling of complex data, numerous features, and their interactions. With capabilities to manage missing data and outliers, STRF stands out for crime prediction, analyzing extensive datasets for informed future forecasts, highlighting its importance in spatiotemporal event forecasting.

*The rationale behind the usage of the technique*: Including feature bagging can significantly decrease the correlation between DTs, resulting in increased predictive accuracy. This avoids highly predictive features that can cause similar splits in trees. A highly predictive feature will only be selected for a limited number of trees, preventing over-reliance on one feature. By incorporating feature bagging in the Random Forest algorithm, the correlation between decision trees is reduced.

*Conditions for the optimal performance of the technique*: The technique can be enhanced by calculating the density coefficients of data points for each class in the training set separately. This is because noisy data points, which are situated farther away from the set center than other data points, exhibit lower density. By calculating density coefficients separately for each class in the training set, the technique can effectively identify and handle noisy data points, which exhibit lower density.

*Limitations of the technique*: The utilization of distance measurement can lead to suboptimal performance when dealing with complex datasets that have multiple scales, varying densities, or high dimensionality. Also, determining the cutoff distance can be challenging since the range of each attribute is often unknown. Complex datasets with varying scales, densities, and high dimensionality can hinder the performance of this technique.

TABLE 21: FEATURING RSEARCH PAPERS (PP.) THAT HAVE EMPLOYED **RANDOM FOREST ESTIMATION-BASED SPATIOTEMPORAL TECHNIQUES** AND THEIR EVALUATION IN TERMS OF SCALABILITY (SC.), INTERPRETABILITY (IN.), ACCURACY (AC.), AND EFFICIENCY (EF.)

| Pp. | Sc. | In. | Ac. | Ef. | Description |
|---|---|---|---|---|---|
| Safat [101] | Fair | Fair | Good | Acceptable | The authors improved crime prediction accuracy using ML algorithms on Chicago and Los Angeles crime data. Random Forest Estimation and KNN were the top performers for Chicago and Los Angeles, respectively. Deep learning with LSTM revealed significant crime count variations in Chicago compared to Los Angeles. |
| Alves [102] | Unsatisfactory | Unsatisfactory | Acceptable | Acceptable | The authors utilized a random forest algorithm to make crime predictions and measure the impact of urban indicators on instances of homicide. Using a random forest regressor, this study achieves up to 97% accuracy in crime prediction. The methodology ensures robustness against minor data variations and ranks urban indicators by their influence. |
| Arora [103] | Acceptable | Fair | Acceptable | Good | The authors proposed Random Forest to predict cybercrimes over social media. Data is classified in the user defined classes as per the training data set. They emphasized the need for law enforcement to prioritize cybercrime. The Random Forest algorithm, with over 80% efficiency and 0.80 precision, emerged as the most suitable choice. |
| Schelter [104] | Fair | Acceptable | Acceptable | Fair | The authors proposed a classification model utilizing an ensemble of randomized decision trees, designed to quickly process unlearning requests. Its implementation leverages vectorized operators to enhance decision tree learning. Its performance in training time and predictive accuracy is comparable to popular tree-based machine learning models like Random Forest. |
| Araujo [105] | Unsatisfactory | Unsatisfactory | Acceptable | Unsatisfactory | The authors proposed a method to enhance patrol planning by identifying potentially dangerous locations and times through a suite of ML algorithms. They introduced a crime hotspot detection approach using MLP, KNN, and Random Forest algorithms. Experimental evaluations highlighted the impact of spatial granularity on model performance. |

#### 2.2.2.2 Bootstrap Aggregation Spatiotemporal Classification

Spatiotemporal Boosted Aggregation (STBA) is a sophisticated machine learning technique that leverages ensemble learning to combine spatiotemporal data with boosting algorithms for crime prediction. This method is particularly effective because it integrates both the spatial and temporal dimensions of crime data, enabling the model to predict not only where but also when future crimes are likely to occur. By accounting for the locations and times of past crimes, STBA provides a comprehensive framework for understanding and anticipating crime patterns. STBA aggregates models via ensemble techniques like majority voting to reduce variance and boost accuracy. It selects random training data samples, with or without replacement, for unbiased variable selection. Once trained, the model applies its insights to predict crime accurately, analyzing data's spatial and temporal aspects. This makes STBA effective for identifying crime hotspots and timing, aiding in prevention and resource planning for law enforcement.

*The rationale behind the usage of the technique*: By utilizing ensemble learning, wherein multiple weak learners collaborate to surpass a single strong learner, variance can be reduced, and overfitting can be avoided during the modeling process. This is important in the presence of highly noisy datasets, as reducing variance can enhance accuracy by introducing variability. Thus, we can get a better sense of the true pattern that underlies the noise. Through the utilization of ensemble learning, which harnesses the collective strength of multiple weak learners, the technique effectively reduces variance and mitigates overfitting.

*Conditions for the optimal performance of the technique:* To enhance the learning performance of the technique, hyperparameters can be fine-tuned through a grid search approach. Also, considering the difficulty level of each instance during the bootstrap sampling process can improve the method. By estimating the difficulty level of each instance, a combination of bootstraps with varying levels of difficulty can be created. Optimizing learning performance includes employing a grid search to fine-tune hyperparameters and considering instance difficulty during bootstrap sampling.

*Limitations of the technique:* The interpretability of the method may be sacrificed, and it may display substantial bias if the correct procedure is not followed. The method can be computationally costly, which may discourage its application in certain situations. Also, it could slightly reduce the effectiveness of stable methods such as K-nearest neighbors, due to its reliance on feature engineering and potential overfitting, but it may still provide valuable insights and predictions in complex data analysis tasks.

TABLE 22: FEATURING RSEARCH PAPERS (PP.) THAT HAVE EMPLOYED **BOOTSTRAP AGREGATION-BASED SPATIOTEMPORAL TECHNIQUES** AND THEIR EVALUATION IN TERMS OF SCALABILITY (SC.), INTERPRETABILITY (IN.), ACCURACY (AC.), AND EFFICIENCY (EF.)

| Pp. | Sc. | In. | Ac. | Ef. | Description |
|---|---|---|---|---|---|
| Beitollahi [106] | Fair | Unsatisfactory | Good | Unsatisfactory | The authors introduced a ML-based technique to tackle Application-layer Distributed Denial of Service (App-DDoS) attacks. It combines a Radial Basis Function (RBF) neural network for classification and the Cuckoo Search Algorithm (CSA) for training, with the Genetic Algorithm guiding feature selection from the NSL-KDD dataset. |
| Sphamadla [107] | Unsatisfactory | Unsatisfactory | Acceptable | Good | The authors introduced a hybrid algorithm that combines the strengths of Random Forest (RF) and Extremely Randomized Trees (ERT), specifically their techniques of bootstrap aggregation and random feature selection, termed as ERTwB. This approach not only surpasses RF and ERT in prediction accuracy and computational efficiency but also emphasizes the necessity of incorporating bagging and randomization in constructing effective supervised prediction models. |

## 3 CRIME PREDICTION BASED ON CLUSTERING

### 3.1 Spatial-Based Clustering

#### 3.1.1 Partitioning-Based Clustering

3.1.1.1 Hierarchical Agglomerative-Based Spatial Clustering

The Spatial-Based Hierarchical Agglomerative (SBHA) approach is a sophisticated technique utilized for crime prediction, combining spatial analysis and clustering methodologies to enhance understanding and forecasting of crime patterns within a city or region. This method begins by subdividing the area of interest into smaller sub-regions or cells, typically employing a grid-based system, to meticulously analyze crime data. By examining the type, frequency, and timing of crimes in each sub-region, the approach seeks to uncover patterns and similarities that may not be immediately apparent.

The SBHA approach starts by considering each sub-region as a distinct cluster and iteratively merges the nearest clusters based on spatial proximity and crime characteristics, such as type, frequency, and timing. This process, governed by a hierarchical agglomerative clustering algorithm, continues until all areas merge into a single cluster, visualized by a tree-like structure indicating cluster mergers. The linkage criterion (e.g., single or average linkage) used in this process significantly impacts the cluster formation and interpretation, with different methods influencing the shape and structure of clusters. Utilizing SBHA, analysts can deeply analyze and predict crime patterns, enabling effective crime prevention strategies by identifying spatial and characteristic similarities of crimes, thus providing a comprehensive framework for combating urban crime.

*The rationale behind the usage of the technique:* Generating the hierarchical arrangement of clusters that can adjust to distinct refinement levels can overcome the challenge of ambiguous similarity between documents and the existence of many subjects or motifs in the documents. It allows clustering process to have multiple levels of refinement, making it beneficial when working with documents that have a broad range of topics. This hierarchical clustering approach facilitates a more nuanced exploration and analysis of diverse document collections, enabling researchers to uncover hidden relationships and gain deeper insights into the data. By iteratively refining clusters, it provides a flexible framework for organizing document structures.

*Conditions for the optimal performance of the technique:* To optimize the efficacy of the method, it is crucial to choose a similarity metric that aligns with the attributes of the text data being analyzed, select a linkage method that is appropriate for consolidating clusters based on the specific data, and determine the number of clusters accurately to balance the need for refinement and comprehensibility (the number depends on the specific goals and context of the analysis). Consideration of the similarity metric, linkage method, and cluster count ensures that the hierarchical clustering process is tailored to the unique characteristics and objectives of the text analysis task.

*Limitations of the technique:* The computational complexity of the hierarchical clustering technique can pose challenges when working with large datasets, making it necessary to consider alternative methods for scalability. Also, the sensitivity to noise and anomalies in the data requires careful preprocessing and outlier detection techniques to mitigate their impact on the clustering results. Determining the optimal number of clusters remains a subjective and challenging aspect, often requiring domain expertise and iterative experimentation for satisfactory outcomes.

TABLE 23: FEATURING RSEARCH PAPERS (PP.) THAT HAVE EMPLOYED **HIERARCHICAL AGGLOMERATIVE-BASED SPATIAL TECHNIQUES** AND THEIR EVALUATION IN TERMS OF SCALABILITY (SC.), INTERPRETABILITY (IN.), ACCURACY (AC.), AND EFFICIENCY (EF.)

| Pp. | Sc. | In. | Ac. | Ef. | Description |
|---|---|---|---|---|---|
| Iqbal [108] | Good | Acceptable | Acceptable | Good | The authors proposed a framework to extract communities from chat logs, identify and condense relevant topics within those communities, and aid in crime detection. Employing WordNet for criminal information mining, it distinguishes cliques and their conversational topics within large suspicious chat logs, demonstrating superior accuracy in topic classification. |
| Boni [109] | Acceptable | Fair | Fair | Fair | The authors proposed crime prediction models using Hierarchical Agglomerative Clustering for specific areas. The models also focus on specific areas but overcome sparse data by sharing information among ZIP codes. This approach combines localized models addressing non-uniform crime patterns with shared information benefits. |
| Sivrjani [110] | Unsatisfactor | Unsatisfactor | Good | Fair | The authors grouped criminal activities using Hierarchical Agglomerative Clustering and utilized the KNN classification to predict crimes. They compared K-Means, Agglomerative, and DBSCAN clustering algorithms to identify the most effective method for crime detection, with K-Means results visualized on Google Maps for clarity. |
| Xu [111] | Acceptable | Fair | Good | Acceptable | The authors proposed a framework to efficiently detect crimes and acquire knowledge about criminal networks, employing automated network analysis and visualization in four stages: network creation, network partition, structural analysis, and network visualization. |

3.1.1.2 K-means-Based Spatial Clustering

Spatial K-means is a data mining technique widely employed for crime prediction, leveraging spatial information to uncover significant patterns within crime data. This method hinges on the K-means clustering algorithm, which groups data points based on their spatial proximity to identify crime hotspots and analyze crime patterns within a specified area. The process begins by assigning initial centroids randomly, followed by the recalibration of each cluster's centroid. This recalibration involves calculating the mean of all data points assigned to a cluster, thereby determining the cluster's new centroid based on these averages.

Subsequently, data points are reassigned to the nearest centroid, considering the updated centroid locations. This iterative procedure of updating centroid positions and reassigning data points continues until the centroids no longer change. Upon achieving this convergence, each data point is definitively allocated to one of the K clusters. Through this systematic optimization, the Spatial K-means method efficiently delineates K distinct clusters, enabling a robust analysis of crime patterns and facilitating the identification of areas requiring heightened attention or intervention.

*The rationale behind the usage of the technique*: K-means text mining enables analysts to rapidly and effectively cluster comparable documents based on their content, by randomly allocating each document to one of the K clusters and subsequently recalculating the centroid of each cluster, then reassigning each document to the cluster whose centroid is nearest to it. By grouping similar documents together, analysts can quickly identify themes and trends within a dataset, allowing for efficient data exploration and extraction of valuable insights. The K-means text mining approach simplifies the analysis process and enhances the scalability of document clustering.

*Conditions for the optimal performance of the technique*: To enhance the method, it is crucial to ensure clear separation between clusters (overlapping clusters can impede precise assignments), select an optimal *k* value (a value that is too small may result in broad clusters, while one that is too large can lead to overfitting), properly initialize centroids (initial placement can impact the sensitivity of the algorithm), and normalize the data appropriately (this prevents the domination of large-scale features). Also, implementing appropriate strategies such as clear cluster separation, optimal *k* value selection, proper centroid initialization, and data normalization is crucial to enhance accuracy.

*Limitations of the technique*: K-means text mining assumes that clusters are spherical and equally sized, which may not hold true in real-world scenarios, leading to suboptimal results. It also requires the user to predefine the number of clusters (K), which can be challenging when there is no prior knowledge or when the dataset lacks distinct clustering patterns. Lastly, K-means is sensitive to the initial placement of centroids, which can result in different outcomes depending on the initial configuration.

TABLE 24: FEATURING RSEARCH PAPERS (PP.) THAT HAVE EMPLOYED **K-MEANS-BASED SPATIAL TECHNIQUES** AND THEIR EVALUATION IN TERMS OF SCALABILITY (SC.), INTERPRETABILITY (IN.), ACCURACY (AC.), AND EFFICIENCY (EF.)

| Pp. | Sc. | In. | Ac. | Ef. | Description |
|---|---|---|---|---|---|
| Hjela [112] | Acceptable | Fair | Good | Acceptable | The authors proposed a crime prediction method based on spatial-based analysis and 2-Dimensional Hotspot analysis, utilizing K-Means clustering to identify clusters within the dataset and locate hotspot areas on a map. It uses a two-dimensional hotspot analysis, employing the KMeans algorithm for clustering, which then informs classification-based models for predicting crime |
| Tayal [113] | Unsatisfactory | Unsatisfactory | Good | Good | The authors employ k-means clustering method for crime detection. It visualizes results on Google Maps for better understanding and uses KNN classification for criminal identification and prediction. It generates two crime clusters through iterative grouping of crimes with similar attributes. Its effectiveness is confirmed by WEKA, showing over 93% accuracy. |
| Ch [114] | Unsatisfactory | Fair | Acceptable | Acceptable | The authors developed an algorithm using the K-Means method to efficiently identify local spatial patterns relevant to creating a global crime pattern from a training set. By analyzing contextual urban features like POI distributions, taxi mobility, and demographic data, they predict future road crime risks using a Zero-Inflated Negative Binomial Regression model. |
| Shriav [115] | Fair | Unsatisfactory | Fair | Fair | The authors explored the effectiveness of the Fuzzy Time Series method for crime forecasting, focusing on its applicability and the provision of practically. They introduced a technique for identifying criminal activities using K-Means clustering, labeling data points in a selected subset of the crime dataset based on a predetermined crime rate. |

### 3.1.2 Density-Based Spatial Clustering

By analyzing the spatial distribution of crimes, patterns and trends can be identified. One approach to analyzing spatial density is to use hotspot analysis, which identifies areas with statistically significant clusters of crime incidents. Another approach is to use predictive modeling to identify areas that are at high risk for future crime. This involves analyzing a variety of data sources, including historical crime data, demographic data, and environmental data, to identify factors that are correlated with crime. This information can then be used to create a spatial density map that highlights areas that are at high risk for future crime.

*The rationale behind the usage of the technique*: Data items that are close to each other and have high density are more likely to belong to the same cluster than data items that are far apart or have low density. Thus, the proximity and density of data points are critical factors that must be considered when deciding which data points should be grouped together into clusters. The method considers proximity and density as factors in determining cluster membership, emphasizing the importance of capturing local density patterns in the data. By incorporating these factors, the algorithm can effectively identify clusters of varying densities and shapes, adapting to the inherent structure of the dataset.

*Conditions for the optimal performance of the technique*: Improving the density-based clustering technique can be achieved by selecting a distance metric that aligns with the data characteristics, such as cosine similarity for textual data or Euclidean distance for numerical data. Determining appropriate density thresholds, such as the minimum number of points and minimum density, allows for effective identification of clusters while avoiding over or under-segmentation. Preprocessing text data through techniques like stop word removal, stemming, and lemmatization can enhance the quality and relevance of features, leading to more accurate and meaningful clustering results.

*Limitations of the technique*: The technique faces challenges with high-dimensional data due to the curse of dimensionality, leading to reduced accuracy and increased computational complexity. The method's performance is highly sensitive to parameter settings, requiring careful tuning for optimal results. Scalability concerns arise when dealing with large datasets, as the algorithm's efficiency may decrease as the data size increases. The technique is vulnerable to the influence of outliers, which can significantly impact the clustering results. The assumption of clusters being defined by areas of high density separated by low density regions restricts its ability to handle datasets with different density regions.

TABLE 25: FEATURING RSEARCH PAPERS **(PP.)** THAT HAVE EMPLOYED **DENSITY-BASED SPATIAL TECHNIQUES** AND THEIR EVALUATION IN TERMS OF SCALABILITY **(SC.)**, INTERPRETABILITY **(IN.)**, ACCURACY **(AC.)**, AND EFFICIENCY **(EF.)**

| Pp. | Sc. | In. | Ac. | Ef. | Description |
|---|---|---|---|---|---|
| Zhao [116] | Fair | Acceptable | Good | Fair | The authors presented a framework for predicting crimes that integrates Density-Based Spatial measurements with intra-region temporal correlation and inter-region spatial correlation. They emphasize the role of big data in collecting and integrating detailed urban, mobile, and public service data, offering richer temporal-spatial insights into crime patterns. |
| Huang [117] | Fair | Unsatisfactory | Acceptable | Unsatisfactory | The authors investigated the correlation between geographical and social features from location-based social networks and crime frequency in San Francisco, across various crime types. They employed the Density-Based Spatial algorithm to determine population density and crime rate by analyzing the number of check-ins and neighboring buildings. |
| Malathi [118] | Unsatisfactory | Fair | Fair | Fair | The authors proposed a hybrid clustering method that combines K-means and DBSCAN to form Density-Based Spatial clustering. This method utilizes density accessibility to measure the proximity to the δ-neighborhood. The authors outlined various clustering algorithms and enhanced these algorithms for future crime prediction. |
| Aryal [119] | Good | Unsatisfactory | Fair | Good | The authors applied a density-based clustering algorithm, named SparkSNN, on Spark to enhance the traditional SNN clustering algorithm's performance for analyzing big data on distributed computing clusters. They enhanced the conventional density-based clustering algorithm by using Spark and distributed computing clusters. |

### 3.2 Spatiotemporal-Based Clustering

### 3.2.1 Structure-Based Spatiotemporal Clustering

#### 3.2.1.1 Kernel Density-Based Spatiotemporal Clustering

A spatiotemporal-based kernel density for crime prediction involves using both the spatial and temporal dimensions to estimate the probability density function. This means that the kernel function used in the estimation process considers both the distance between locations and the time elapsed between events. Once the probability density function has been estimated, it can be used to predict the likelihood of a crime occurring at a given location and time in the future.

*The rationale behind the usage of the technique* (1) Kernel density estimation (KDE) allows for the visualization and analysis of crime patterns over a geographic area. By applying KDE, one can identify hotspots or areas with high concentrations of crime. This is crucial for law enforcement agencies to allocate resources more effectively and for policymakers to develop targeted interventions, (2) This technique combines spatial and temporal data to predict not just where crimes are likely to occur but also when. The integration of time into the analysis provides a more dynamic view of crime trends, enabling predictions about future crime occurrences based on past patterns. This is particularly useful for anticipating seasonal variations in crime rates or identifying times of day that are more prone to specific types of crime, and (3) By utilizing empirical data, the Kernel Density-Based Spatiotemporal technique supports a more objective and quantifiable approach to crime prediction and prevention. This data-driven method helps in moving away from intuition-based decisions towards evidence-based strategies, leading to potentially more effective and efficient law enforcement practices.

*Conditions for the optimal performance of the technique:* (1) Choose the right grid cell size or KDE bandwidth to balance detail and generalization in spatial analysis, (2) Select the appropriate time window for analysis based on crime types and occurrence patterns, (3) Consider external factors like socio-economic indicators, weather, and events to enhance prediction accuracy, (4) Handle crime data sensitively, protecting privacy and addressing biases, and (5) Combine KDE with other analytical techniques like machine learning for a more robust prediction and classification system.

*Limitations of the technique:* (1) Inaccurate or missing data can lead to biased results, (2) Low resolution or sparse data can limit insights, (3) Incorrect kernel or bandwidth selection affects accuracy, (4) Assumes stable crime patterns over time, which may not hold, (5) Descriptive nature doesn't explain why crimes occur, and (6) High data volume can be computationally expensive.

TABLE 26: FEATURING RSEARCH PAPERS (PP.) THAT HAVE EMPLOYED **KERNEL DENSITY-BASED SPATIOTEMPORAL TECHNIQUES** AND THEIR EVALUATION IN TERMS OF SCALABILITY (**SC.**), INTERPRETABILITY (**IN.**), ACCURACY (**AC.**), AND EFFICIENCY (**EF.**)

| Pp. | Sc. | In. | Ac. | Ef. | Description |
|---|---|---|---|---|---|
| Kim [120] | Unsatisfactory | Fair | Acceptable | Good | The authors utilized crime statistics data from the Seoul Metropolitan Police Agency to generate three types of maps: point-based, area-based, and density-based. They also investigated the selection of bandwidth for Kernel Density Estimation (KDE) used in creating density-based maps. |
| Hu et [121] | Fair | Unsatisfactory | Acceptable | Fair | The authors proposed a technique to predict criminal activities by integrating various data sources, such as Spatiotemporal crime dataset and zoning district dataset. They used KDE and zoning district dataset to overcome the challenges in crime prediction. |
| Clougty [122] | Unsatisfactory | Unsatisfactory | Fair | Acceptable | The authors investigated the spatial and temporal patterns of sexual assault occurrences in the University of Virginia from 1990 to 2015. They employed KDE, logistic regression, and random forest modeling techniques. Utilizing a logistic regression model revealed that after 1998, the distance to sex offender residences became significant predictors. |

3.2.1.2 Density-Based Spatiotemporal Clustering

Spatiotemporal density-based clustering is a technique used to identify clusters of data points in both spatial and temporal dimensions. It combines the concepts of spatial density-based clustering and temporal data analysis to discover patterns and structures in data that evolve over time. The basic idea is to identify regions in space and time where the density of data points is high, indicating the presence of a cluster. The density is typically defined based on distance and time thresholds. Data points that are close to each other in space and time are considered to belong to the same cluster.

*The rationale behind the usage of the technique:* (1) Crime data exhibits spatiotemporal patterns, which density-based techniques can detect, revealing clusters of criminal activities in specific locations and timeframes, (2) Density-based clustering helps identify spatial crime hotspots, aiding efficient resource allocation for law enforcement, (3) These methods consider temporal trends in crime rates, enabling the detection of patterns at different times, enhancing understanding, (4) Density-based clustering adapts to data variations, making it suitable for analyzing crime data with diverse clustering levels, (5) Identifying spatiotemporal crime patterns aids resource allocation, enhancing crime prevention and response strategies, (6) Spatiotemporal clustering aids predictive modeling by revealing historical crime patterns, improving forecasting accuracy.

*Conditions for the optimal performance of the technique:* (1) Choose an appropriate density-based clustering algorithm (e.g., DBSCAN or HDBSCAN) capable of handling varying cluster densities and noise, (2) Tune the density threshold parameter to control cluster size and shape based on data density, (3) Adjust temporal and spatial sensitivity parameters to determine data point proximity within clusters, (4) Implement noise handling mechanisms for noisy data points (e.g., filter or create separate clusters), (5) Ensure interpretability of clusters for actionable insights (use visualization techniques), and (6) Iteratively refine clustering by modifying parameters, features, or preprocessing for improved results.

*Limitations of the technique:* (1) DBSCAN and similar algorithms require careful parameter tuning, impacting cluster quality, (2) Large datasets pose computational and memory challenges, leading to longer processing times, (3) DBSCAN assumes dense, compact clusters, struggling with irregular shapes or varying densities, (4) Performance relies on data distribution; complex distributions may hinder cluster identification, (5) Clusters lack inherent labels, requiring manual interpretation, especially in complex contexts, and (6) DBSCAN focuses on grouping, requiring additional steps for predictive power.

TABLE 27: FEATURING RSEARCH PAPERS (PP.) THAT HAVE EMPLOYED **DENSITY-BASED SPATIOTEMPORAL TECHNIQUES** AND THEIR EVALUATION IN TERMS OF SCALABILITY (**SC.**), INTERPRETABILITY (**IN.**), ACCURACY (**AC.**), AND EFFICIENCY (**EF.**)

| Pp. | Sc. | In. | Ac. | Ef. | Description |
|---|---|---|---|---|---|
| Catlett [123] | Fair | Fair | Good | Acceptable | The authors proposed an approach for identifying high crime risk areas and predicting crime trends using spatial analysis. They created a Spatiotemporal crime forecasting model consisting of crime-dense regions with corresponding predictors, each estimating the number of crimes in the associated region. It used a density-based clustering algorithm that identifies clusters based on the estimated density distribution of the data. |
| Aryal [124] | Acceptable | Acceptable | Fair | Fair | The authors proposed a clustering algorithm for predicting crime by using a density-based Spatiotemporal approach. It utilizes geo-located data points and expands on the SNN clustering technique, incorporating location and time. This method can identify clusters of varying shapes, sizes, and densities. |

### 3.2.1.3 Self-Exciting Point Process-Based Spatiotemporal Clustering

The Self-exciting Point Process (SEPP), a technique rooted in spatial and temporal analysis, is utilized to identify patterns in crime occurrences based on their locations and times. This method is integral in pinpointing hotspots and analyzing crime patterns, which in turn facilitates the prediction of where and when future crimes are likely to emerge. The process begins with the collection of historical crime data to train the model. A suitable model, often a Hawkes process, is chosen to align with the specific characteristics of the data.

The SEPP method trains on historical crime data, utilizing maximum likelihood estimation or other statistical techniques to fine-tune parameters like the intensity function and self-excitation term. This enables precise spatial and temporal predictions of future crimes. Its effectiveness is gauged through precision, recall, and F1 score metrics, comparing forecasts to actual events. The outcomes of this evaluation facilitate model refinement, incrementally improving prediction accuracy for enhanced crime prevention strategies.

*The rationale behind the usage of the technique:* Since crime data is often characterized by clustering and periodicity, employing the self-exciting point process (SEPP) is advantageous due to its ability to: (1) capture the complex spatiotemporal patterns of crime, (2) be flexible (it can be customized to specific crime types and locations), and (3) be used for real-time crime prediction and resource allocation. Also, the SEPP model can assist law enforcement agencies in optimizing patrol routes, allocating personnel, and strategically deploying resources to the identified hotspots, ultimately aiding in crime prevention and mitigation efforts.

*Conditions for the optimal performance of the technique:* To improve the technique: (1) calibrate SEPP model to appropriate spatial and temporal resolution for the crime being predicted, (2) construct SEPP model based on appropriate assumptions about the underlying crime process, (3) validate model using appropriate techniques to assess accuracy, possibly by testing its ability to predict future crime patterns with historical data. Continuous monitoring and evaluation of the technique's performance, along with feedback from end-users and stakeholders, can drive iterative improvements and ensure its practical applicability in the ever-evolving field of crime prevention and public safety.

*Limitations of the technique:* Limitations: (1) limited spatial and temporal resolution as SEPP models assume crime events are independent and future events only depend on past events in the same location, (2) limited interpretability, (3) limited generalizability as patterns and factors contributing to crime vary across contexts, making it not generalizable, across different geographical areas or time periods. The reliance on historical data for training the model may introduce bias and fail to capture emerging crime trends or changes in the underlying dynamics, necessitating the need for continuous monitoring and adaptation of the technique to evolving crime patterns and contexts.

**TABLE 28:** FEATURING RSEARCH PAPERS **(PP.)** THAT HAVE EMPLOYED **SELF-EXCITING POINT PROCESS-BASED SPATIOTEMPORAL TECHNIQUES** AND THEIR EVALUATION IN TERMS OF SCALABILITY **(SC.)**, INTERPRETABILITY **(IN.)**, ACCURACY **(AC.)**, AND EFFICIENCY **(EF.)**

| Pp. | Sc. | In. | Ac. | Ef. | Description |
|---|---|---|---|---|---|
| Mohler [125] | Fair | Fair | Good | Acceptable | The authors investigated the use of self-exciting point processes in criminology. Utilizing residential burglary data from the Los Angeles Police Department, it showcases the application of these models to urban crime analysis. A nonparametric estimation method is used to examine the space-time triggering function and analyze temporal trends in burglary rates. |
| Jaiswal [126] | Fair | Unsatisfactory | Acceptable | Fair | The authors introduced an approach to forecast homicides in the Chicago crime dataset. Their method involved using an epidemic type of aftershock sequence model, which was based on a marked self-exciting point process, to enhance the prediction accuracy. |

### 3.2.2 Partitioning-Based Spatiotemporal Clustering

#### 3.2.2.1 Hierarchical Structure-Based Spatiotemporal Clustering

The approach integrates spatiotemporal hierarchical structuring with advanced clustering methods to efficiently analyze crime data across different regions. Initially, the study area is divided into hierarchical regions based on various factors such as population density, demographics, land use, and crime rates, with each region containing smaller sub-regions. This structure facilitates the use of spatiotemporal modeling techniques, including spatial and temporal autoregressive models, to predict crime rates in each sub-region by considering the spatial and temporal relationships between neighboring areas and their crime histories. The technique combines hierarchical and heuristic clustering to refine crime data analysis. Starting with hierarchical clustering for initial structure, it then applies heuristic clustering to merge data points based on proximity to the nearest centroid, continuing until a set convergence criterion is reached. This dual-method approach efficiently captures both broad and detailed data patterns, providing a robust framework for crime prediction that considers various spatial and temporal factors.

*The rationale behind the usage of the technique:* By combining hierarchical and heuristic clustering, the benefits of both methods can be achieved. Hierarchical clustering detects nested clusters and handles complex data structures but can be computationally expensive. Heuristic clustering is faster and manages large datasets but may not be as effective at detecting complex structures or nested clusters. By merging them, we leverage their strengths both, resulting in an approach that can handle complex data structures, detect nested clusters, and efficiently process large datasets. This combined technique provides a balance between accuracy and computational efficiency, making it a valuable tool for clustering tasks in various domains.

*Conditions for the optimal performance of the technique:* Selecting an appropriate stopping criterion is crucial for enhancing the technique. The stopping criterion determines when to terminate the algorithm after achieving the desired number of clusters. This criterion may rely on cluster similarity or variation within clusters. Proper configuration of clustering parameters is critical. Choosing an appropriate linkage criterion for merging clusters is essential for addressing the data, characteristics, and the desired clustering outcome. Common linkage criteria include single-linkage, complete-linkage, and average-linkage, each with their own strengths and weaknesses. Proper parameter tuning can ensure that the technique is effectively applied to address the challenges in the dataset.

*Limitations of the technique:* The sensitivity of the technique to the initial placement of data points or centroids can make it susceptible to getting stuck in local optima, potentially leading to suboptimal clustering results. Careful consideration and experimentation with different initialization strategies can help mitigate this issue. The oversimplification of data structure through lengthy chains of clusters can result in a loss of fine-grained details and hinder the ability to capture more intricate relationships and patterns within the data. Advanced techniques, such as density-based clustering or graph-based clustering, may be worth exploring to address these limitations and handle noisy datasets effectively.

TABLE 29: FEATURING RSEARCH PAPERS (PP.) THAT HAVE EMPLOYED **HIERARCHICAL STRUCTURE-BASED SPATIOTEMPORAL TECHNIQUES** AND THEIR EVALUATION IN TERMS OF SCALABILITY (SC.), INTERPRETABILITY (IN.), ACCURACY (AC.), AND EFFICIENCY (EF.)

| Pp. | Sc. | In. | Ac. | Ef. | Description |
|---|---|---|---|---|---|
| Wu [127] | Unsatisfactory | Fair | Good | Good | The authors proposed a hierarchical Spatiotemporal network model to predict crimes, which captures the shared interactions between regions, time, and categories in a multi-dimensional Spatiotemporal latent space. This model considers the inter-dependencies across time and space, which helps to address the issue of data imbalance. |
| Yu [128] | Fair | Fair | Acceptable | Unsatisfactory | The authors introduced a method for detecting hierarchical Spatiotemporal patterns and building a predictive model based on these patterns at varying resolutions. The method involves generating indicators from the raw data within different Spatiotemporal spaces. A distributed Spatiotemporal pattern (DSTP) is then extracted from a distribution, which is formed by multi-clustering locations. |
| Wang [129] | Unsatisfactory | Acceptable | Fair | Acceptable | The authors centered their attention on the spatiotemporal distribution of crime point patterns and the statistical methods used to analyze them. They utilized a hierarchical spatiotemporal algorithm to effectively depict and visualize crime clusters, which could be described using various levels of spatial clusters based on specific criteria. |

3.2.2.2 Hierarchical Density-Based Spatiotemporal Clustering

Hierarchical Density (HD) is a statistical method used for crime prediction, focusing on the density of crime incidents in both time and space dimensions. This approach employs a hierarchical clustering algorithm to group crime incidents into clusters, taking into account their proximity in time and space. Once these clusters are formed, the HD algorithm proceeds to calculate a density score for each cluster. This score reflects the number of crimes that have occurred within the cluster relative to its size, effectively measuring the concentration of crime within each cluster. The calculated density scores are pivotal in predicting the likelihood of future crime incidents occurring in the same area. By analyzing both the number of crimes and the size of the clusters, the HD algorithm enables the identification of high-density crime areas. These areas are deemed more prone to future incidents, thereby allowing law enforcement and public safety organizations to allocate resources more efficiently and implement preventative measures in areas most at risk.

*The rationale behind the usage of the technique*: By analyzing the patterns in the types of crimes committed within each cluster, HD can provide insights into the underlying factors or characteristics that contribute to the prevalence of specific crimes in certain areas. This knowledge can aid in understanding the root causes of crime and inform targeted interventions, such as community-based programs or environmental modifications, to mitigate the risk factors associated with particular crime types. The identification of crime hotspots and patterns through HD can support evidence-based decision-making, resource allocation, and proactive policing strategies, enhancing crime prevention efforts.

*Conditions for the optimal performance of the technique*: Incorporating advanced analytical techniques such as machine learning algorithms, neural networks, or deep learning models can enhance the predictive capabilities of the technique by capturing complex relationships and nonlinear patterns in the data. Regular model evaluation and validation using appropriate metrics and cross-validation techniques are crucial to assess the performance and reliability of the crime prediction system. Collaboration and knowledge sharing between researchers and law enforcement agencies can facilitate the continuous refinement of the technique, ensuring its effectiveness in addressing evolving crime dynamics.

*Limitations of the technique*: The assumption of spatial homogeneity in the technique may oversimplify the reality of crime patterns, as crime densities can vary significantly within a given region. The limited scope of the technique restricts its predictive capabilities to predefined regions, potentially overlooking emerging crime patterns in areas not covered by the predefined regions. Moreover, relying solely on past data may not account for changes in social, economic, or demographic factors that can influence crime patterns over time, requiring regular updates and adjustments to maintain accuracy. Considering additional contextual factors and incorporating dynamic data sources can enhance the predictive power.

TABLE 30: FEATURING RSEARCH PAPERS (PP.) THAT HAVE EMPLOYED **HIERARCHICAL DENSITY-BASED SPATIOTEMPORAL TECHNIQUES** AND THEIR EVALUATION IN TERMS OF SCALABILITY (SC.), INTERPRETABILITY (IN.), ACCURACY (AC.), AND EFFICIENCY (EF.)

| Pp. | Sc. | In. | Ac. | Ef. | Description |
|---|---|---|---|---|---|
| Butt [130] | Good | Fair | Good | Good | The authors put forward an economical and feasible approach to identify crime hotspots and forecast the frequency of criminal activities in those areas. The approach relies partly on utilizing Hierarchical Density-Based Spatial Clustering of Applications with Noise (HDBSCAN) to spot the hotspots with an elevated likelihood of crime incidents. |
| Sravni [131] | Acceptable | Fair | Good | Acceptable | The authors proposed a technique to forecast the category of crime that is likely to occur, utilizing engineered spatial features and identifying crime hotspots through a hierarchical density approach. The number of clusters in the detection process was optimized by considering the silhouette score. The approach aimed to predict the type of crime based on the given crime data. |
| Baqir [132] | Fair | Unsatisfactory | Acceptable | Fair | The authors [130] conducted a comparison of two hierarchical clustering algorithms for identifying crime hot-spots in urban regions: Hierarchical Density-based spatial clustering of applications with noise and Hierarchical Agglomerative Clustering (HAC). The results of the analysis indicated that HDBSCAN exhibited superior performance over HAC in terms of accuracy. |

## 4 CRIME PREDICTION BASED ON REGRESSION

### 4.1 Variable Independency-Based Regression

#### 4.1.1 Bayesian Logistic Regression

Bayesian logistic regression is a statistical technique that models the relationship between categorical predictor variables and a binary outcome variable, making it particularly useful in the field of crime investigation. It operates by specifying a prior distribution for model parameters, which is informed by past crime statistics or other relevant prior information. This prior distribution is then updated using Bayes' theorem, in conjunction with the likelihood function derived from the data. This approach allows for the incorporation of both prior knowledge and observed evidence into the estimation of model parameters, enhancing the accuracy of predictions. Specifically, in crime investigation, Bayesian logistic regression can model the probability of a crime being committed based on various factors such as location, time, and demographics, making it a powerful tool for predicting crime occurrences and understanding the dynamics behind them.

*The rationale behind the usage of the technique*: Since crime data is often characterized by clustering and periodicity and there may be many different factors that contribute to the likelihood of a crime occurring, the following characteristics of the Bayesian Logistic Regression is advantageous: incorporation of prior knowledge, flexible modeling, probabilistic inference, and handling missing data. The Bayesian Logistic Regression approach is advantageous in analyzing crime data due to its ability to handle the inherent clustering and periodicity patterns often observed. Its flexibility in modeling various factors influencing crime likelihood, along with its capability for probabilistic inference.

*Conditions for the optimal performance of the technique*: To improve the technique, it is crucial to ensure the use of accurate and comprehensive data, as the reliability of predictions heavily depends on the quality of input data. Selecting appropriate prior probabilities based on expert knowledge and the specific context of the crime investigation can enhance the model's ability to capture relevant patterns and make more reliable predictions. Also, choosing suitable models that align with the characteristics of the data and crime phenomena can lead to accurate results. Continuous updating with new data allows for adapting to evolving patterns over time. Employing proper regularization can help prevent overfitting and improve generalization.

*Limitations of the technique*: The technique has limitations. First, specifying priors can be challenging when there is limited information or a lack of informative prior knowledge, which may affect the accuracy of the results. As the number of predictor variables increases, the computational complexity of Bayesian logistic regression grows, potentially posing challenges for large-scale analyses. The reliability of results heavily relies on assumptions about the data, and if these assumptions are violated, the predictions may become less reliable. Bayesian logistic regression may not be suitable for situations with limited data, as the model relies on having enough data to estimate the parameters.

TABLE 31: FEATURING RSEARCH PAPERS (PP.) THAT HAVE EMPLOYED **BAYESIAN LOGISTIC REGRESSION-BASED TECHNIQUES** AND THEIR EVALUATION IN TERMS OF SCALABILITY (SC.), INTERPRETABILITY (IN.), ACCURACY (AC.), AND EFFICIENCY (EF.)

| Pp. | Sc. | In. | Ac. | Ef. | Description |
|---|---|---|---|---|---|
| Zhang [133] | Acceptable | Fair | Good | Acceptable | The authors Zhang [131] discussed machine learning and sampling approaches for detecting money laundering and rare events. They analyzed five algorithms, including binary Bayes logistic regression. |
| Tabezki [134] | Unsatisfactory | Fair | Good | Good | The authors used multiple techniques, including logistic regression, to determine the number of crimes, minimizing classification errors. |
| Chang [135] | Fair | Unsatisfactory | Acceptable | Acceptable | The authors [130] conducted a comparison of two hierarchical clustering algorithms for identifying crime hot-spots in urban regions: Hierarchical Density-based spatial clustering of applications with noise and Hierarchical Agglomerative Clustering (HAC). The results of the analysis indicated that HDBSCAN exhibited superior performance over HAC in terms of accuracy. |

#### 4.1.2 Linear Regression

Linear Regression involves creating a model that represents the correlation between a dependent variable (y) and one or multiple independent variables (x). The dependent variable, y, may be a textual variable, such as a sentiment score, while the independent variable, x, can consist of various characteristics of the text, such as word frequencies, character n-grams, or syntactic patterns " and "In the text-based numerical prediction technique, the initial step involves collecting relevant data. Subsequently, features are extracted from the text using methods such as Bag-of-Words (BoW) or Term Frequency-Inverse Document Frequency (TF-IDF). The dataset is then split into training and testing sets to train a linear regression model, followed by evaluating its performance. To improve the model's accuracy, hyperparameters can be adjusted, regularization techniques can be applied, and further feature extraction techniques can be employed to refine the model.

*The rationale behind the usage of the technique*: Linear regression modeling provides coefficient estimates for each predictor variable, indicating the direction and magnitude of their impact on the dependent variable. This enables the identification of significant keywords or phrases that have a strong association with the topic or sentiment under investigation. Moreover, the simplicity and interpretability of linear regression allow for intuitive insights and effective communication of the results to stakeholders. However, it's important to note that linear regression assumes a linear relationship between the predictor variables and the dependent variable, which may not always hold in complex text data.

*Conditions for the optimal performance of the technique:* To improve the technique, careful modification of hyperparameters is necessary, such as fine-tuning regularization parameters like L1 or L2 regularization, which can help control model complexity and prevent overfitting. Model performance can be assessed using metrics such as Mean Absolute Error (MAE) or Mean Squared Error (MSE) to gauge the accuracy of the predictions. To ensure reliable results, it is important to avoid including highly correlated features in the model as they can introduce multicollinearity issues. Scaling the features using techniques like normalization or standardization is beneficial to ensure uniform scaling and avoid potential biases. Adhering to the assumptions of linear regression, such as linearity, independence of errors, and homoscedasticity, is crucial for accurate modeling.

*Limitations of the technique:* It is crucial to consider the limitations of the linearity assumption in linear regression when dealing with nonlinear relationships between predictor and dependent variables. Approaches like TF-IDF may not fully capture the semantic relationships between words, limiting the model's understanding of context. High dimensionality, often encountered in data, can increase the risk of overfitting. Multicollinearity, caused by high correlation among predictor variables, can impact the interpretability of the model and introduce uncertainty in estimating the effects of correlated variables. Sparse data, where most entries are zero, can pose challenges in model training and convergence due to the limited information available.

TABLE 32: FEATURING RSEARCH PAPERS (PP.) THAT HAVE EMPLOYED LINEAR REGRESSION-BASED TECHNIQUES AND THEIR EVALUATION IN TERMS OF SCALABILITY (SC.), INTERPRETABILITY (IN.), ACCURACY (AC.), AND EFFICIENCY (EF.)

| Pp. | Sc. | In. | Ac. | Ef. | Description |
|---|---|---|---|---|---|
| Alotaibi [136] | Fair | Unsatisfactory | Good | Good | The authors investigated how normative beliefs and subjective norms impact cyber bullying and the expected social consequences. The collected data was analyzed using linear regressions. The study indicated that attitudes, social norms, perceived control over behavior, and the use of social media are all linked to the likelihood of engaging in cyberbullying. |
| Misyris [137] | Fair | Acceptable | Good | Acceptable | The authors investigated the potential link between the occurrence of criminal activities in Portland, USA. They utilized linear regression techniques to make precise forecasts. In their analysis of a dataset spanning 59 years to forecast the occurrence of crimes in Delhi-India. |
| Gera [138] | Unsatisfactory | Unsatisfactory | Acceptable | Fair | The authors employed a linear regression model to forecast a range of crimes in the city of Delhi over the next fifteen years, using data on various crimes collected over the past fifty-nine years. Forecasts were made for total cognizable crimes, murder, kidnapping, and abduction, dacoity, robbery, burglary, theft, and rioting. |
| McClendon [139] | Acceptable | Fair | Fair | Acceptable | The authors carried out a study in which they compared the patterns of violent crime using Communities and Crime Dataset, with actual crime statistics. They utilized Linear Regression, employing an identical finite set of features, on the Dataset. The linear regression algorithm has demonstrated significant effectiveness and accuracy in predicting crime statistics. |

## 4.2 Past Observations-Based Regression

### 4.2.1 Logistic Regression

The logistic regression model employs the logistic (or sigmoid) function to calculate the log-odds (logarithm of the odds) of the outcome, based on the values of the independent variables. This function, defined as $f(x) = 1 / (1 + e^{-x})$, where e is the base of the natural logarithm, smoothly maps any real-valued input to a probability value between 0 and 1. Consequently, this mapping enables the classification of data into two distinct classes by transforming the log-odds using the logistic function, thus obtaining the predicted probabilities. These probabilities represent the likelihood of each category or class. A decision threshold is typically applied to assign the input to the most probable category. This process effectively converts the input features into a probability output, facilitating the prediction and categorization tasks inherent in logistic regression analysis.

*The rationale behind the usage of the technique:* By utilizing a logistic function to link input features to a probability output, several benefits are obtained including the ability to easily understand and interpret the algorithm, effective handling of high-dimensional data, accommodation of non-linear relationships, an output that is easily interpretable, and the prevention of overfitting, leading to improved generalization performance. These advantages make the logistic function a valuable tool in binary classification tasks.

*Conditions for the optimal performance of the technique:* Employing techniques such as dimensionality reduction, such as Principal Component Analysis (PCA) or feature selection algorithms like Recursive Feature Elimination, can help reduce noise and improve efficiency. Data preprocessing steps, including normalization and scaling can contribute to the overall performance. Incorporating advanced optimization algorithms, such as stochastic gradient descent or Adam, helps expedite the model convergence and improve its training speed. Regular model evaluation and performance monitoring on unseen data are crucial to maintain accuracy over time.

*Limitations of the technique:* The assumption of linearity may not hold for complex cases where the relationship between input features and the target variable is nonlinear. Feature selection becomes challenging when dealing with many features, which can lead to overfitting. In imbalanced datasets, where one class is underrepresented, the technique may exhibit lower accuracy in classifying the minority class. Multicollinearity can pose a problem when independent variables exhibit high correlation, as it can impact the model's interpretability and stability.

TABLE 33: FEATURING RSEARCH PAPERS (PP.) THAT HAVE EMPLOYED **LOGISTIC REGRESSION-BASED TECHNIQUES** AND THEIR EVALUATION IN TERMS OF SCALABILITY (SC.), INTERPRETABILITY (IN.), ACCURACY (AC.), AND EFFICIENCY (EF.)

| Pp. | Sc. | In. | Ac. | Ef. | Description |
|---|---|---|---|---|---|
| Badal [140] | Acceptable | Good | Good | Fair | The authors implemented four predictive models—logistic regression, neural networks, decision trees, and random forests—leveraging police intelligence on known fraudulent companies to pinpoint others likely involved in money laundering. This approach aims to maximize the identification of fraudsters while minimizing the misclassification of legitimate businesses. |
| Bharati [141] | Unsatisfactory | Fair | Acceptable | Acceptable | The authors predicted the likelihood of future crimes based on conditions in a dataset of multiple crimes, using different models including logistic regression and KNN classification to determine the most effective model for crime prediction. Additionally, they visualize the data to identify patterns, such as peak times or months for criminal activity. |
| Mati.[142] | Fair | Acceptable | Fair | Fair | The authors used random forest and logistic regression to predict future crime rates using timing information from previous incidents and hot spot characteristics. The study identifies when and where assaults mainly happen through time series and hot spot analysis. It creates hourly assault prediction models based on land use with logistic regression. |
| Rumens [143] | Acceptable | Unsatisfactory | Fair | Good | The authors proposed a combined model that integrates logistic regression with neural network analysis to generate bi-weekly forecasts for the year 2014, drawing on crime statistics from the preceding three years. Additionally, the study produces monthly forecasts that differentiate between day and night, providing a finer temporal breakdown. |

### 4.2.2 Auto-Regressive Analysis

Auto-regressive regression is a statistical method that models the relationship between a dependent variable, such as crime rates in a particular area, and a combination of its past values and other independent variables, including demographics, economic conditions, and policing strategies. This technique is particularly useful in crime prediction, as it can uncover patterns and trends in criminal activity by analyzing the current crime rate in relation to its lagged values. The process of selecting the model's order is critical, balancing the need to capture sufficient historical information without overfitting. A higher order model may allow for the incorporation of more complex dynamics, but it also increases the risk of amplifying noise or relying too heavily on irrelevant past data. Thus, choosing the appropriate order for an auto-regressive model requires a careful mix of statistical techniques, domain knowledge, and empirical evaluation to ensure accuracy and relevance in predictions.

*The rationale behind the usage of the technique*: By capturing the temporal dependence of crime rates, auto-regressive modeling enables forecasting future crime rates based on historical data. The model considers the inherent patterns and correlations within the time series, allowing for a more accurate prediction of potential crime trends. The ability to anticipate future crime rates can provide valuable insights for policymakers, law enforcement agencies, and community organizations in implementing preventive measures and allocating resources effectively. However, it is important to note that while auto-regressive modeling is a powerful tool, other factors beyond historical crime rates should also be considered to comprehensively understand and address crime dynamics.

*Conditions for the optimal performance of the technique*: To enhance the technique, consider the following. Ensuring time series stationarity is crucial as non-stationary data can significantly impact pattern and trend identification. This can be achieved through techniques like differencing or transformation. Selecting an appropriate model such as AR (Autoregressive) or ARMA (Autoregressive Moving Average) based on the data's features is essential. Careful consideration should be given to the data's autocorrelation and seasonality. Precise parameter estimation is vital for accurate modeling. This involves using estimation techniques such as maximum likelihood estimation or Bayesian methods to optimize the model parameters. Updating the model with new data is important to maintain accuracy and account for any changes in the patterns.

*Limitations of the technique*: The technique assumes stationarity, which may not hold true for crime data that can exhibit non-stationary behavior due to various social, economic, and environmental factors. The technique relies primarily on historical data and may not adequately consider external factors such as policy changes, socio-political events, or shifts in law enforcement strategies. Therefore, it may fail to capture the full complexity of crime patterns and trends. The technique may struggle to detect and account for sudden changes in crime patterns, such as spikes or dips, which may require more advanced models or adaptive techniques. It requires expertise in time series analysis and statistical modeling, which can pose a challenge for practitioners without specialized knowledge.

TABLE 34: FEATURING RSEARCH PAPERS (PP.) THAT HAVE EMPLOYED **AUTO-REGRESSIVE ANALYSIS-BASED TECHNIQUES** AND THEIR EVALUATION IN TERMS OF SCALABILITY (SC.), INTERPRETABILITY (IN.), ACCURACY (AC.), AND EFFICIENCY (EF.)

| Pp. | Sc. | In. | Ac. | Ef. | Description |
|---|---|---|---|---|---|
| Payne [144] | Acceptable | Unsatisfactory | Good | Fair | The authors conducted a correlation analysis on the impact of pandemics on economic growth and found a robust connection between unemployment and crime. They also proposed an auto-regressive model for crime prediction, which was used to predict crime rates for the next six months in Queensland, Australia. |
| Dash [145] | Unsatisfactory | Acceptable | Acceptable | Acceptable | The authors used an auto-regressive model to forecast the number of crimes that may occur in a specific community during a particular month and year. This approach combines crime reports with information from schools, libraries, police stations, and 311 service requests using a network analysis method, improving the precision of our forecasts. |
| Yi [146] | Fair | Unsatisfactory | Good | Unsatisfactory | The authors developed a model that uses auto-regression for temporal correlation and spatial correlation across areas to predict crime, integrating these analyses to forecast future crime rates. The Clustered-CCRF model and tree-structured clustering algorithm effectively combine these correlations. They validated the approach using real crime data from Chicago. |

## 4.3 Time-Series Analysis-Based Regression

Time-series analysis-based regression is a potent method for predicting future crime trends. This approach involves studying historical crime data to uncover patterns over time, enabling the development of predictive models. Commonly employed time-series regression models include Autoregressive Integrated Moving Average (ARIMA), Seasonal ARIMA, as well as more advanced options like Vector Autoregression (VAR) and recurrent neural networks (RNNs) such as LSTM. The process begins with establishing a mathematical connection between a dependent variable (crime rates) and independent variables across time. To ensure the reliability of the model, preprocessing is essential, which includes assessing the stationarity of the time series data. Subsequently, relevant independent variables are carefully chosen for integration into the regression model. Once the variables are selected, an appropriate regression method is applied to fit the data, calculating the coefficients that describe the relationship between the variables. To gauge the model's performance and accuracy, various metrics are employed for evaluation. This comprehensive approach enables the development of robust crime prediction models that can inform law enforcement and policymakers effectively.

*The rationale behind the usage of the technique*: The technique integrates the following approaches, which provides an accurate comprehension of the data and gaining insights in the complex relationships between variables that vary over time: (1) time-series regression, which enables the analysis and modeling of variable relationships over time, and (2) text mining, which facilitates the extraction of insights from text data. By leveraging time-series regression, researchers can analyze the dynamic relationships between variables and capture temporal patterns and trends. Incorporating text mining allows for the extraction of valuable information from data, enabling deeper understanding and uncovering hidden insights within the text-based information.

*Conditions for the optimal performance of the technique*: The technique can be improved by: (1) gathering relevant and sufficient data, (2) collecting data of superior quality, (3) verifying and modifying the data to ensure stationarity, as needed, to satisfy the model's assumptions, (4) addressing autocorrelation, a typical feature in time-series data, (5) selecting appropriate independent variables through careful feature selection and considering their relevance to the dependent variable. Additionally, (6) employing advanced regression techniques, such as ARIMA or LSTM, tailored for time-series analysis can enhance the accuracy and predictive power of the model.

*Limitations of the technique*: The limitations include: (1) the complexity and diversity of language, which can make preprocessing textual data challenging, (2) the difficulty of converting this data into a useful format for analysis, (3) time-series regression models may become overly complex when dealing with high-dimensional data, and (4) the technique may struggle to capture non-linear relationships and interactions between variables, which can limit its ability to fully explain complex phenomena. (5) the technique assumes that the underlying relationships between variables remain stable over time, which may not always hold true in dynamic environments.

TABLE 35: FEATURING RSEARCH PAPERS (PP.) THAT HAVE EMPLOYED TIME-SERIES ANALYSIS-BASED TECHNIQUES AND THEIR EVALUATION IN TERMS OF SCALABILITY (SC.), INTERPRETABILITY (IN.), ACCURACY (AC.), AND EFFICIENCY (EF.)

| Pp. | Sc. | In. | Ac. | Ef. | Description |
|---|---|---|---|---|---|
| Garcia [147] | Unsatisfactory | Fair | Acceptable | Unsatisfactory | The authors introduced an approach to time series regression that allows the examination of spatiotemporal crime patterns at a street-level granularity. It consists of two primary components that address the challenges of the sparsity of spatial data and the spreading of crimes across urban areas. |
| Yuan [148] | Fair | Unsatisfactory | Fair | Unsatisfactory | The authors examined criminal networks with a time-based approach using dynamic network analysis. They analyzed the trends in the networks' variations and focused on monitoring changes in the overall structure of networks. They demonstrated the advantage of comprehending the entire network's structure to make it difficult for criminals to conceal their activities. |
| Jamson [149] | Unsatisfactory | Fair | Fair | Fair | The authors assessed the strength of correlations within data to uncover spatiotemporal patterns across various scales. They merged time-series analysis with outcomes derived from random matrix theory. They applied cross-correlation measures, eigenvalue spectrum analysis, and insights from random matrix theory. |

## 5 CRIME PREDICTION BASED ON SOCIAL NETWORK ANALYSIS

### 5.1 Path-Based Analysis

#### 5.1.1 Shortest Path-Based Analysis

Shortest path-based analysis is a technique employed in crime prediction to pinpoint high-risk areas for criminal activity. This method revolves around the identification of the shortest path connecting known crime locations and its application in forecasting potential future crime hotspots. The underlying principle hinges on the assumption that criminals tend to repeat offenses in familiar territories, such as places they have targeted before. By scrutinizing these shortest paths among known crime sites, we can discern areas most susceptible to future criminal activities.

To implement this technique, algorithms like Dijkstra's algorithm come into play. These algorithms calculate the shortest route between two nodes by taking into account the weight of the connecting edges. In this context, edge weight symbolizes the intensity of the relationship between two individuals, incorporating factors such as the duration of their acquaintance, shared experiences, and the level of trust between them. This integration of algorithmic analysis and relationship strength assessment enhances the accuracy of crime prediction and assists law enforcement agencies in proactive crime prevention strategies.

*The rationale behind the usage of the technique*: Determining the shortest path between two nodes is an efficient way for revealing information, resources, or influence to flow. This is useful for understanding social networks, identifying key players, and evaluating network efficiency and robustness. Nodes with high connectivity are influential and connect different parts of the network. Identifying alternative paths can assess redundancy and critical nodes for maintaining connectivity.

*Conditions for the optimal performance of the technique*: To enhance the technique: (1) choose a suitable algorithm for the specific network being analyzed, like Dijkstra and Floyd-Warshall algorithms, (2) choose appropriate node selection, (3) select appropriate optimization parameters. Analyzing the entire network may not be necessary. Instead, it may be useful to select representative nodes relevant to the analysis. The algorithm and node selection can impact the analysis performance.

*Limitations of the technique*: The limitations include: (1) ignoring the importance of weak ties, (2) failing to capture multiple pathways that may exist between nodes, (3) lack of consideration for the context in which relationships occur, (4) neglecting the role of intermediary nodes (it ignores the role of intermediary nodes or brokers that can facilitate the spread of information in a network), (5) assuming homogeneity of relationships (e.g., assuming all paths have the same weight).

TABLE 36: FEATURING RSEARCH PAPERS (PP.) THAT HAVE EMPLOYED **SHORTEST PATH-BASED ANALYSIS TECHNIQUES** AND THEIR EVALUATION IN TERMS OF SCALABILITY (SC.), INTERPRETABILITY (IN.), ACCURACY (AC.), AND EFFICIENCY (EF.)

| Pp. | Sc. | In. | Ac. | Ef. | Description |
|---|---|---|---|---|---|
| Wiil [150] | Good | Unsatisfactory | Acceptable | Fair | The authors introduced a fresh approach for examining the significance of connections within terrorist networks, drawing inspiration from studies on transportation networks. They integrated the link importance metric into CrimeFighter Assistant and assessed its effectiveness on established terrorist networks. By utilizing various methods such as co-occurrence analysis, the shortest path algorithm, and a heuristic approach. |
| Chau [151] | Acceptable | Unsatisfactory | Fair | Fair | The authors were able to enhance the accuracy and efficiency of a link analysis system. The incorporation of these techniques effectively enabled the identification of connections in criminal networks. Subjects in the experiments preferred heuristic-based paths over simple co-occurrence analysis and felt an automated system would greatly benefit crime investigations. |
| Xu [152] | Fair | Unsatisfactory | Acceptable | Fair | The authors were able to enhance the accuracy and efficiency of a link analysis system by utilizing various methods such as co-occurrence analysis, the shortest path algorithm, and a heuristic approach, The incorporation of these techniques effectively enabled the identification of connections in criminal networks. |

### 5.1.2 Minimum Spanning Tree-Based Analysis

A Minimum Spanning Tree (MST) algorithm plays a pivotal role in crime prediction, assisting in the identification of critical locations within a city or neighborhood where crime prevention efforts should be concentrated. This technique is grounded in graph theory and aims to determine the smallest set of edges required to connect all nodes in a graph, while minimizing the total edge weights. The resultant MST serves as a valuable tool for pinpointing potential crime hotspots and guiding the allocation of resources for effective crime prevention.

To construct this essential tree structure spanning all graph vertices with the least possible total edge weight, MST algorithms come into play. Among these algorithms, Kruskal's algorithm takes the approach of sorting edges by weight and adding them in increasing order. Prim's algorithm, on the other hand, begins with a single vertex and progressively incorporates the minimum-weight edges to unvisited vertices. This iterative process continues until all vertices have been visited, ultimately building the Minimum Spanning Tree. These algorithms, when applied judiciously, empower law enforcement agencies to strategically deploy resources and proactively combat crime in high-risk areas, making cities and neighborhoods safer for their residents.

*The rationale behind the usage of the technique*: Finding the most efficient way to connect a set of nodes in a graph while minimizing the total weight or cost of the connections can help in identifying groups of nodes that are closely connected and form clusters. This helps in identifying groups of nodes that are closely connected and form clusters. The approach is also an effective mechanism for selecting the minimum set of edges that connect all the nodes in the graph. This ensures that all nodes are connected with minimum latency.

*Conditions for the optimal performance of the technique*: The optimal performance of the MST technique is achieved when: (1) the graph is sparse (when the graph has fewer edges, the MST algorithms can run faster), (2) the edges of the graph have unique weights (if the edges of the graph have unique weights, the MST can be found in a single run of the MST algorithm), (3) the algorithm used is efficient (some MST algorithms have better performance than others), (4) the graph is connected (if it is not connected, the MST cannot be applied).

*Limitations of the technique*: The limitations include: (1) it assumes edge weights are distinct ( if there are multiple edges with the same weight, the algorithm may not choose the optimal edge to include in the MST), (2) it can be expensive for large graphs, (3) it does not consider edge direction (the MST algorithm only considers the weight of edges, but not their direction), and (4) it may not represent all relationships (the MST algorithm only considers the connections between nodes that form a tree, and may not represent all the relationships in the graph).

TABLE 37: FEATURING RSEARCH PAPERS (PP.) THAT HAVE EMPLOYED **MINIMUM SPANNING TREE-BASED ANALYSIS TECHNIQUES** AND THEIR EVALUATION IN TERMS OF SCALABILITY (SC.), INTERPRETABILITY (IN.), ACCURACY (AC.), AND EFFICIENCY (EF.)

| Pp. | Sc. | In. | Ac. | Ef. | Description |
|---|---|---|---|---|---|
| Taha [153] | Acceptable | Acceptable | Good | Fair | The authors developed ECLfinder, a forensic analysis system for detecting crimes and identifying influential criminals. This system builds a network representing a criminal organization based on its Mobile Communication Data or crime incident reports. The system constructs a MST where each vertex is an individual criminal. Using existence dependency, ECLfinder identifies the important vertices representing the influential criminals in the organization. A vertex $u$ is existence dependent on vertex $v$ if $u$ cannot reach any other vertex in the network through MST paths after v's removal. |
| Wang [154] | Unsatisfactory | Good | Acceptable | Fair | The authors proposed an improved MST clustering algorithm to detect potentially fraudulent money laundering transactions in financial applications, using a dissimilarity metric. They introduced a unique dissimilarity metric tailored to the nature of financial datasets and the identification of money laundering transactions. This metric gauges the variance among outliers. |

## 5.2 Vertices Connections-Based Analysis

### 5.2.1 Association Edge Analysis

Analyzing the edge strength within a social network can yield valuable insights for predicting crimes. Edge strength represents the measure of connection or interaction between individuals within the network, encompassing factors like the frequency and duration of interactions, the level of trust or influence, and shared interests among entities. This comprehensive evaluation of connections unveils patterns that may signal potential criminal behavior. Social network analysis uncovers sub-groups and the dynamics of criminal networks, measuring relationship strength. This data aids crime prediction using machine learning, including logistic regression, decision trees, random forests, and neural networks. Splitting the dataset for testing refines model performance. Algorithms like Apriori, FP-Growth, and Eclat find associations, offering law enforcement versatile tools for proactive crime prevention.

*The rationale behind the usage of the technique*: By examining the association edges in social networks, one can gain a deeper understanding of the social structures and dynamics that underlie human behavior and interaction, which is helpful in crime prediction. This can reveal potential opportunities for collaboration or areas of conflict within the network. By analyzing the connections and interactions between individuals, we can identify key players in criminal activities, potential collaborators or accomplices, and areas of conflict within the network.

*Conditions for the optimal performance of the technique*: The method can be improved by: (1) appropriate measurement of network ties (e.g., degree centrality, betweenness centrality, and closeness centrality) to determine the importance of individual nodes or groups in the network), (2) consideration of network context (the analysis should consider the broader context of the network, such as its size, structure, and dynamics, to understand the nature and implications of the relationships identified), (3) adequate statistical techniques to identify significant relationships in the network and control for confounding variables).

**Limitations of the technique**: The limitations include: (1) it only focuses on direct connections (social network does not account for indirect connections or relationships), (2) it does not account for the context (social network does not consider the context in which connections between nodes occur; for example, two individuals may be connected in a social network, but the nature of their connection may differ), (3) it assumes all connections are equal (some connections may be stronger than others), (4) it may not capture changes over time (social network are static).

**TABLE 38:** FEATURING RSEARCH PAPERS **(PP.)** THAT HAVE EMPLOYED **ASSOCIATION EDGE-BASED ANALYSIS TECHNIQUES** AND THEIR EVALUATION IN TERMS OF SCALABILITY **(SC.)**, INTERPRETABILITY **(IN.)**, ACCURACY **(AC.)**, AND EFFICIENCY **(EF.)**

| Pp. | Sc. | In. | Ac. | Ef. | Description |
|---|---|---|---|---|---|
| Alzabi [155] | Fair | Fair | Good | Good | The authors proposed a technique that can assess the significance of nodes in a graph in relation to a specified set of query nodes. It employs the strength of the connections between the nodes as a measure of their relative importance. It tackles incomplete and inconsistent contributions of query nodes, ensuring precise calculation of relative importance. |
| Taha [156] | Unsatisfactory | Fair | Good | Acceptable | The authors introduced a novel forensic analysis system known as IICCC. This system utilizes mobile phone communications data of a criminal organization's members to identify high-level criminals and prioritize crucial communication channels (i.e., association edges) accurately. |
| Ozgul [157] | Unsatisfactory | Fair | Acceptable | Fair | The authors developed a model, which improves upon the Social Detection Model by incorporating neighborhood cohabitation as a new feature. They examined the possibility of utilizing social similarity characteristics to identify individuals within criminal networks, using the association edges between nodes within the network as a basis for detection. |
| Jhee [158] | Fair | Good | Good | Unsatisfactory | The authors put forward a method for a vast criminal network. The method resembles a sandwich panel, was created with a distinctive configuration: one side comprises a crime case network, while the other comprises a people network consisting of victims, criminals, witnesses, and others. Association edges connecting each case to its related people link the two networks. |
| Taha [159] | Good | Unsatisfactory | Fair | Acceptable | The authors proposed CLDRI, a forensic analysis system that aims to identify the key members of a criminal organization. Their approach involves developing formulas that assign importance scores to individual criminals based on the strength of their associations to other members within the organization. |

### 5.2.2 K-Core-Based Analysis

K-core analysis is a robust method for exploring the intricacies of social network structures, involving the discovery of densely interconnected subgraphs, known as k-cores, within a larger network. A k-core is essentially a subgraph where each node is linked to at least k other nodes within the same subgraph. It serves as a valuable tool for pinpointing closely-knit and cohesive communities within the social network, whether grounded in shared interests, geographic proximity, or other factors, shedding light on the network's dynamics.

To perform k-core analysis, we first represent the network as a graph, with nodes signifying individual actors and edges representing their relationships. We then identify the k-core by systematically removing nodes with fewer than k connections until all remaining nodes possess at least k connections. This iterative process repeats until no further nodes can be removed. The result is a k-core subgraph derived from the original network, where all nodes maintain at least k connections. This analytical technique provides profound insights into the underlying structure and dynamics of the social network, aiding researchers in understanding its inner workings.

*The rationale behind the usage of the technique*: The technique allows one to identify the most tightly connected subgroups within a network. This can be useful for understanding the structure of a network and the relationships between individuals within the network. Nodes that are part of the k-core of a network are likely to be important hubs or connectors within the network. These nodes may be particularly influential in the spread of information or the transmission of behaviors within the network.

*Conditions for the optimal performance of the technique:* The technique can be improved by: (1) selection of appropriate *k* value to ensure that the analysis captures the most relevant and informative structures in the network), (2) network size and density (large and dense networks can be computationally intensive and may require specialized algorithms), and (3) choice of algorithm (some algorithms may be more suitable for large or dense networks, while others may be better suited for sparse or networks).

*Limitations of the technique:* The limitations include: (1) limited information about the global structure (k-core analysis focuses only on the local structure of the network, ignoring the global structure), (2) sensitivity to parameter choices (if *k* is set too high, the analysis may miss important nodes and connections; if *k* is set too low, the analysis may include too many nodes and connections, making it difficult to draw meaningful conclusions), (3) limited insight into network dynamics (it does not consider the dynamics of social networks).

**TABLE 39:** FEATURING RSEARCH PAPERS **(PP.)** THAT HAVE EMPLOYED **K-CORE-BASED ANALYSIS TECHNIQUES** AND THEIR EVALUATION IN TERMS OF SCALABILITY **(SC.)**, INTERPRETABILITY **(IN.)**, ACCURACY **(AC.)**, AND EFFICIENCY **(EF.)**

| Pp. | Sc. | In. | Ac. | Ef. | Description |
|---|---|---|---|---|---|
| Taha [160] | Fair | Acceptable | Good | Acceptable | The authors introduced SIIMCO, a forensic analysis system capable of identifying the key members of a criminal organization as well as the immediate leaders of lower-level criminals on a provided criminal network. Unlike current methods, SIIMCO utilizes k-core analysis to overcome the limitations of incomplete and inconsistent contributions. |
| Ozgul [161] | Unsatisfactory | Unsatisfactory | Acceptable | Acceptable | The authors compared criminal network detection models based on crime features with the k-core and n-clique algorithms. These models employ relational data and utilize an inner-join query approach to establish connections between entities. A scoring function is used to determine the maximum likelihood similarities between the entities. The performance of the n-clique and k-core algorithms was compared against crime-specific criminal network detection models. |

## 7 EXPERIMENTAL EVALUATIONS

In this section, we assess and rank the various techniques described in this paper through experimentation. For each group of algorithms utilizing the same underlying technique, we chose one to serve as a representative. These selected algorithms were then evaluated and ranked. They were executed on a Windows 10 Pro machine equipped with an Intel(R) Core(TM) i7-6820HQ processor running at 2.70 GHz and 16 GB of RAM.

### 7.1 Datasets

The evaluations were conducted using the following datasets:

- *Chicago Crime Dataset:* The Chicago Police Department dataset is a public resource that provides information on reported crime incidents that have occurred in Chicago since 2001. The data is regularly updated using information from the Police Department's CLEAR system and includes unverified reports. This dataset contains information about each reported crime, such as the type of crime, location, date and time of occurrence, and whether or not an arrest was made. The dataset includes fields such as date and time of occurrence, address, crime type, FBI code, incident description, and location type. The dataset can be downloaded from [162].
- *San Francisco crime dataset:* It contains 39 categories of crimes, and the most common crime in the dataset is larceny/theft. However, the frequencies of all the classes in the dataset are not equally distributed, meaning that some crimes occur more frequently than others in San Francisco. It is important to consider these variations when analyzing crime data, as they can affect the accuracy and usefulness of the results. The dataset can be downloaded from [163].
- *Caviar crime dataset [164, 165]:* The dataset known as Caviar pertains to the communications of a Canadian gang named Caviar, which was involved in drug trafficking activities in Montreal, Canada. The dataset is based on phone conversations among the gang members during the period from 1994 to 1996, and it represents a network of 110 nodes, with each node corresponding to a different gang member. The gang's main activities involved distributing hashish and cocaine.
- *Diyarbakir crime dataset [166, 167]:* Diyarbakir is a city in southeastern Turkey with a population over one million. The Diyarbakir dataset encompasses 1,370 instances of criminal activity related to drug trafficking, documenting the apprehension and subsequent arrest of 2,552 individuals engaged in such activities between the years 2006 and 2011. Furthermore, this dataset identifies 221 previously recognized drug dealing networks and includes comprehensive demographic information on the offenders. This information comprises surnames, given names, residential addresses, dates of birth, provinces of residence, and places of birth.

### 7.2 Methodology for Selecting a Representative Algorithm for Each Technique, Ranking the Various Techniques, Sub-Techniques, Sub-Categories, and Categories

We conducted the experimental evaluations using the following approach:

> ➤ **Evaluating each sub-technique**: After reviewing the papers that reported algorithms utilizing a particular sub-technique, we identified the most influential paper. The algorithm presented in this paper was chosen to serve as the sub-technique's representative. To determine the most significant paper among all papers reporting algorithms that use the same sub-technique, we assessed different factors, including how cutting-edge it is and its publication date. Table 40 shows the list of selected papers as well as the time took for their training.
> ➤ **Ranking the sub-techniques that belong to the same overall technique**: We calculated the average scores of the chosen algorithms that utilized the same sub-technique. Then, we ranked the sub-techniques that fall under the same technique based on their scores.
> ➤ **Ranking the various techniques that belong to the same sub-category**: We calculated the average scores of the chosen algorithms that utilized the same technique. Then, we ranked the techniques that fall under the same sub-category based on their scores.
> ➤ **Ranking the various sub-categories that belong to the same category**: We calculated the average scores of the chosen algorithms that utilized the same sub-category. Then, we ranked the sub-categories that fall under the same category based on their scores.

We searched for publicly available codes for the algorithms that we selected as representative of their techniques. We could obtain codes for the following papers: [29[1], 61[2], 67[3], 73[4], 32[5], 44[6], 133[7]]. As for the remaining representative papers, we developed our own implementations using TensorFlow [168] and trained them with Adam optimizer [168]. TensorFlow's APIs allow users to create their own ML algorithms [165]. We used Python 3.6 as our development language and TensorFlow 2.10.0 as the backend for the models.

### 7.3 Evaluation Setup

- *For CNN-based and Residual Network-based models:* To begin the training, we established a baseline of 20 epochs and a learning rate of 0.01, which we then fine-tuned based on the results. The Adam optimizer was utilized to optimize the loss function and update the weights during training. We maintained a constant length of 30 for the temporal sliding window and used convolution filters with a fixed size of 3 x 3.
- *For LSTM-based and BiLSTM-based models:* A 64-unit LSTM model was utilized to process the dataset. The embedding dimension was set to 64 and the maximum feature set was limited to 2500. A spatial dropout 1D layer with a dropout rate of 0.4 units was added after the first LSTM layer, followed by another LSTM layer with a recurrent dropout rate of 0.2 units. A 64-unit dense layer with a RELU activation function was added next. Finally, a final dense layer with 6 units and a Softmax activation function was added. The categorical cross-entropy was used as the loss function, and the Keras Adam optimizer was employed with a learning rate of 0.01. The model was trained using a 20% testing and 20% validation split over 16 batches and 5 epochs. To extract features from the input signal, we employed a kernel size 3. In order to capture temporal dependencies in the data, we utilized a bidirectional LSTM layer with 64 units, followed by another bidirectional LSTM layer with 32 units. To avoid overfitting, we incorporated dropout regularization. We added hidden layers with 30 dimensions between the softmax layer and the final output of the BiLSTM, and for neural network-based approaches, we set the node representation dimensions to 300. We initialized with random vectors.
- *For Deep Neural Network-based models:* We chose L1 as a regularization method and sigmoid as the activation function. We employed Adam optimizer to update the parameters of the network during training. We used dropout to randomly drop out some neurons during training, which can prevent overfitting and improve generalization.
- *For Decision Tree-based models:* We started with 10 trees and gradually increased the number until a suitable value is found. We set the maximum depth of each tree to 10. We set the minimum number of samples required to split a node to 2.
- *For Kernel Density-based models:* We experimented with a variety of bandwidth values, ranging from 0.1 to 1.0 in steps of 0.1. For each bandwidth value, we evaluated different minimum point thresholds for forming a cluster, ranging from 5 to 20 in increments of 5.
- *For Linear Regression-based models:* The regression line was adjusted with various intercept values including -10, -5, 0, 5, and 10. After evaluating the fit using the R-squared value, we found that the intercept of 0 gave the best performance. Therefore, we utilized this intercept value to fit the data.
- *For Logistic Regression-based models:* The logistic regression model was fitted using different values of the regularization parameter C, including 0.1, 1, and 10, and penalty functions of 'l1' and 'l2'. Additionally, we tested the model using class weights of 'balanced' and 1.0. To evaluate the classification performance, we used the area under the ROC curve, and the best result was obtained with C = 1, penalty = 'l2', solver = 'liblinear', and class_weight = 'balanced'.

We employed the following metrics to evaluate the performance of the methods:

- For the methods that employ classification and clustering-based techniques, we utilized accuracy and F1-score metrics. The accuracy metric determines the ratio of correct predictions made by a model to the total predictions. The F1-score metric combines precision and recall into a single metric by taking their harmonic mean.
- For the methods that employ regression-based techniques, we employed the RMSE (Root Mean Squared Error) metric, which calculates the average difference between the predicted and actual values of a continuous variable.
- For the methods that employ social network-based techniques, we employed recall, precision, and F1-score metrics. Recall is the proportion of actual positives correctly identified. Precision is the proportion of positive identifications that were actually correct.

We implemented 10-fold cross-validation. Table 40 presents the training duration allocated to each method. The division of the datasets into training, validation, and test sets was as follows:

- **Deep Learning Models** (*CNN-based [24], Residual Network [29, 48], CNN-BiLSTM [32], Feature-based [36], BERT-based [38], LSTM Graph Attention [50], LSTM Neural Network [52], Bi-directional LSTM [55], Recurrent CNN [61], Convolutional Graph Attention [67], Deep Neural Network [73]*): These models benefit from large training datasets due to their complexity and capacity to model high-level abstractions. Therefore, the division of the datasets should ensure they have ample data to learn from. We adopted the following divisions of the training, validation, and test datasets: Training Set: 80%, Validation Set: 10%, and Test Set: 10%.
- **Feature-based and Traditional Machine Learning Models** (*Gradient Boosting Machine [76], Adaptive Boosting [80], Decision Tree [87], Boosted Decision Trees [90], K-Nearest Neighbor [93], Kernel Density Estimation Neighbor [97], Random Forest Estimation [101], Bootstrap Aggregation [107]*): These models might not require as much data as deep learning models to perform well. However, using the same dataset division allows for a fair comparison across all models. We adopted the following division of the training, validation, and test datasets: Training Set: 70%, Validation Set: 15%, and Test Set: 15%.

---

[1] https://github.com/kenshohara/3D-ResNets
[2] https://github.com/IBM/EvolveGCN
[3] https://github.com/hexiangnan/attentional_factorization_machine
[4] https://github.com/shadowkshs/Grid-based-crime-prediction-model-using-geographical-features
[5] https://github.com/cstorm125/thai2fit
[6] https://github.com/cstorm125/thai2fit
[7] https://cran.r-project.org/web/packages/C50/index.html

TABLE 40. LIST OF SELECTED REPRESENTATIVE PAPERS (REP. PAP.). THE TABLE ALSO SHOWS THE TIME IT TOOK TO TRAIN EACH METHOD

| Technique/sub-technique | Rep. Pap. | Training Time (hr.) | Technique/sub-technique | Rep. Pap. | Training Time (hr.) | Technique/sub-technique | Rep. Pap. | Training Time (hr.) |
|---|---|---|---|---|---|---|---|---|
| Spatial CNN-based | [25] | 8.7 | Spatiotemporal Deep Neural Network | [73] | 3.4 | Spatiotemporal Density Clustering | [123] | 0.73 |
| Spatial Residual Network-based | [29] | 13.8 | Spatial Gradient Boosting Machine | [76] | 1.95 | Self-Exciting Point Process | [125] | 26.6 |
| Spatial CNN-BiLSTM-based | [32] | 37.6 | Spatial Adaptive Boosting | [80] | 1.3 | Spatiotemporal Hierarchical structure | [127] | 33.6 |
| Spatial Feature-based | [36] | 2.6 | Spatial Naïve Bays Estimation | [84] | 1.2 | Spatiotemporal Hierarchical Density | [130] | 55.2 |
| Spatial BERT-based | [38] | 49.2 | Spatial Decision Tree | [87] | 0.6 | Shortest Path-based | [150] | 1.2 |
| Spatiotemporal CNN-based | [44] | 9.3 | Spatial Boosted Decision Tree | [90] | 1.55 | Minimum Spanning Tree-based | [153] | 0.89 |
| Spatiotemporal Residual Network | [48] | 14.6 | Spatial K-Nearest Neighbor | [93] | 0.7 | Association Edge Analysis-based | [155] | 22.4 |
| Spatiotemporal LSTM Graph Att. | [51] | 38.1 | Spatial Kernel Density E. Neighbor | [97] | 0.95 | K-Core Analysis-based | [160] | 0.68 |
| LSTM Neural Network | [52] | 22.4 | Spatiotemporal Random Forest Es. | [101] | 1.2 | Spatial Density | [116] | 1.3 |
| Spatiotemporal Bi-Directional LSTM | [55] | 27.9 | Spatiotemporal Bootstrap Aggregation | [106] | 1.1 | Spatiotemporal Kernel Density | [120] | 1.6 |
| Spatiotemporal Recurrent CNN | [61] | 31.3 | Spatial Hierarchical Agglomerative | [108] | 0.96 | | | |
| Convolutional Graph Attention | [67] | 38.1 | Spatial K-Means | [112] | 0.45 | | | |

## 7.4 The Experimental Results

We provide in Subsections 7.4.1-7.4.4 the scores achieved by the different methods and their rankings, as follows. Subsection 7.4.1 presents the accuracy values and F1-scores of the methods that employ classification-based techniques using both Chicago and San Francisco crime datasets. Subsection 7.4.2 presents the accuracy values and F1-scores of the methods that employ clustering-based techniques using both Chicago and San Francisco crime datasets. Subsection 7.4.3 presents the RMSE values of the methods that employ regression-based techniques using both Chicago and San Francisco crime datasets. Subsection 7.4.4 presents the Recall, Precision, and F1-score values of the methods that employ social network-based techniques using both Caviar and Diyarbakir datasets.

### 7.4.1 The Methods Employing Classification-based Techniques

Tables 41 and 42 show the ranking of the sub-techniques, techniques, and sub-categories based on the accuracies and F1-scores, respectively, of the methods. Figs. 2 and 3 depict the individual accuracies and F1-scores, respectively, of the methods grouped based on their techniques.

TABLE 41. RANKING OF THE CLASSIFICATION'S SUB-TECHNIQUES, TECHNIQUES, AND SUB-CATEGORIES BASED ON THEIR ACCURACY VALUES

| Categ. | Sub-Categ. | Technique | Sub-Technique | Selected Papers | Datasets | Acc % | Sub-Tech Rank | Tech. Rank | Sub-Categ. Rank |
|---|---|---|---|---|---|---|---|---|---|
| Deep Learning Methods | Spacial-based | Spatial-based Neural Network | CNN-based | [24] | Chi / SFO | 81.6 / 86.2 | 1 | 5 | 2 |
| | | | Residual Network | [29] | Chi / SFO | 81.3 / 85.4 | 3 | | |
| | | | CNN-BiLSTM | [32] | Chi / SFO | 82.1 / 86.7 | 2 | | |
| | | Pre-trained-based | Feature-based | [36] | Chi / SFO | 78.7 / 82.3 | 2 | 4 | |
| | | | BERT-based | [38] | Chi / SFO | 77.4 / 79.6 | 1 | | |
| | Spatio-Temporal | Spatiotemporal-based Neural Network | CNN-based | [44] | Chi / SFO | 82.5 / 87.2 | 2 | 2 | 1 |
| | | | Residual Networks | [48] | Chi / SFO | 82.8 / 87.7 | 3 | | |
| | | | LSTM Graph Attention | [51] | Chi / SFO | 83.5 / 88.7 | 1 | | |
| | | Recursive-based Neural Network | LSTM Neural Network | [52] | Chi / SFO | 84.3 / 89.5 | 2 | 1 | |
| | | | Bi-directional LSTM | [55] | Chi / SFO | 83.9 / 89.1 | 1 | | |
| | | | Recurrent CNN | [61] | Chi / SFO | 83.2 / 88.3 | 3 | | |
| | | Graph Attention | Convolutional G. Attention | [67] | Chi / SFO | 79.2 / 83.3 | 2 | 3 | |
| | | | Deep Neural Network | [73] | Chi / SFO | 80.9 / 84.4 | 1 | | |
| Classical Methods | Spacial-based | Adaptive Ensemble | Gradient Boosting Machine | [76] | Chi / SFO | 79.6 / 82.5 | 1 | 3 | 1 |
| | | | Adaptive Boosting | [80] | Chi / SFO | 79.3 / 81.9 | 2 | | |
| | | Naive Bayes Estimation | | [84] | Chi / SFO | 75.1 / 76.4 | | 5 | |
| | | Decision Support | Decision Tree | [87] | Chi / SFO | 78.8 / 81.1 | 2 | 1 | |
| | | | Boosted Decision Trees | [90] | Chi / SFO | 79.1 / 81.4 | 1 | | |
| | | Distribution of Datapoints | K-Nearest Neighbor | [93] | Chi / SFO | 75.4 / 77.3 | 2 | 4 | |
| | | | Kernel density Est. Neighbor | [96] | Chi / SFO | 77.6 / 79.2 | 1 | | |
| | Spatio-Temp. | Random Forest Estimation | | [101] | Chi / SFO | 74.6 / 75.2 | | 6 | 2 |
| | | Bootstrap Aggregation | | [106] | Chi / SFO | 78.3 / 80.6 | | 2 | |

**TABLE 42** RANKING OF THE **CLASSIFICATION'S** SUB-TECHNIQUES, TECHNIQUES, AND SUB-CATEGORIES BASED ON THEIR **F1-SCORE** VALUES

| Categ. | Sub-Categ. | Technique | Sub-Technique | Selected Papers | Data sets | F1-score | Sub-Tech. Rank | Tech. Rank | Sub-Categ. Rank |
|---|---|---|---|---|---|---|---|---|---|
| Deep Learning Methods | Spacial-based | Spatial-based Neural Network | CNN-based | [24] | Chi / SFO | 53.2 / 62.3 | 1 | 5 | 2 |
| | | | Residual Network | [29] | Chi / SFO | 49.3 / 56.1 | 3 | | |
| | | | CNN-BiLSTM | [32] | Chi / SFO | 51.4 / 59.5 | 2 | | |
| | | Pre-trained-based | Feature-based | [36] | Chi / SFO | 55.2 / 64.6 | 2 | 4 | |
| | | | BERT-based | [38] | Chi / SFO | 57.3 / 68.0 | 1 | | |
| | Spatio-Temporal | Spatiotemporal-based Neural Network | CNN-based | [44] | Chi / SFO | 64.5 / 75.2 | 2 | 2 | 1 |
| | | | Residual Networks | [48] | Chi / SFO | 61.3 / 76.6 | 3 | | |
| | | | LSTM Graph Attention | [51] | Chi / SFO | 67.1 / 79.7 | 1 | | |
| | | Recursive-based Neural Network | LSTM Neural Network | [52] | Chi / SFO | 72.0 / 85.2 | 2 | 1 | |
| | | | Bi-directional LSTM | [55] | Chi / SFO | 76.1 / 84.4 | 1 | | |
| | | | Recurrent CNN | [61] | Chi / SFO | 69.2 / 81.5 | 3 | | |
| | | Graph Attention | Convolutional G. Attention | [67] | Chi / SFO | 59.2 / 70.3 | 2 | 3 | |
| | | | Deep Neural Network | [73] | Chi / SFO | 60.4 / 72.0 | 1 | | |
| Classical Methods | Spacial-based | Adaptive Ensemble | Gradient Boosting Machine | [76] | Chi / SFO | 58.6 / 64.1 | 1 | 3 | 1 |
| | | | Adaptive Boosting | [80] | Chi / SFO | 55.3 / 62.3 | 2 | | |
| | | Naive Bayes Estimation | | [84] | Chi / SFO | 45.6 / 48.4 | | 5 | |
| | | Decision Support | Decision Tree | [87] | Chi / SFO | 61.3 / 71.2 | 2 | 1 | |
| | | | Boosted Decision Trees | [90] | Chi / SFO | 61.5 / 74.6 | 1 | | |
| | | Distribution of Datapoints | K-Nearest Neighbor | [93] | Chi / SFO | 49.1 / 53.2 | 2 | 4 | |
| | | | Kernel density Est. Neighbor | [96] | Chi / SFO | 53.5 / 58.3 | 1 | | |
| | Spatio-Temp. | Random Forest Estimation | | [101] | Chi / SFO | 42.0 / 44.2 | | 6 | 2 |
| | | Bootstrap Aggregation | | [106] | Chi / SFO | 60.3 / 67.5 | | 2 | |

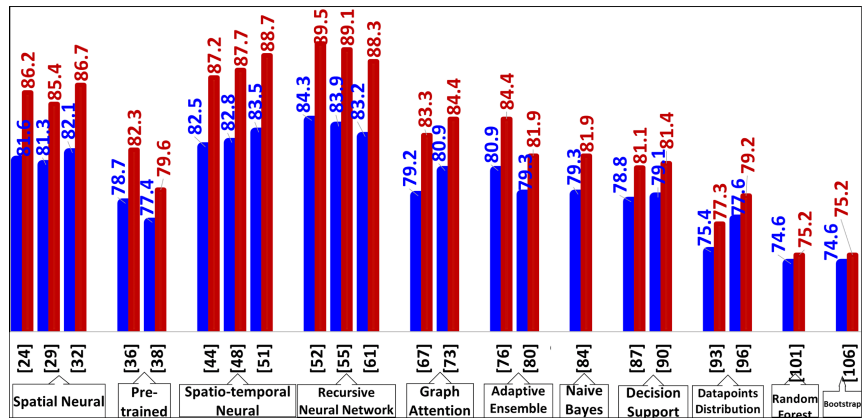

Fig. 2. The individual **accuracy** values of the **classification** algorithms. The algorithms are grouped based on their techniques.

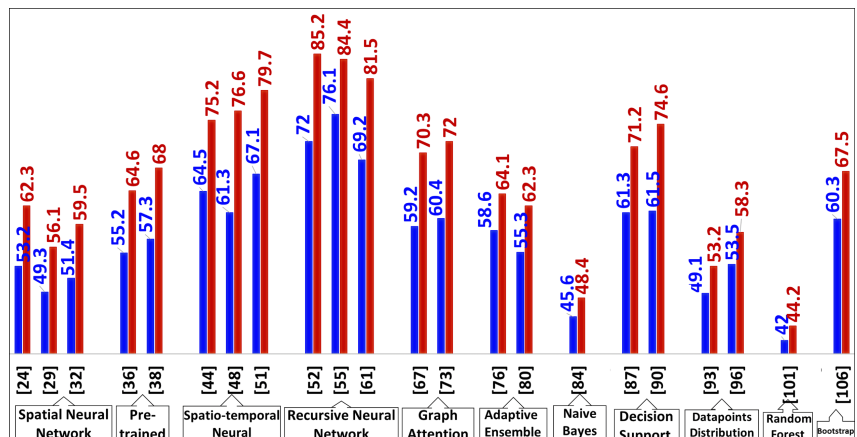

Fig. 3. The individual **F1-score** values of the **classification** algorithms. The algorithms are grouped based on their techniques.

### 7.4.2 The Methods that Employ Clustering-based Techniques

Tables 43 and 44 show the ranking of the sub-techniques, techniques, and sub-categories based on their accuracies and F1-scores, respectively. Figs. 4 and 5 depict the accuracies and F1-scores, respectively, of the methods that employ clustering-based techniques, grouped based on their techniques.

TABLE 43 RANKING OF THE **CLUSTERING'S** SUB-TECHNIQUES, TECHNIQUES, AND SUB-CATEGORIES BASED ON THEIR **ACCURACY** VALUES

| Categ. | Technique | Sub-Technique | Selected Papers | Datasets | Acc % | Sub-Tech. Rank | Tech. Rank | Categ. Rank |
|---|---|---|---|---|---|---|---|---|
| Spatial | Partitioning based | Hierarchical Agglomerative | [108] | Chi | 87.3 | 2 | 2 | 2 |
| | | | | SFO | 88.5 | | | |
| | | K-means | [112] | Chi | 85.6 | 1 | | |
| | | | | SFO | 88.3 | | | |
| | Spatial Density | | [116] | Chi | 88.7 | N/A | 1 | |
| | | | | SFO | 89.4 | | | |
| Spatio-temporal | Structure based | Kernel Density Clustering | [120] | Chi | 95.2 | 1 | 1 | 1 |
| | | | | SFO | 97.4 | | | |
| | | Density-based Clustering | [123] | Chi | 91.5 | 3 | | |
| | | | | SFO | 92.1 | | | |
| | | Self-exciting Point Process | [125] | Chi | 94.3 | 2 | | |
| | | | | SFO | 96.6 | | | |
| | Partitioning based | Hierarchical Structure | [127] | Chi | 90.6 | 2 | 2 | |
| | | | | SFO | 91.2 | | | |
| | | Hierarchical Density | [130] | Chi | 92.8 | 1 | | |
| | | | | SFO | 93.4 | | | |

TABLE 44 RANKING OF THE **CLUSTERING'S** SUB-TECHNIQUES, TECHNIQUES, AND SUB-CATEGORIES BASED ON THEIR **F1-SCORE** VALUES

| Categ. | Technique | Sub-Technique | Selected Papers | Datasets | F1-score | Sub-Tech. Rank | Tech. Rank | Categ. Rank |
|---|---|---|---|---|---|---|---|---|
| Spatial | Partitioning based | Hierarchical Agglomerative | [108] | Chi | 84.3 | 2 | 1 | 2 |
| | | | | SFO | 85.6 | | | |
| | | K-means | [112] | Chi | 85.7 | 1 | | |
| | | | | SFO | 86.4 | | | |
| | Spatial Density | | [116] | Chi | 82.7 | N/A | 2 | |
| | | | | SFO | 84.9 | | | |
| Spatio-Temporal | Structure based | Kernel Density Clustering | [120] | Chi | 90.1 | 2 | 1 | 1 |
| | | | | SFO | 91.2 | | | |
| | | Density-based Clustering | [123] | Chi | 87.8 | 3 | | |
| | | | | SFO | 89.7 | | | |
| | | Self-exciting Point Process | [125] | Chi | 90.3 | 1 | | |
| | | | | SFO | 92.6 | | | |
| | Partitioning based | Hierarchical Structure | [127] | Chi | 87.8 | 1 | 2 | |
| | | | | SFO | 89.1 | | | |
| | | Hierarchical Density | [130] | Chi | 87.2 | 2 | | |
| | | | | SFO | 88.4 | | | |

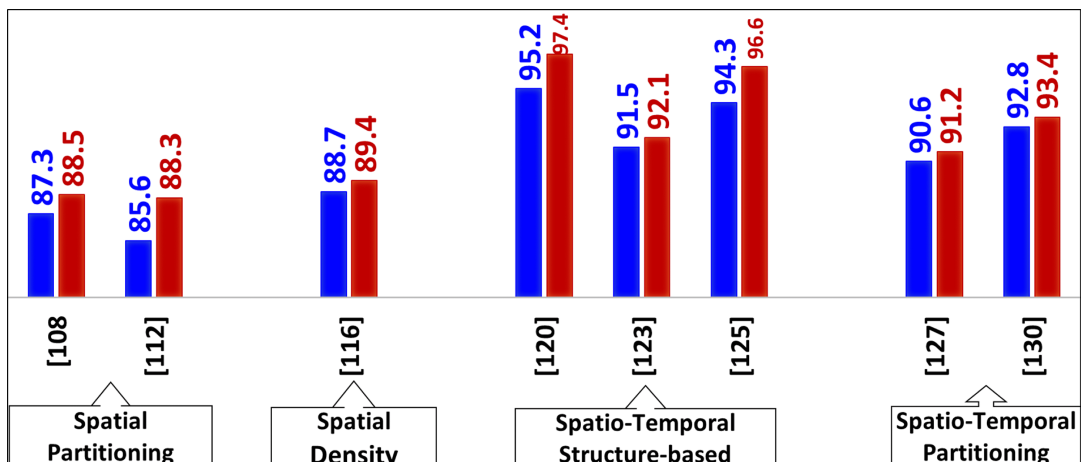

Fig. 4. The individual **accuracy** values of the **clustering** algorithms. The algorithms are grouped based on their techniques.

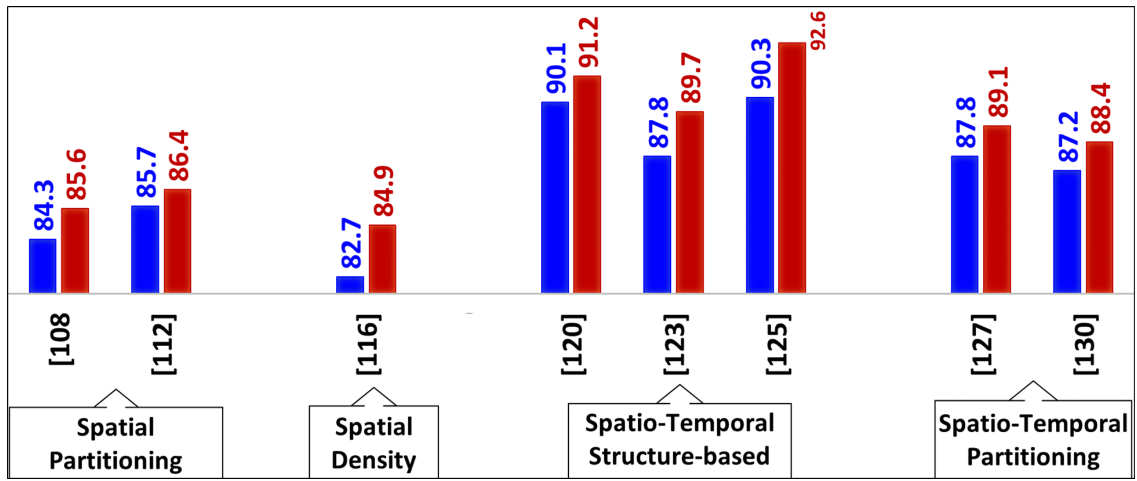

**Fig. 5.** The individual **F1-score** values of the **clustering** algorithms. The algorithms are grouped based on their techniques.

### 7.4.3 The Methods that Employ Regression-based Techniques

Fig. 6 depicts the individual RMSE scores of the methods that employ regression-based techniques, grouped based on their techniques. Table 45 shows the ranking of the regression's sub-technique and techniques according to their corresponding RMSE values.

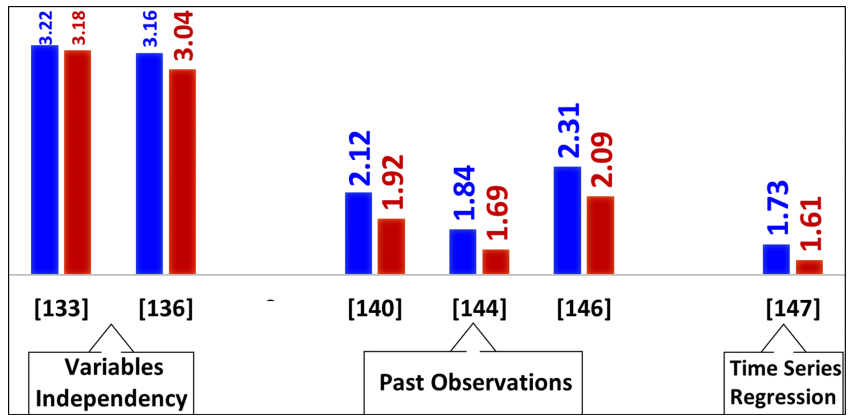

**Fig. 6.** The individual **RMSE** scores of the **regression** algorithms. The algorithms are grouped based on their techniques.

TABLE 45. RANKING OF THE REGRESSION'S SUB-TECHNIQUES AND TECHNIQUES BASED ON THEIR RMSE SCORES

| Technique | Sub-Technique | Selected Papers | Data sets | RMSE | Sub-Technique Rank | Technique Rank |
|---|---|---|---|---|---|---|
| Variables Independency | Bayesian Logistic Regression | [133] | Chi | 3.22 | 2 | 3 |
| | | | SFO | 3.18 | | |
| | Linear Regression | [136] | Chi | 3.16 | 1 | |
| | | | SFO | 3.04 | | |
| Past Observations | Logistic Regression | [140] | Chi | 2.12 | 2 | 2 |
| | | | SFO | 1.92 | | |
| | Auto Regressive | [144] | Chi | 1.84 | 1 | |
| | | | SFO | 1.69 | | |
| | Support Vector Regression | [146] | Chi | 2.31 | 3 | |
| | | | SFO | 2.09 | | |
| Time-Series Regression | | [147] | Chi | 1.73 | N/A | 1 |
| | | | SFO | 1.61 | | |

### 7.4.4 The Methods that Employ Social Network-based Techniques

Figs. 7 and 8 depict the recalls, precisions, and F1-scores of the methods using the Caviar and Diyarbakir datasets, respectively. The methods are grouped based on the techniques they employ. Tables 46 and 47 illustrate the ranking of the techniques according to their corresponding scores using the Caviar and Diyarbakir datasets, respectively.

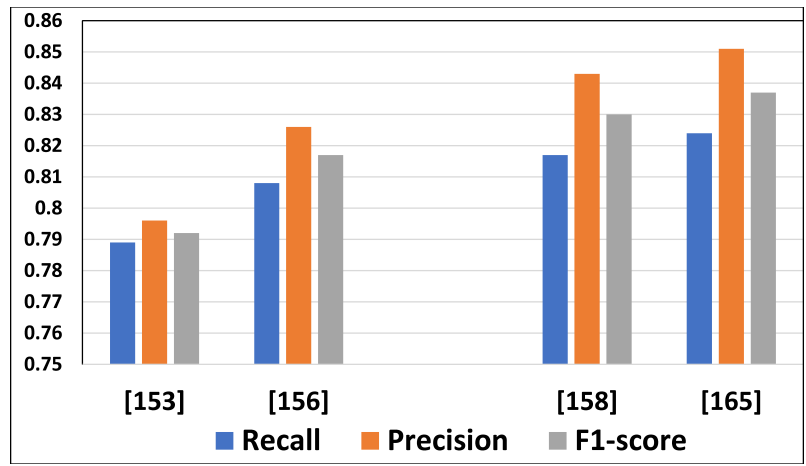

**Fig. 7.** The individual scores of the algorithms that employ **Social Network**-based techniques using the **Caviar dataset**.

**TABLE 46.** RANKING OF THE **SOCIAL NETWORK**-BASED'S SUB-TECHNIQUES AND TECHNIQUES USING THE **CAVIAR DATASET**

| Technique | Sub-Technique | Selected Papers | Metric | Score | Sub-Technique Rank | Technique Rank |
|---|---|---|---|---|---|---|
| Path based | Shortest Path | [150] | Recall | 0.789 | 2 | 2 |
| | | | Precision | 0.796 | | |
| | | | F1-score | 0.792 | | |
| | Minimum Spanning Tree | [153] | Recall | 0.808 | 1 | |
| | | | Precision | 0.826 | | |
| | | | F1-score | 0.817 | | |
| Vertices Connections | Association Edge Analysis | [155] | Recall | 0.817 | 2 | 1 |
| | | | Precision | 0.843 | | |
| | | | F1-score | 0.830 | | |
| | K-core Analysis | [16] | Recall | 0.824 | 1 | |
| | | | Precision | 0.851 | | |
| | | | F1-score | 0.837 | | |

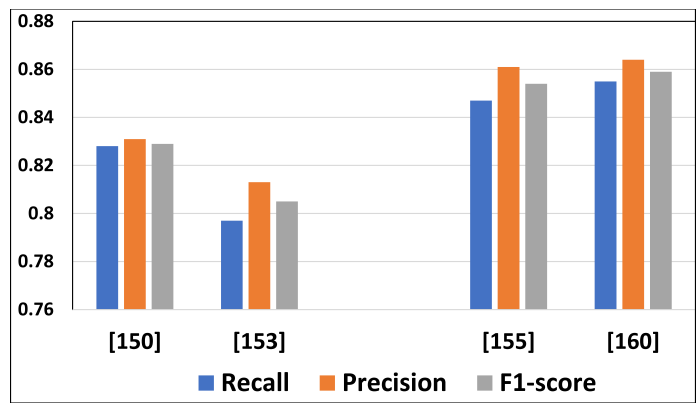

**Fig. 8.** The scores of the algorithms that employ **Social Network**-based techniques using the **Diyarbakir dataset**.

**TABLE 47.** RANKING OF THE **SOCIAL NETWORK**-BASED'S SUB-TECHNIQUES AND TECHNIQUES USING THE **DIYARBAKIR DATASET**

| Technique | Sub-Technique | Selected Papers | Metric | Score | Sub-Technique Rank | Technique Rank |
|---|---|---|---|---|---|---|
| Path based | Shortest Path | [150] | Recall | 0.828 | 1 | 2 |
| | | | Precision | 0.831 | | |
| | | | F1-score | 0.829 | | |
| | Minimum Spanning Tree | [153] | Recall | 0.797 | 2 | |
| | | | Precision | 0.813 | | |
| | | | F1-score | 0.805 | | |
| Vertices Connections | Association Edge Analysis | [155] | Recall | 0.847 | 2 | 1 |
| | | | Precision | 0.861 | | |
| | | | F1-score | 0.854 | | |
| | K-core Analysis | [160] | Recall | 0.855 | 1 | |
| | | | Precision | 0.864 | | |
| | | | F1-score | 0.859 | | |

## 7.5 Discussion of the Experimental Results

We now examine and interpret the outcome results obtained from our experiments. We will elaborate on the importance of the findings, clarify any trends or patterns that surfaced during the experiments, and highlight any constraints or shortcomings.

### 7.5.1 Discussion of the Results of the Classification Algorithms

#### 7.5.1.1 The Algorithms that Employ Convolutional Neural Networks-Based Approaches [24, 32, 44, 61, 67]

These algorithms, particularly the one that utilizes spatiotemporal mechanisms, exhibited impressive precision and were able to capture semantic features by modifying the convolution kernel's height to address long-range connections and process more data. However, the experiments indicated that they had significant computational complexity and necessitated a significant amount of training data. Additionally, due to the networks' increased depth, the algorithms were susceptible to gradient dispersion.

Algorithms [44, 61, 67], which leverage spatiotemporal mechanisms demonstrated notable precision, effectively capturing semantic features through adaptive convolution kernel adjustments for enhanced data processing and long-range connections. However, these benefits come at the cost of increased computational complexity, extensive training data requirements, and susceptibility to gradient dispersion due to network depth.

Further analysis revealed the capability of neural network models to predict crime locations within neighboring zip codes, suggesting a partial geographical clustering of crime types. This insight underscores the significance of spatial and temporal resolution, alongside crime type, in influencing neural network performance. The effectiveness of these models hinges on appropriate data preprocessing and the deployment of an optimal network architecture.

The results highlight a correlation between non-urgent social events and crime incidents, though capturing the complex spatial-temporal relationship between them remains challenging. A notable limitation is the performance dip in scenarios of sparse crime events, affecting spatio-temporal prediction resolution. This challenge has been partly addressed through innovative approaches like the spatio-temporal weighted graph, enhancing the representation of crime data for improved prediction outcomes.

Overall, the experimental results affirm that CNN-based models' success in crime prediction depends on multiple factors, including spatial-temporal resolution, crime type, data preprocessing, and network architecture.

#### 7.5.1.2 The Algorithms that Employ Recursive Neural Networks-Based Approaches [51, 55, 61]

Our investigation into the impact of training data volume, model architecture, and hyperparameter optimization on prediction performance yielded insightful results. Notably, we determined that while both insufficient and excessive training data did not guarantee improved outcomes, a balanced approach was essential for effective model training. Specifically, our analysis revealed that the LSTM-based models outperformed traditional neural network architectures in predictive accuracy.

Further experimentation led to the identification of optimal models' configurations for LSTM-based models across various datasets. An ideal training duration was established at 290 epochs, with the model architecture comprising 54 layers proving most effective. Additionally, the number of neurons, significantly influenced by the cell state parameter, was optimized at a cell state of 57, enhancing models' performance.

The influence of grid size on prediction accuracy was also examined, highlighting that while larger grids covering an entire city failed to yield meaningful predictions due to the low likelihood of crime in vast areas, overly small grids resulted in sparse data matrices, impeding useful forecasting. Thus, selecting an appropriate grid size emerged as a crucial factor in developing predictive models, underscoring the importance of balancing granularity with data density to optimize prediction accuracy.

We attributed the remarkable performance of these algorithms to several factors:
   i. Their capability to process each sentence in the datasets separately and then integrate the information to classify it. This is possible because they are designed to accommodate inputs of varying lengths. This is crucial for crime text classification since the length of crime reports can vary significantly.
   ii. Their aptitude to capture long-term dependencies, which allowed them to capture the temporal relationships between words in a sentence or between sentences in a paragraph. This enabled them to process each sentence or paragraph independently and combined the information to classify it.
   iii. Their proficiency in understanding the context of the dataset's terms by considering the relationships between words in a sentence. This feature helped the algorithms to identify the true meaning of the sentence, even in cases where the words themselves may not provide a clear indication of the crime.
   iv. These algorithms are adept at handling long text by breaking it down into smaller components and recursively applying a neural network to each component. This capability allowed them to mitigate the vanishing gradient problem that occurs in deep neural networks when processing long sequences.
   v. The algorithms excelled in dealing with complex sentence structures, which is crucial for crime-related text that often features multiple clauses/phrases. They could effectively handle such intricate structures by recursively applying the same neural network to each constituent part of the sentence.

#### 7.5.1.3 The Algorithms that Employ Decision Support-Based Approaches [87, 90]

Upon evaluating the experimental outcomes, it was observed that the algorithms employing traditional methods and incorporating gradient-boosted decision trees displayed a high level of competence in accurately classifying text, resulting in impressive precision scores. However, these algorithms demonstrate a tendency to overfit when employed on limited datasets. The improved performance can be attributed to their ability to retain instances with substantial gradients while disregarding those with minor gradients.

The boosted decision tree model showed a significant improvement over the decision tree model, suggesting that the boosting technique effectively reduces overall prediction errors. The higher precision of the boosted decision tree model indicates a lower rate of false positives (i.e., fewer instances where non-crimes were predicted as crimes). The model's higher recall suggests it is better at identifying actual crimes, reducing the number of missed crime predictions. Overall, the results suggest that the Boosted Decision Tree model is effective for crime prediction tasks.

#### 7.5.1.4 The Algorithm that Employs Naïve Bays-Based Approach [84]

By leveraging labeled data from a distribution, this algorithm was successful in estimating an initial model. Its inaccurate results were primarily caused by assuming feature non-interdependence, indicating a lack of consideration for the connections and interplay between various dataset features. Despite this constraint, the algorithm produced some useful outcomes.

The results suggest that the Naive Bayes algorithm can effectively predict certain types of crime with reasonable accuracy. However, performance varied significantly based on the crime type, with some crimes being more predictable than others based on the available features. The model's strengths lie in its simplicity and the probabilistic understanding of the features' relationships with crime occurrences.

#### 7.5.1.5 The Algorithms that Employ K-Nearest Neighbor and Kernal Density Estimation Neighbor-Based Approachs [93, 96]

These algorithms demonstrated satisfactory accuracy scores through the implementation of distance-based classification and normalization of training data. However, the presence of irrelevant features and inconsistent scaling of features posed challenges for the algorithms. Irrelevant features introduced noise and bias to the results, as they did not significantly impact the classification model. Inconsistent scaling of features led to inaccuracies since larger scaled features may dominate over others. In all datasets, as the cell size grows, the execution time exhibited an exponential decrease. Furthermore, when the cell size was small, the time expense was so significant that employing it in interactive systems is advised against.

#### 7.5.1.1 The Algorithms that Employ Pre-Trained-Based Approaches [36, 38]

Although these algorithms captured word contexts and analyzed important contextual information, their accuracy scores were only moderate. Also, their dependence on extensive training datasets limited their ability to perform well on larger corpora. The analysis of misclassification of the algorithms through the confusion matrix revealed the models' robustness in distinguishing between criminal and non-criminal classes accurately. This addressed potential class imbalance issues effectively. This balanced dataset composition ensured that the models' performance was not adversely affected by an imbalance in class representation.

### 7.5.2 Discussion of the Results of the Clustering Algorithms

#### 7.5.2.1 The Algorithms that Employ Partitioning-Based Approaches [108, 112]

These algorithms proved to be effective in utilizing local information to reduce the distance within clusters and increase it between different ones, ultimately aiding in the comprehension of the data. These algorithms could identify and group texts that shared similar content or themes and separate them from dissimilar texts. Moreover, these algorithms were more reliable when the number of centroids exceeded the possible number of clusters. The common neighbor-based algorithms were found to be efficient in processing large datasets, which makes them suitable for handling large amounts of text data. However, their accuracy was found to be lower than other methods when dealing with larger networks or handling overlapping clusters. This is because the algorithms are limited by their inability to handle complex relationships and interactions between texts present in such situations.

#### 7.5.2.2 The Algorithms that Employ Structure-Based Approaches [120, 123]

The results of the experiments showed that combining density-based or self-exciting point process techniques with clustering methods resulted in a significant improvement in performance. This integration enabled a more precise representation of the data, leading to better clustering outcomes. Moreover, using multiple types of context information simultaneously further enhanced the algorithms' effectiveness, particularly for non-linearly separable datasets. Nonetheless, the experiments also highlighted the sensitivity of the algorithms to parameter settings. Inaccurate parameter configurations considerably affected the accuracy of clustering results. Additionally, the algorithms were ineffective in accurately clustering datasets with varying densities.

#### 7.5.2.3 The Algorithm that Employ Density-Based Approach [130]

Based on the density of points rather than specific distance measures, this approach was found to be less susceptible to noise in the data according to the experimental results. It was able to detect clusters of any shape and size, including those that overlap, a feat that can be challenging for other clustering methods. Additionally, the approach could pinpoint outlier points that did not belong to any cluster, making it valuable for tasks such as anomaly detection or data cleansing.

### 7.5.3 Discussion of the Results of the Regression Algorithms

#### 7.5.3.1 The Algorithms that Employ Variable Independency-Based Approachs [133, 136]

*The Algorithms that Employ Linear Regression-based Approaches:* These algorithms could identify the most significant predictors of crime rates. They presented a quantitative measure of the association between crime rates and predictor variables, allowing for the identification of the strength and direction of the relationship. The algorithms provided fast and efficient crime rate predictions; however, they faced difficulties in accurately predicting crimes when the relationships were non-linear. This was due to the linear regression assumption that the relationship between the dependent variable (crime rates) and independent variables (predictors) was linear. Also, the approach was susceptible to overfitting, where the model fits the noise in the data rather than the underlying relationship between the variables, resulting in inadequate predictive performance on new data. The models showed positive responses to sampling, achieving performance that was either improved or on par with traditional logistic regression models under conditions of high event frequency. If biases or errors remain untreated in the data, they will be transferred to the model construction phase, leading to inaccurate models.

#### 7.5.3.2 The Algorithm that Employs Time Series-Based Approach [147]

Based on the results of the experiment, it was found that combining DNNs and other linear models improved the accuracy of the original single linear time series regression model. Furthermore, it was observed that integrating RNNs into a hybrid model, which combines the strengths of multiple models, resulted in a minor decrease in the RMSE. The inclusion of RNNs allowed the model to capture temporal dependencies and patterns in the data, leading to improved predictions. However, the approach did not perform well when applied to non-stationary data.

### 7.5.4 Discussion of the Results of the Social Network-Based Algorithms

#### 7.5.4.1 The Algorithms that Employ Path-Based-Based Approachs [150, 153]

Based on the experimental results, it was found that the algorithms utilizing shortest path-based approaches were effective in detecting crime in specific scenarios, particularly those where the crime data was clearly defined and the network was simple. These algorithms were capable of quickly identifying the most direct route between two points within the datasets, and accurately predicting crime hotspots. However, the experiments also revealed certain limitations of the shortest path-based algorithms, including the fact that they did not account for the dynamic nature of crime patterns, which can change over time and space, nor did they consider factors that could influence the likelihood of criminal activity in specific areas. As a result, the accuracy of the shortest path-based algorithms was found to be limited in more complex scenarios, such as those involving large datasets with a high density of crime incidents. According to the experimental findings, the algorithms utilizing the minimum spanning tree approach proved to be more efficient than those using the shortest path approach. This was due to their ability to consider the relative importance or influence of various nodes or areas within the network, thereby providing a more nuanced comprehension of crime patterns. Nonetheless, these algorithms based on the minimum spanning tree were insufficient in accounting for the dynamic nature of crime patterns or the influence of external factors on crime rates.

#### 7.5.4.2 The Algorithms that Employ Vertices Connections-Based Approachs [155, 160]

These algorithms demonstrated high accuracy in detecting crime patterns within well-established and well-connected networks, as evidenced by the experimental results. However, their effectiveness decreased significantly when trained on biased or incomplete data, making it difficult for them to identify important nodes and relationships within highly decentralized or poorly connected networks. The study revealed that the accuracy of k-core-based algorithms relied heavily on the quality and completeness of the network data used. When the data was incomplete, false positives or false negatives occurred, negatively impacting the algorithms' performance. The algorithms couldn't accurately capture all relevant information and connections due to the complex/dynamic networks.

## 8 POTENTIAL FUTURE PERSPECTIVES FOR CRIME PREDICTION

### 8.1 Future Perspectives for Classification Approaches

1. Deep learning models, such as CNNs, RNNs, and GNNs, are highly effective in crime text classification. However, optimizing technical constraints such as layer depth, regularization, and network learning rate is necessary for optimal performance. Adversarial samples can significantly reduce the effectiveness of the model, and enhancing model robustness in their presence is a research challenge. Although DNNs excel in feature extraction and semantic mining, designing accurate models necessitates a deeper understanding of the underlying theories. Improving model performance and interpretability remains a challenge due to the absence of clear guidelines for optimization.
2. Deep learning techniques like CNNs and RNNs can be employed to extract relevant crime features from text data, which can then be classified using conventional clustering algorithms for integration with deep learning.
3. Domain knowledge can be incorporated to improve crime text classification by customizing the algorithm for subjects and crime topics' significance.
4. Techniques like data imputation and outlier detection can be employed to manage noisy and incomplete crime data typically found in real-world data, improving text classification algorithms' handling of such data.
5. The difficulty of understanding why documents is grouped together can be solved using interpretable classification algorithms that employ human-readable cluster descriptions.
6. There is a growing trend in developing classification algorithms that can merge information from multiple modalities to comprehend the relationship between different data types, particularly in multimedia analysis.
7. Feature selection is critical to the success of crime prediction using classification. Future perspectives for crime prediction using classification should focus on developing more effective feature selection techniques that can identify the most relevant features for prediction.
8. In the realm of foreign exchange prediction, past research has predominantly concentrated on traditional econometric models, focusing on macroeconomic indicators, such as interest rates, inflation rates, and economic growth metrics. The shift to data-driven methods in foreign exchange prediction emphasizes the use of ML and AI, including historical data, sentiment analysis, and pattern recognition. Deep learning models like LSTMs and CNNs are now key for identifying market patterns over time. The trend is towards integrating big data for immediate analysis and combining traditional econometric methods with advanced ML to enhance the accuracy and speed of predicting forex rates.

### 8.2 Future Perspectives for Clustering Approaches

1. Continuous learning is a form of machine learning that enables a model to learn from a continuous stream of data. In the case of crime text clustering, it can enhance the model's ability to adapt to new language patterns, resulting in improved accuracy and adaptability.
2. Interpretability is crucial for creating trust when using crime text. Further research into interpretable models for crime text clustering can provide insights into prediction mechanisms and increase transparency.
3. Crime prediction using clustering can be integrated with various data sources such as social media, surveillance cameras, mobile devices, and other sensors, to improve the accuracy of predictions.
4. Clustering can be used to identify high-risk areas for crime, which helps to focus on those areas. This is known as predictive policing and has already been used in some cities.
5. Clustering can be used to identify crime patterns across different jurisdictions, which could facilitate collaboration between law enforcement agencies. By sharing information and resources, agencies could more effectively prevent crime and catch criminals.

### 8.3 Future Perspectives for Regression Approaches

1. Deep learning techniques, such as CNNs and RNNs, can be integrated with text regression analysis to enhance accuracy and scalability. By learning complex features from text data, deep learning algorithms can handle unstructured and structured text data with greater precision.
2. To improve crime text regression analysis, domain-specific knowledge can be incorporated into the analysis. For instance, in healthcare, medical terminology and knowledge were utilized to enhance the prediction of disease severity or treatment outcomes from patient notes.
3. Due to the noisy and incomplete nature of crime text data, it can be challenging to construct accurate regression models. Therefore, future research can focus on developing techniques to manage these problems, such as outlier detection, data cleaning, and data imputation.
4. Crime text data is often accompanied by other types of data, such as images, audio, or video. To better predict numerical values, multi-modal regression analysis can be employed to combine information from various modalities.
5. To facilitate the interpretation of crime text regression analysis, it is essential to develop explainable models that provide insights into the factors driving the predicted numerical values. Feature importance analysis and visualization are among the techniques that can be utilized.

### 8.4 Future Perspectives for Social Network Approaches

Crime prediction utilizing social network analysis emerges as a novel, evolving domain with substantial potential to enrich crime prevention and law enforcement tactics. Enhanced perspectives include:

1. The application of social network analysis extends to an array of data sources beyond mere social media platforms. Incorporating data from social networks, mobile devices, and additional sources enables more comprehensive and precise criminal activity predictions.
2. The sophistication of machine learning algorithms is on the rise, facilitating finer accuracy in predictions derived from social network data. The employment of artificial intelligence and deep learning techniques augments the predictive capability and enriches insights into criminal behaviors.
3. The amalgamation of social network analysis with Geographic Information Systems offers a fuller depiction of criminal activities, empowering law enforcement to pinpoint and preempt crime in targeted geographic locales.
4. With the growing use of social networks for cybercrime, including phishing, identity theft, and fraud, employing social network analysis for the identification and monitoring of such activities equips law enforcement with effective tools against cybercrime.

## 9 CONCLUSIONS

This survey paper presented a comprehensive examination of crime prediction, focusing on the most recent advancements in machine learning and statistical modeling techniques. We explored a range of modeling approaches, including regression, classification, clustering, social network, and deep learning, with the goal of predicting criminal activities. An assessment of each technique's strengths, limitations, and crucial considerations for model selection is provided. We also conducted an extensive review of the current landscape of crime prediction research, emphasizing noteworthy strategies and obstacles within the field. We introduced a structured taxonomy consisting of four tiers, commencing with the methodology category, and concluding with the methodology sub-techniques.

In addition to presenting a comprehensive taxonomy for crime predication algorithms, this survey undertook empirical and experimental evaluations to assess the efficacy of these diverse approaches. Through rigorous experimentation, the study compared and ranked various algorithmic categories and techniques. This includes evaluating different sub-techniques using the same technique, different techniques utilizing the same methodology sub-category, different methodology sub-categories within the same category, as well as different methodology categories. Our methodological taxonomy, empirical assessments, and comparative experiments collectively enhance researchers' understanding of crime prediction algorithms. This comprehensive perspective enables researchers to make informed choices when selecting an algorithm or technique to address their specific research inquiry or problem.

Recursive neural network approaches proved by our experimental results to be the most successful among the various methods utilizing classification-based techniques. Similarly, Spatiotemporal approaches demonstrated superior performance compared to other methods within the clustering-based techniques. Time series regression approaches emerged as the top performers among the different methods employing regression-based techniques. Within the realm of social network-based techniques, the methods employing k-core analysis yielded the best results.

**REFERENCES**

[1] Mohler, G. "Marked point process hotspot maps for homicide and gun crime prediction in Chicago," *International Journal of Forecasting*, Volume 30(3) 491-497, 2014.

[2] A. Iriberri and G. Leroy, "Natural Language Processing and e-Government: Extracting Reusable Crime Report Information," *2007 IEEE International Conference on Information Reuse and Integration*, Las Vegas, NV, USA, pp. 221-226, 2007.

[3] Curry A, Latkin C, Davey-Rothwell M. "Pathways to depression: the impact of neighborhood violent crime on inner-city residents in Baltimore, Maryland, USA". *Social Science & Medicine*. 67(1):23-30, 2008.

[4] Harper, W. R., & Harris, D. H. "The Application of Link Analysis to Police Intelligence". *Human Factors*, 17(2), pp. 157-164, 975.

[5] McAndrew, D. "The structural analysis of criminal networks", in: D. Canter, L. Alison (Eds.), *The Social Psychology of Crime: Groups, Teams, and Networks, Offender Profiling Series*, Aldershot, Dartmouth, 1999.

[6] V. Krebs. "Mapping Networks of Terrorist Cells," *Connections*, 24 (3), pp. 43–52, 2002.

[7] W.F. Coady, "Automated link analysis: artificial intelligence-based tool for investigators", *Police Chief*, 52(9):22-23, 1985.

[8] Sparrow, K. "The application of network analysis to criminal intelligence: an assessment of the prospects," *Social Network*, 13:251-274, 1991.

[9] Xia, L., Huang, C., Xu, Y., Dai, P., Bo, L., Zhang, X., and Chen, T. "Spatial-Temporal Sequential Hypergraph Network for Crime Prediction with Dynamic Multiplex Relation Learning". IJCAI, 1631-1637, 2021.

[10] Kadar, C., Maculan, R., and Feuerriegel, S. "Public decision support for low population density areas: An imbalance-aware hyper-ensemble for spatio-temporal crime prediction", *Decision Support Systems*, Volume 119, pp. 107-117, 2019.

[11] Zhang, Y., Siriaraya, P, Kawai, Y., and Jatowt, A. "Predicting time and location of future crimes with recommendation methods", *Knowledge-Based Systems*, Volume 210, 2020.

[12] Zhao, X., Fan, W., Liu, H. and Tang, J. "Multi-type Urban Crime Prediction". In the *AAAI Conference on Artificial Intelligence*, Vol. 36. 4388—439, 2022.

[13] Thomas, A., Sobhana, N. "A survey on crime analysis and prediction", *Materials Today: Proceedings*, 58(1):310-315, 2022.

[14] Yang, H., Huang, C., Liang, H., Ding, W., and Li, X. "A Survey of Property Crime Incident Links and Their Discovery Techniques". In *2021 International Conference on Mechanical, Aerospace and Automotive Engineering (CMAAE 2021)*. NY, USA, pp. 93–97, 2022.

[15] Kawthalkar, I., Jadhav, S., Jain, D. and Nimkar, A. "A Survey of Predictive Crime Mapping Techniques for Smart Cities," In *2020 National Conference on Emerging Trends on Sustainable Technology and Engineering Applications (NCETSTEA)*, Durgapur, India, pp. 1-6, 2020.

[16] Kounadi, O. Ristea, A., Araujo, A. Leitner, M. "A systematic review on spatial crime forecasting". *Crime Science*, 9(1), 2020.

[17] Saravanan, P., et al. "Survey on Crime Analysis and Prediction Using Data Mining and Machine Learning Techniques". In: Zhou, N., Hemamalini, S. (eds) Advances in Smart Grid Technology. Lecture Notes in Elect. Eng., v. 688. Springer, Singapore, 2021

[18] Prabakaran, S. and Mitra, S. "Survey of Analysis of Crime Detection Techniques Using Data Mining and Machine Learning," *Journal of Physics: Conference Series*, 1000(1), 2018.

[19] Ramesh, P. and Maheswari, D. "Survey of cyber crime activities and preventive measures". In *the Second International Conference on Computational Science, Engineering and Information Technology (CCSEIT '12)*., New York, NY, USA, pp. 301–305, 2012.

[20] Alsaqabi, A., Aldhubayi, F. and Albahli, S. "Using Machine Learning for Prediction of Factors Affecting Crimes in Saudi Arabia". In *the International Conference on Big Data Engineering (BDE)*, New York, NY, USA, pp. 57–62, 2019.

[21] Belesiotis, A., Papadakis, G. and Skoutas, D. "Analyzing and Predicting Spatial Crime Distribution Using Crowdsourced and Open Data". *ACM Transactions on Spatial Algorithms Syst.* 3, 4, Article 12, 2018.

[22] Bshayer S. et al. 2020. "A Comparative Study of Decision Tree and Naive Bayes Machine Learning Model for Crime Category Prediction in Chicago". In *the 2020 the 6th International Conference on Computing and Data Engineering (ICCDE)*. New York, NY, USA, pp. 34–38, 2020.

[23] Zhuo, E. and Libed, J. "Analysis of Crime Rates in Rizal Province using Crime Forecasting Models". In *the 2020 the 3rd International Conference on Computers in Management and Business (ICCMB 2020)*, New York, NY, USA, pp. 64-69, 2020.

[24] Duan, L., Hu, T., Cheng, E., Zhu, J., and Gao, C. "Deep convolutional neural networks for spatiotemporal crime prediction," in *the International Conference on Information and Knowledge Engineering (IKE)*, Las Vegas, NV, USA, pp. 61–67, 2017.

[25] Fu, K., Chen, Z., and Lu, C. "StreetNet: preference learning with convolutional neural network on urban crime perception". In *the 26th ACM International Conference on Advances in Geographic Information Systems (SIGSPATIAL '18)*, New York, NY, USA, pp. 269–278, 2018.

[26] Wei, Y., Liang, W., Wang, Y., "CrimeSTC: A Deep Spatiotemporal-Categorical Network for Citywide Crime Prediction". In *the 2020 3rd International Conference on Computational Intelligence and Intelligent Systems (CIIS '20)*. New York, NY, USA, 75–79, 2020.

[27] Onan, A. "Sentiment analysis on product reviews based on weighted word embeddings and deep neural networks", *Concurrency and Computation: Practice and Experience*, vol. 33, no. 23, Art. no. e5909, 2021.

[28] Patel, C., Bhatt, D., Sharma, U., Patel, R., Pandya, S., Modi, K., Cholli, N., Patel, A., Bhatt, U.. Khan, M.A., et al. "DBGC: Dimension-Based Generic Convolution Block for Object Recognition". *Sensors*, 22, 1780, 2022.

[29] Hara, K., Kataoka, H., and Satoh, Y. "Learning Spatiotemporal Features with 3D Residual Networks for Action Recognition," In *the IEEE International Conference on Computer Vision Workshops*, Venice, Italy, pp. 3154–3160, 2017.

[30] Jan, A. and Khan, G. "Malicious Activity Detection In Safe City Environment," *International Conference on Artificial Intelligence (ICAI)*, Islamabad, Pakistan, pp. 170-174, 2021.

[31] Matereke, T., Nyirenda, C. and Ghaziasgar, M. "A Performance Evaluation of 3D Deep Learning Algorithms for Crime Classification," In *the IEEE AFRICON*, Arusha, Tanzania, pp. 1-6, 2021.

[32] Krungklang, W., Sinthupinyo, S. "An Analysis of Natural Language Text Relating to Thai Criminal Law," In *the 2020 12th International Conference on Electronics, Computers and Artificial Intelligence (ECAI)*, Bucharest, Romania, pp. 1-6, 2020.

[33] Patel, J., Sisodiya, R., Doshi, N. and Mishra, S. "A CNN-BiLSTM based Approach for Detection of SQL Injection Attacks," In *the International Conference on Computational Intelligence and Knowledge Economy (ICCIKE)*, Dubai, UAE, pp. 378-383, 2021.

[34] Nidhi, A., C, D. and Sowjanya K, "Crime Forecasting : A Theoretical Approach," In *the IEEE 7th International Conference on Recent Advances and Innovations in Engineering (ICRAIE)*, Mangalore, India, pp. 37-41, 2022.

[35] Saha, R., Naskar, A., Dasgupta,T., and Dey, L. "A System for Analysis, Visualization and Retrieval of Crime Documents". In *the 7th ACM IKDD CoDS and 25th COMAD (CoDS COMAD 2020)*. New York, NY, USA, 317–321, 2020.

[36] Verma, H., Lotia, S. and Singh, A. "Convolutional Neural Network Based Criminal Detection," In *the IEEE Region 10 Conference (TENCON)*, Osaka, Japan, pp. 1124-1129, 2020.

[37] Barathi, B., Aadesh, P., Balajee, R., Balaji, M. "Suspicious action and behavior detection using CNN", *International Journal of Computer Science and Mobile Computing*, 9(5):51-59, 2020.

[38] Sandagiri, S., Kumara, B. and Kuhaneswaran, B. "ANN Based Crime Detection and Prediction using Twitter Posts and Weather Data," In *the International Conference on Data Analytics for Business and Industry (ICDABI)*, Sakheer, Bahrain, pp. 1-5, 2020.

[39] Alkhatib, A., Mushtaq, M., Ghauch, H., Danger, J. "CAN-BERT do it? Controller Area Network Intrusion Detection System based on BERT Language Model", In *the 2022 IEEE/ACS 19th International Conference on Computer Systems and Applications (AICCSA)*, PP. 1-8, 2022.

[40] Rifat, N., Ahsan, M., Chowdhury,M. and Gomes, R. "BERT Against Social Engineering Attack: Phishing Text Detection," *IEEE Intern. Conference on Electro Inform. Tech. (eIT)*, Mankato, MN, USA, pp. 1-6, 2022.

[41] Sandagiri, S., Kumara, B. and Kuhaneswaran, B. "Deep Neural Network-Based Approach to Identify the Crime Related Twitter Posts," In *the International Conference on Decision Aid Sciences and Application (DASA)*, Sakheer, Bahrain, pp. 1000-1004, 2020.

[42] Onan, A. "Hierarchical graph-based text classification framework with contextual node embedding and BERT-based dynamic fusion", *Journal of King Saud University - Computer and Information Sciences*, Volume 35, Issue 7, 2023.

[43] Bhatt, D., Patel, C., Talsania, H., Patel, J., Vaghela, R., Pandya, S., Modi, K., Ghayvat, H. "CNN Variants for Computer Vision: History, Architecture, Application, Challenges and Future Scope". *Electronics* ,10, 2470, 2021.

[44] Zhuang, Y., Almeida, M, Morabito, M., and Ding, W. "Crime hot spot forecasting: A recurrent model with spatial and temporal information," In *the 2017 IEEE International Conference on Big Knowledge (ICBK)*, Hefei, China, 2017, pp. 143-150, 2017.

[45] Esquivel, N., Nicolis, O., Peralta, B. and Mateu, J. "Spatiotemporal Prediction of Baltimore Crime Events Using CLSTM Neural Networks," in *IEEE Access*, vol. 8, pp. 209101-209112, 2020.

[46] Wawrzyniak, Z. *et al.*, "Data-driven models in machine learning for crime prediction," In *the 26th International Conference on Systems Engineering (ICSEng)*, Sydney, NSW, Australia, pp. 1-8, 2018.

[47] Matereke, T., Nyirenda, C. and Ghaziasgar, M. "A Comparative Evaluation of Spatio Temporal Deep Learning Techniques for Crime Prediction," In *the IEEE AFRICON*, Arusha, Tanzania, pp. 1-6, 2021.

[48] Khoei, T., Hu, W. and Kaabouch, N. "Residual Convolutional Network for Detecting Attacks on Intrusion Detection Systems in Smart Grid," In the *IEEE International Conference on Electro Information Technology (eIT)*, Mankato, MN, USA, pp. 231-237, 2022.

[49] Z. Li, C. Huang, L. Xia, Y. Xu and J. Pei, "Spatial-Temporal Hypergraph Self-Supervised Learning for Crime Prediction," In the *2022 IEEE 38th International*

[50] Rayhan, Y. and Hashem, T. "AIST: An Interpretable Attention-Based Deep Learning Model for Crime Prediction". *ACM Transactions on Spatial Algorithms and Systems*. 9, 2, Article 13, June 2023.

[51] Manju, D., Seetha M, Sammulal, P., "Early action prediction using 3DCNN with LSTM and bidirectional LSTM" *Turkish Journal of Computer and Mathematics Education* 12(6):2275-2281, 2021.

[52] Yi, F., Yu, Z., Zhuang, F. and Guo, B. "Neural network based continuous conditional random field for fine-grained crime prediction". In *the 28th International Joint Conference on Artificial Intelligence (IJCAI)*, Macao, 2019.

[53] Yuan, Z., Zhou, X. and Yang, T. "Hetero-ConvLSTM: A Deep Learning Approach to Traffic Accident Prediction on Heterogeneous Spatiotemporal Data". In *the 24th ACM SIGKDD International Conference on Knowledge Discovery & Data Mining (KDD)*, London United Kingdom, 984–992, 2018.

[54] Mei, Y. and Li. F. "Predictability Comparison of Three Kinds of Robbery Crime Events Using LSTM". In *the 2019 2nd International Conference on Data Storage and Data Engineering (DSDE 2019)*, Jeju Island, South Korea, pp. 22–26, 2019.

[55] Siami-Namini, S., Tavakoli, N. and Namin, N. "The Performance of LSTM and BiLSTM in Forecasting Time Series," In *the IEEE International Conference on Big Data (Big Data)*, Los Angeles, CA, USA, pp. 3285-3292, 2019.

[56] Ren, F., Jiang. Z. and Liu, J. "A Bi-Directional LSTM Model with Attention for Malicious URL Detection," In the *Advanced Information Technology*, Electronic and *Automation Control Conference (IAEAC)*, Chengdu, China, pp. 300-305, 2019.

[57] Freitas, N., Silva, l., Macêdo, J. Vasconcelos, L. and Junior, F. "Crime Monitor: Monitoring Criminals from Trajectory Data," In *the 22nd IEEE International Conference on Mobile Data Management (MDM)*, Toronto, ON, Canada, pp. 225-228, 2021.

[58] Onan, A. "SRL-ACO: A text augmentation framework based on semantic role labeling and ant colony optimization", *Journal of King Saud University - Computer and Information Sciences*, Volume 35, Issue 7, 2023.

[59] Onan, A. "Bidirectional convolutional recurrent neural network architecture with group-wise enhancement mechanism for text sentiment classification", *Journal of King Saud University - Computer and Information Sciences*, 34(5):2098-2117, 2022.

[60] Onan, A. and Toçoğlu, M. A. "A Term Weighted Neural Language Model and Stacked Bidirectional LSTM Based Framework for Sarcasm Identification," in *IEEE Access*, vol. 9, pp. 7701-7722, 2021.

[61] Pareja, A., Domeniconi, G., Chen, J. and Ma, T. "EvolveGCN: evolving graph convolutional networks for dynamic graphs," in *the AAAI Conference on Artificial Intelligence*, 34(4):5363–5370, February 2020.

[62] Stec A, Klabjan D. "Forecasting crime with deep learning". *arXiv* preprint arXiv:180601486, 2018.

[63] Kabir, M., Safir, F., Shahen, S., Maua, J., Awlad, I. "Human Abnormality Classification Using Combined CNN-RNN Approach," In the *IEEE 17th International Conference on Smart Communities: Improving Quality of Life Using ICT, IoT & AI*, Charlotte, USA, pp. 204-208, 2020.

[64] Huang, C., Zhang, J., Zheng, Y. and Chawla, N. "DeepCrime: Attentive Hierarchical Recurrent Networks for Crime Prediction". In the *27th ACM International Conference on Information and Knowledge Management (CIKM '18)*, Torino Italy, pp. 1423–1432, 2018.

[65] Onan, A. "Mining opinions from instructor evaluation reviews: a deep learning approach", *Computer Applications in Engineering Education*, vol. 28, no. 1, pp. 117–138, 2020.

[66] Wang, C., Lin, Z., Yang, X., Sun, J., Yue, M., Shahabi, C. "HAGEN: Homophily-Aware Graph Convolutional Recurrent Network for Crime Forecasting". *AAAI*, 36, 4193-4200, 2022

[67] Xiao, J., Ye, H., He, X., Zhang, H., Wu,F. and Chua, T. "Attentional factorization machines: learning the weight of feature interactions via attention networks". In *the 26th International Joint Conference on Artificial Intelligence (IJCAI'17)*. Melbourne Australia, pp. 3119–3125, 2017.

[68] Hu, K., Li, L., Liu, J. and Sun, D. "DuroNet: A Dual-robust Enhanced Spatiotemporal Learning Network for Urban Crime Prediction". *ACM Transactions* on *Internet Technology*. 21, 1, Article 24, February 2021.

[69] Yu, Y., Jiao, L., Zhou, N., Zhang, L. and Yin, H. "Enhanced factorization machine via neural pairwise ranking and attention networks". *Pattern Recognition Letters* 140 (2020), 348–357, 2020.

[70] Mascorro, G. *et al.* "Suspicious Behavior Detection on Shoplifting Cases for Crime Prevention by Using 3D Convolutional Neural Networks", *ArXiv*, 2020.

[71] Onan, A. "GTR-GA: Harnessing the power of graph-based neural networks and genetic algorithms for text augmentation", *Expert Systems with Applications*, Volume 232, 2023.

[72] S. Zhao, R. Liu, B. Cheng and D. Zhao, "Classification-Labeled Continuousization and Multi-Domain Spatio-Temporal Fusion for Fine-Grained Urban Crime Prediction," In *IEEE Transactions on Knowledge and Data Engineering*, 35(7):6725-6738, 1 July 2023

[73] Lin, Y., Yen, M. and Yu, L. "Grid-based crime prediction using geographical features," *ISPRS International Journal of Geo-Information*, 7(8):298, 2018.

[74] Chun, S., Paturu, V., Yuan, S., Pathak, R., Atluri, V. "Crime Prediction Model using Deep Neural Networks". In *the 20th Annual International Conference on Digital Government Research (dg.o)*. pp. 512–514, 2019.

[75] Guo, H., Tang, R., Ye, Y., Li, , Spain, pp. 1725–1731, 2017.

[76] Vomfell, L., Härdle, W., Lessmann, S. "Improving crime count forecasts using Twitter and taxi data", A decision support system (DSS), 113:73-85, 2018.

[77] Ingilevich, V. and Ivanov, S. "Crime rate prediction in the urban environment using social factors", *Procedia Computer Science*, 136, pp. 472–478, 2018.

[78] Asad, A., Mansur, R., Zawad, S. "Analysis of Malware Prediction Based on Infection Rate Using Machine Learning Techniques," In *the IEEE Region 10 Symposium (TENSYMP)*, Dhaka, Bangladesh, pp. 706-709, 2020.

[79] Yu, C., Ding, W., Chen, P., Morabito, M. "Crime forecasting using Spatiotemporal pattern with ensemble learning". In *the18th Pacific-Asia Conference on Knowledge Discovery and Data Mining*, Tainan, Taiwan, pp. 174–185, 2014.

[80] Hossain, S., Abtahee, A., Kashem, I., Hoque, M. and Sarker, I. "Crime prediction using Spatiotemporal data," *arXiv:2003.09322*, 2020.

[81] Yuki, J., Sakib, M., Zamal, Z., Habibullah, K., Das, A. "Predicting Crime Using Time and Location Data". In the *7th International Conference on Computer and Communications Management (ICCCM )*, Bangkok, Thailand, pp. 124–128, 2019.

[82] Al-Ghushami, A., Syed, D., Sessa, J., Zainab, A. "Intelligent Automation of Crime Prediction using Data Mining," In *the IEEE 31st Intern. Symposium on Industrial Electronics (ISIE)*, Anchorage, AK, USA, pp. 245-252, 2022.

[83] Onan, A. "Sentiment analysis on massive open online course evaluations: A text mining and deep learning approach". *Computer Applications in Engineering Education*, *29*(3), 572–589, 2021.

[84] Kumar R, and Nagpal, B. "Analysis and prediction of crime patterns using big data," *International Journal of Information Technology*, 11(4):799-805, 2019.

[85] Jangra, M. and Kalsi, S. "Naïve Bayes Approach for the Crime Prediction in Data Mining", *International Journal of Computer Applications*, 178 (4), 2019.

[86] Phua, C., Alahakoon, D., Lee, V. "Minority Report in Fraud Detection: Classification of Skewed Data", *Sigkdd Explorations*, 6(1),51-5, 2004.

[87] Ivan, M., Ahishakiye, E., Taremwa, D. "Crime Prediction Using Decision Tree (J48) Classification Algorithm", *International Journal of Computer and Information Technology, 6(3),* 2017.

[88] Kiani, R., Mahdavi, S. and Keshavarzi, A. "Analysis and Prediction of Crimes by Clustering and Classification", *International Journal of Advanced Research in Artificial Intelligence*, 4(8), 2015.

[89] Aldossari, B., Alqahtani, F., Alshahrani, N., Alzamanan, R. "A Comparative Study of Decision Tree and Naive Bayes Machine Learning Model for Crime Category Prediction in Chicago", In *the 6th International Conference on Computing and Data Engineering (ICCDE)*, Sanya China. pp. 34–38, 2020.

[90] Kim, S., Joshi P, Kalsi, PS., Taheri, P. "Crime analysis through machine learning". In *the IEEE 9th annual information technology electronics and mobile communication conference*. Vancouver 1-3 November 2018.

[91] Hu, Y., Wang, F., Guin, C., Zhu, H. "A Spatiotemporal kernel density estimation framework for predictive crime hotspot mapping and evaluation", *Applied*

*Geography*, Volume 99, Pages 89-97, 2018.

[92] Alsirhani, A., Sampalli, S., Bodorik, P. "DDoS Detection System: Utilizing Gradient Boosting Algorithm and Apache Spark," In *the IEEE Canadian Conf. Elect. & Comp, Engin. (CCECE)*, Quebec, QC, Canada, pp. 1-6, 2018.

[93] Kumar, A. Verma, A., Shinde, G., Sukhdeve, Y. and Lal, N. "Crime Prediction Using K-Nearest Neighboring Algorithm," In t*he International Conference on Emerging Trends in Information Technology and Engineering (ic-ETITE)*, Vellore, India, pp. 1-4, 2020.

[94] Llaha, O. "Crime Analysis and Prediction using Machine Learning," In *the 43rd International Convention on Information, Communication and Electronic Technology (MIPRO)*, Opatija, Croatia, pp. 496-501, 2020.

[95] Tayebi, A., Gla, U., Brantingham, P. "Learning where to inspect: Location learning for crime prediction." In *the IEEE International Conference on Intelligence and Security Informatics (ISI)*, Baltimore, MD, USA, pp. 25-30, 2015.

[96] Neto, J., Santos, E and Vidal, C. "MSKDE - Using Marching Squares to Quickly Make High Quality Crime Hotspot Maps," In *the 29th SIBGRAPI Conference on Graphics, Patterns and Images (SIBGRAPI)*, Sao Paulo, Brazil, pp. 305-312, 2016.

[97] Yang, D., et al. "CrimeTelescope: crime hotspot prediction based on urban and social media data fusion". *World Wide Web, volume* 21, 5, pp. 1323–1347, 2018.

[98] Wang, X., Gerber, MS., Brown, DE. "Auto matic crime prediction using events extracted from twitter posts". In *Social Computing, Behavioral Cultural Modeling and Prediction*, pp. 231 238. Springer 514, 2012.

[99] Al Boni, M. and Gerber, M. "Automatic Optimization of Localized Kernel Density Estimation for Hotspot Policing," In *the 15th IEEE Inter. Conf. Machine Learning & Applications (ICMLA)*, Anaheim, CA, USA, pp. 32-38, 2016.

[100] Gerber, M. "Predicting crime using Twitter and kernel density estimation". *Decision Support Systems* 61, pp. 115–125, 2014.

[101] Safat, W., Asghar, S. and Gillani, S. "Empirical Analysis for Crime Prediction and Forecasting Using Machine Learning and Deep Learning Techniques," in *IEEE Access*, vol. 9, pp. 70080-70094, 2021.

[102] Alves, L., Ribeiro, H. and Rodrigues, F. "Crime prediction through urban metrics and statistical learning," *Statistical Mechanics* and *its Applications.*, vol. 505, pp. 435–443, 2018.

[103] Arora, T., Sharma, M. and Khatri, S. "Detection of Cyber Crime on Social Media using Random Forest Algorithm," In the *international Conference on Power Energy, Environment and Intelligent Control (PEEIC)*, Greater Noida, India, pp. 47-51, 2019.

[104] Schelter, S., Grafberger, S. and Dunning, T. "HedgeCut: Maintaining Randomised Trees for Low-Latency Machine Unlearning". In *the International Conference on Management of Data (SIGMOD '21)*, Virtual Event China, pp. 1545–1557, 2021.

[105] Araujo, A., Cacho, N., Bezerra, L., Vieira, C. and Borges, J. "Towards a crime hotspot detection framework for patrol planning,'' In *the. 20th* Annual *International Conference* on *High Performance Computing*, Hainan, China, pp. 1256-1263, 2018.

[106] Beitollahi, H., Sharif, D. and Fazeli, D. "Application Layer DDoS Attack Detection Using Cuckoo Search Algorithm-Trained Radial Basis Function," in *IEEE Access*, vol. 10, pp. 63844-63854, 2022.

[107] Sphamandla, S. May, Omowunmi, E. Isafiade, and Olasupo, O. Ajayi. "Hybridizing Extremely Randomized Trees with Bootstrap Aggregation for Crime Prediction". In *the International Conference on Artificial Intelligence and Pattern Recognition (AIPR)*, Xiamen, China, pp. 536–541, 2021.

[108] Iqbal, F., Fung, B., Debbabi, M., Batool, R. and Marrington, A. "Wordnet-Based Criminal Networks Mining for Cybercrime Investigation," in *IEEE Access*, vol. 7, pp. 22740-22755, 2019.

[109] Boni, M. and Gerber, M. "Area-Specific Crime Prediction Models," In the *2016 15th IEEE International Conference on Machine Learning and Applications (ICMLA)*, Anaheim, CA, USA, pp. 671-676, 2016.

[110] Sivaranjani, S., Sivakumari, S. and Aasha, M. "Crime prediction and forecasting in Tamilnadu using clustering approaches," In *the International Conference on Emerging Trends in Engineering and Technology*, Kollam, India, pp. 1-6, 2016.

[111] Xu, J., Chen, H. "CrimeNet explorer: a framework for criminal network knowledge discovery". *ACM Transactions on Information Systems*, 23(2):201–226, 2005.

[112] Hajela, G., Chawla, M., Rasool, A. "A Clustering Based Hotspot Identification Approach For Crime Prediction", *Procedia Computer Science*, 167:1462-1470, 2020.

[113] Tayal, D.K., Jain, A., Arora, S., Agarwal, S., Gupta, T. and Tyagi, N., "Crime detection and criminal identification in India using data mining techniques". *AI & society*, *30*(1), pp.117-127, 2015

[114] Ch., Y., Ding, W., Chen, P., Morabito, M. " Crime Forecasting Using Spatiotemporal Pattern with Ensemble Learning". In: *Tseng, V.S., Ho, T.B., Zhou, ZH., Chen, A.L.P., Kao, HY. (eds) Advances in Knowledge Discovery and Data Mining. PAKDD 2014. Lecture Notes in Computer Science*, vol 8444. Springer, Cham, 2014.

[115] Shrivastav, A., Ekata "Applicability of Soft computing technique for Crime Forecasting A Preliminary Investigation." In the *International Journal of Computer Science & Engineering Technology*, pp 415-421, 2012.

[116] Zhao, X and Tang, J. "Modeling Temporal-Spatial Correlations for Crime Prediction". In *the 2017 ACM on Conference on Information and Knowledge Management (CIKM '17)*, Singapore Singapore, pp. 497–506, 2017.

[117] Huang, Y., Li, C. and Jeng, C. "Mining location-based social networks for criminal activity prediction," In *the 24th Wireless and Optical Communication Conference (WOCC)*, Taipei, Taiwan, pp. 185-189, 2015.

[118] Malathi, A., Baboo, S. "An enhanced algorithm to predict future crime using data mining". *International Journal of Computer Applications* (**IJCA**), 21(1):16, 2011.

[119] Aryal, A. and Wang, S. "SparkSNN: A density-based clustering algorithm on spark," In *the 2018 IEEE 3rd International Conference on Big Data Analysis (ICBDA)*, Shanghai, China, pp. 433-437, 2018.

[120] Kim, Y., Kim, G., Lee, Y., and Jang, K. "Bandwidth Selection of Kernel Density Estimation for GIS-based Crime Occurrence Map Visualization," In the *International Conference on Information and Communication Techno. Convergence*, Jeju, Korea, pp. 1705-1708, 2020.

[121] Hu, Y. et al. "Crimes Prediction Using Spatiotemporal Data and Kernel Density Estimation", In t*he* 2019 *Asia Pacific Conference on Research in Industrial and Systems Engineering (APCoRISE)*, Depok, Indonesia, pp. 1-6, 2019.

[122] Clougherty, E., Clougherty, J., Liu, X., Brown, D. "Spatial and temporal analysis of sex crimes in Charlottesville, Virginia," In *the System and Information Engineering Design Symposium*, Charlottesville, VA, USA, pp. 69-74, 2015.

[123] Catlett, C., Cesario, E., Talia, D., Vinci, A., "Spatiotemporal crime predictions in smart cities: A data-driven approach and experiments", *Pervasive and Mobile Computing*, Volume 53, 2019.

[124] Aryal, A. and Wang, S. "Discovery of patterns in Spatiotemporal data using clustering techniques," In *the 2017 2nd International Conference on Image, Vision and Computing (ICIVC)*, Chengdu, pp. 990-995, 2017.

[125] Mohler, G., Short, M., Brantingham, P., Schoenberg, F. & Tita, G. "Self-Exciting Point Process Modeling of Crime", *Journal of the American Statistical Association*, 106:493, pp.100-108, 2011.

[126] Jaiswal, B., Chandra, B., and Paul, K. "Geodetic Distance and Dynamic Outlier Exclusion in EM Optimization of Self Exciting Point Process for Homicide Prediction in Chicago," In *the International Congress on Advanced Applied Informatics*, Kitakyushu, Japan, pp. 534-541, 2020.

[127] Wu, X., Huang, C., Zhang, C. and Chawla, N. "Hierarchically Structured Transformer Networks for Fine-Grained Spatial Event Forecasting". In *the Web Conference 2020 (WWW '20)*, Taipei Taiwan, pp. 2320–2330, 2020.

[128] Yu, C., Ding, W., Morabito, M. and Chen, P. "Hierarchical Spatiotemporal Pattern Discovery and Predictive Modeling," In *IEEE Transactions on Knowledge and Data Engineering*, 28(4):979-993, 2016.

[129] Wang, S., Li, X., Cai, Y. and Tian, J. "Spatial and temporal distribution and statistic method applied in crime events analysis," In *the 19th International Conference on Geoinformatics*, Shanghai, China, pp. 1-6, 2011.

[130] Butt, U. *et al.*, "Spatiotemporal Crime Predictions by Leveraging Artificial Intelligence for Citizens Security in Smart Cities," in *IEEE Access*, vol. 9, pp. 47516-47529, 2021.

[131] Sravani, T. and Suguna, M. "Comparative Analysis Of Crime Hotspot Detection And Prediction Using Convolutional Neural Network Over Support Vector Machine with Engineered Spatial Features Towards Increase in Classifier Accuracy," In the *International Conference on Business Analytics for Technology & Security (ICBATS)*, Dubai, UAE, pp. 1-5, 2022.

[132] Baqir, A., Rehman, S., Malik, S., Mustafa, F. and Ahmad, U. "Evaluating the Performance of Hierarchical Clustering algorithms to Detect Spatiotemporal Crime Hot-Spots," *International Conference on Computing, Mathematics, Engineering, and Technology.*, Sukkur, Pakistan, pp. 1-5, 2020.

[133] Zhang, Y., & Trubey, P. "Machine learning and sampling scheme: An empirical study of money laundering detection". *Computational Economics, 54(3)*, 1043-1063, 2019.

[134] Tabedzki, C. and Thirumalaiswamy, A. and van Vliet, P. and Sun, S. "Yo Home to Bel-Air: Predicting Crime on The Streets of Philadelphia" *University of Pennsylvania*, CIS 520, 2018.

[135] Chang, M., Yih, W. and Meek, C. "Partitioned logistic regression for spam filtering", In *the 14th ACM SIGKDD international conference on Knowledge discovery and data mining (KDD '08)*, Las Vegas Nevada USA, pp. 97–105, 2008.

[136] Alotaibi, N. "Cyber Bullying and the Expected Consequences on the Students' Academic Achievement," in *IEEE Access*, vol. 7, pp. 153417-153431, 2019.

[137] Misyrlis, M., Cheung, C., Srivastava, A., Kannan, R. and Prasanna, V. "Spatiotemporal Modeling of Criminal Activity". In *the International Workshop on Social Sensing (SocialSens'17)*, pp. 3–8, 2017.

[138] Gera, P. and Vohra, R. "Predicting Future Trends in City Crime using Linear Regression", *IJCSMS (International Journal of Computer Science & Management Studies)*, 14(7):58–64, 2014.

[139] McClendon L., Meghanathan N. "Using machine learning algorithms to analyze crime data". *Mach Lear Appl Int J* 2(1):1–12, 2015.

[140] Badal-Valero, E., Alvarez-Jareno, A., & Pavia, M. "Combining Benford's Law and machine learning to detect money laundering". *An actual Spanish court case. Forensic Science International*, 282, pp. 24–34, 2018.

[141] Bharati, A., Sarvanaguru, RAK. "Crime prediction and analysis using machine learning". *International Research Journal of Engineering and Technology*, (9):1037–1042, 2018.

[142] Matijosaitiene, I., Zhao, P., Jaume, S. "Prediction of Hourly Effect of Land Use on Crime", *International Journal of Geo-Information*, 8, 16, 2018.

[143] Rummens, A., Hardyns, W., Pauwels, L. "The use of predictive analysis in spatiotemporal crime forecasting: building and testing a model in an urban context". Applied Geography, 86:255–261. 2007.

[144] Payne, J and Morgan, A. "COVID-19 and violent crime: A comparison of recorded offence rates and dynamic forecasts (ARIMA) for march 2020 in Queensland, Australia," *Tech. Rep.*, 2020.

[145] Dash, S., Safro, I. and Srinivasamurthy, R. "Spatiotemporal prediction of crimes using network analytic approach," *IEEE International Conference on Big Data (Big Data)*, Seattle, WA, USA, pp. 1912-1917, 2018.

[146] Yi, F., Yu, z., Zhuang, f., Zhang, x., and Xiong, H. "An Integrated Model for Crime Prediction Using Temporal and Spatial Factors". In the *2018 IEEE International Conference on Data Mining (ICDM)*, Singapore, 2018, pp. 1386-1391, 2018.

[147] Garcia-Zanabria, G et al. "CriPAV: street-level crime patterns analysis and visualization". *IEEE Transactions on Visualization and Computer Graphics,* Vol. 28, No. 12, December 2022.

[148] Yuan,J., Cao, J., Xia, B. "Arresting Strategy Based on Dynamic Criminal Networks Changing over Time", *Discrete Dynamics in Nature and Society*, vol. 2013, Article ID 296729, 2012.

[149] Jameson L. Toole, Nathan Eagle, Joshua B. Plotkin. "Spatiotemporal correlations in criminal offense records". *ACM Transactions Intelligence System Technology 2, 4, Article 38*, 2011.

[150] Wiil, U., Gniadek, J. and Memon, N. "Measuring Link Importance in Terrorist Networks," In the *International Conference on Advances in Social Networks Analysis and Mining*, Odense, Denmark, pp. 225-232, 2010.

[151] Schroeder, J. Xu, J., Chen, H. and Chau, M. "Automated criminal link analysis based on domain knowledge,*" Journal of the American Society for Information Science and Technology*, 58(6):842_855, Apr. 2007.

[152] Xu, J. and Chen, H. "Fighting organized crimes: using shortest-path algorithms to identify associations in criminal networks". *Decision Support Systems*, 38(3):473–487, 2004.

[153] Taha, K. and Yoo, P. "Using the Spanning Tree of a Criminal Network for Identifying Its Leaders," in *IEEE Transactions on Information Forensics and Security*, 12(2):445-453, Feb. 2017.

[154] Wang, X. and Dong, G. "Research on Money Laundering Detection Based on Improved Minimum Spanning Tree Clustering and Its Application," *System Knowledge Acquisition Modeling.*, Wuhan, China, pp. 62-64, 2009.

[155] Alzaabi, M., Taha, K. and Martin, T. "CISRI: A Crime Investigation System Using the Relative Importance of Information Spreaders in Networks Depicting Criminals Communications," in *IEEE Transactions on Information Forensics and Security*, 10(10):2196-2211, 2015.

[156] Taha, K. and Yoo, P. "Shortlisting the Influential Members of Criminal Organizations and Identifying Their Important Communication Channels," in *IEEE Transactions on Information Forensics and Security*, 14(8):1988-1999, Aug. 2019.

[157] Ozgul,F. and Erdem, Z. "Detecting Criminal Networks Using Social Similarity," In the *IEEE/ACM International Conference on Advances in Social Networks Analysis and Mining*, Istanbul, Turkey, pp. 581-585, 2012.

[158] Jhee, J., Kim, M., Park, M., Yeon, J., Kwak, Y. and Shin, H. "Fast Prediction for Suspect Candidates from Criminal Networks," In *the 2023 IEEE International Conference on Big Data and Smart Computing (BigComp)*, Jeju, Korea, Republic of, pp. 353-355, 2023.

[159] Taha, K. and Yoo, P. "A system for analyzing criminal social networks,", In *the IEEE/ACM International Conference on Advances in Social Networks Analysis and Mining (ASONAM)*, Paris, France, pp. 1017-1023, 2015.

[160] Taha, K. and Yoo, P. "SIIMCO: A Forensic Investigation Tool for Identifying the Influential Members of a Criminal Organization," in *IEEE Transactions on Information Forensics & Security*, 11(4):811-822, 2016.

[161] Ozgul, F., Erdem, Z., Bowerman, C. "Comparison of Feature-Based Criminal Network Detection Models with k-Core and n-Clique," In *the International Conference on Advances in Social Networks Analysis and Mining*, Odense, Denmark, pp. 400-401, 2010.

[162] Chicago Crime Dataset, available at: https://data.cityofchicago.org/Public-Safety/Crimes-2001-to-Present/ijzp-q8t2

[163] San Francisco Crime Dataset, available: https://data.sfgov.org/Public-Safety/Police-Department-Incident-Reports-Historical-2003/tmnf-yvry

[164] Bahulkar, A., Szymanski, B., Baycik, N. and Sharkey, T. "Community Detection with Edge Augmentation in Criminal Networks," In *the IEEE/ACM International Conference on Advances in Social Networks Analysis and Mining (ASONAM)*, Barcelona, Spain, pp. 1168–1175, Aug. 2018.

[165] 171. Morselli, C. and Giguere, C. "Legitimate strengths in criminal networks," *Crime, Law Social Change*, 45(3):185–200, 2006.

[166] Ozgul, F. and Erdem, Z. "Detecting Criminal Networks Using Social Similarity," In *the 2012 IEEE/ACM International Conference on Advances in Social Networks Analysis and Mining*, Istanbul, Turkey, pp. 581-585, 2012.

[167] Ozgul, F and Erdem, Z. "Which crime features are important for criminal network members?," In *the 2013 IEEE/ACM International Conference on Advances in Social Networks Analysis and Mining (ASONAM 2013)*, Niagara Falls, ON, Canada, pp. 1058-1060, 2013

[168] K. P. Sinaga and M. -S. Yang, "Unsupervised K-Means Clustering Algorithm," in *IEEE Access*, vol. 8, pp. 80716-80727, 2020.